\documentclass[runningheads]{llncs}
\usepackage{graphicx}

\usepackage{tikz}
\usepackage{comment}
\usepackage{amsmath,amssymb} %
\usepackage{color}

\usepackage[accsupp]{axessibility}  %

\newif\ifArxivVersion

\ArxivVersiontrue

\usepackage[width=122mm,left=12mm,paperwidth=146mm,height=193mm,top=12mm,paperheight=217mm]{geometry}

\usepackage{subcaption}
\usepackage{tabularx}
\usepackage{multirow}
\usepackage{booktabs}

\usepackage{hhline}
\usepackage{arydshln}
\usepackage{graphbox}
\usepackage[export]{adjustbox}
\usepackage{enumitem}

\newcommand{\fig}[1]{Figure~\ref{fig:#1}}

\newcommand{\tab}[1]{Table~\ref{tab:#1}}

\usepackage[pagebackref=true,breaklinks=true,letterpaper=true,colorlinks,bookmarks=false]{hyperref}
\usepackage{cleveref}

\newcommand{\p}{\mathcal{P}}
\newcommand{\e}{e}
\newcommand{\E}{E}
\def\@onedot{\ifx\@let@token.\else.\null\fi\xspace}

\begin{document}

\title{Realistic One-shot Mesh-based Head Avatars}

\newcommand{\topic}[1]{\vspace{1mm}\noindent\textbf{#1}}
\newcommand{\modelname}{ROME\xspace}
\def\etal{et al\onedot}
\newcommand{\TK}[1]{\authornote{TK}{ballblue}{#1}} %
\setcounter{footnote}{0}  %

\titlerunning{ROME}
\author{Taras Khakhulin\inst{1,2}
\and Vanessa Sklyarova\inst{1,2}
\and\\ Victor Lempitsky\inst{3}
\and  Egor Zakharov \inst{1,2}}

\authorrunning{T. Khakhulin et al.}
\institute{
    Samsung AI Center -- Moscow
    \and Skolkovo Institute of Science and Technology
    \and Yandex Armenia \\[7pt]  \url{https://samsunglabs.github.io/rome/}
}

\maketitle

\begin{abstract}
We present a system for realistic one-shot mesh-based human head avatars creation, ROME for short. Using a \emph{single} photograph, our model estimates a person-specific head mesh and the associated neural texture, which encodes both local photometric and geometric details. The resulting avatars are rigged and can be rendered using a neural network, which is trained alongside the mesh and texture estimators on a dataset of in-the-wild videos. In the experiments, we observe that our system performs competitively both in terms of head geometry recovery and the quality of renders, especially for the cross-person reenactment.
\end{abstract}

\newlength{\wid}
\newlength{\mrgone}
\newlength{\mrgtwo}
\newlength{\mrgthree}

\section{Introduction}

Personalized human avatars are becoming a key technology across several application domains, such as telepresence, virtual worlds, and online commerce. In many practical cases, it is sufficient to personalize only a part of the avatar's body, while the remaining areas can then be picked from a pre-defined library of assets or omitted from the interface. Towards this end, many applications require personalization at the head level, i.e., the creation of person-specific head models, thus making it an important and viable intermediate step between personalizing only the face and the entire body. Alone, face personalization is often insufficient, while the full-body reconstruction remains a complicated task and leads to the reduced quality of the models or requires cumbersome data collection.

Acquiring human avatars from a single photograph (in a ``one-shot'' setting) offers the highest convenience for the end-user. However, their creation process is particularly challenging and requires strong priors on human geometry and appearance. To this end, parametric models are long known to offer good personalization solutions~\cite{Blanz1999AMM} and were recently adapted to one-shot performance~\cite{Feng2020LearningAA,Guo2020Towards,Sanyal2019LearningTR}. Such models can be learned from a relatively small dataset of 3D scans and represent geometry and appearance via textured meshes, making them compatible with many computer graphics applications and pipelines. However, they cannot be trivially expanded to the whole head region due to the large geometric variability of the non-facial parts such as hair and neck.

Our proposed system addresses this issue and allows parametric face models to represent the non-facial parts. In order to handle the increased geometric and photometric variability, we train our method on a large dataset of in-the-wild videos~\cite{Chung2018VoxCeleb2DS} and use neural networks to parameterize both the geometry and the appearance. For the appearance modeling, we follow the deferred neural rendering~\cite{Thies2019DeferredNR} paradigm and employ a combination of neural textures and rendering networks. In addition, a neural rendering framework~\cite{Ravi2020Pytorch3D} is used to enable end-to-end training and achieve high visual realism of the resulting head models. After training, the geometric and appearance networks can be conditioned on the information extracted from a single photograph, enabling one-shot realistic avatar generation.

To the best of our knowledge, our system is the first that is capable of creating realistic personalized human head models in a rigged mesh format from a single photograph. This distinguishes our model from a growing class of approaches that a) recover neural head avatars without explicit geometry~\cite{Wang2021OneShotFN,Siarohin2019FirstOM,Zakharov2019FewShotAL,Zakharov2020FastBN}, b) can personalize the face region but not the whole head~\cite{Thies2016Face,Kim2018Deep,Blanz1999AMM,Feng2020LearningAA}, and c) from commercial systems that create non-photorealistic mesh avatars from a single image~\cite{AvatarSDK,Pinscreen}. Alongside our main model, we also discuss its simplified version based on a linear blendshapes basis and show how to train it using the same video dataset. Below, we refer to the avatars generated by our system as ROME avatars (Realistic One-shot Mesh-based avatars).

\begin{figure}[t!]
    \setlength{\wid}{0.11\textwidth}
    \setlength{\mrgone}{-0.1cm}
    \setlength{\mrgtwo}{0.12cm}
    \centering
    \begin{tabular}{cccc | cccc}
        \includegraphics[width=\wid]{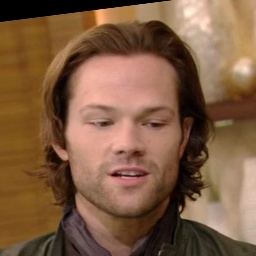} & \hspace{\mrgone}
        \includegraphics[width=\wid]{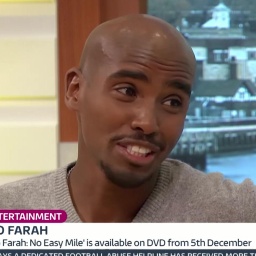} & \hspace{\mrgone}
        \includegraphics[width=\wid]{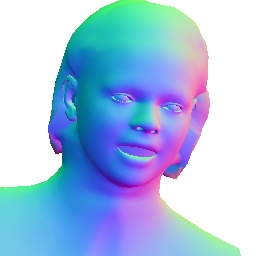} & \hspace{\mrgone}
        \includegraphics[width=\wid]{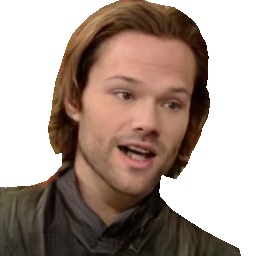} \hspace{0.1cm} & \hspace{0.1cm}
        \includegraphics[width=\wid]{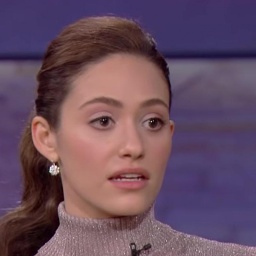} & \hspace{\mrgone}
        \includegraphics[width=\wid]{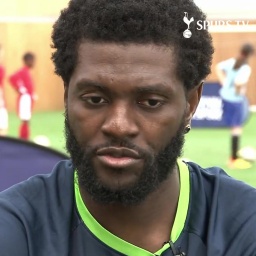} & \hspace{\mrgone}
        \includegraphics[width=\wid]{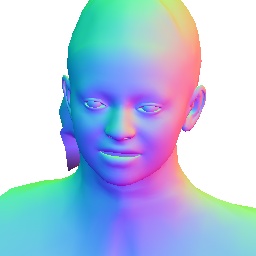} & \hspace{\mrgone}
        \includegraphics[width=\wid]{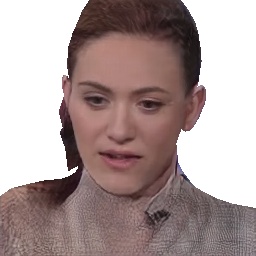} \\
        \includegraphics[width=\wid]{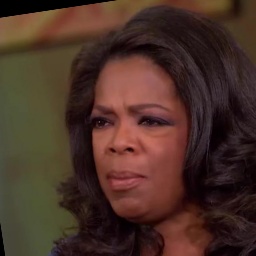} & \hspace{\mrgone}
        \includegraphics[width=\wid]{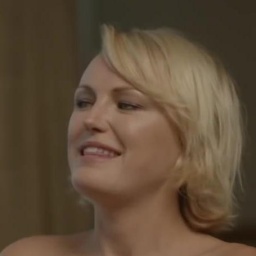} & \hspace{\mrgone}
        \includegraphics[width=\wid]{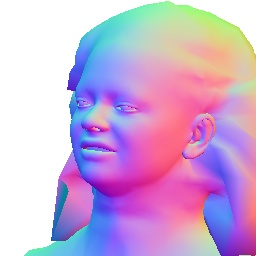} & \hspace{\mrgone}
        \includegraphics[width=\wid]{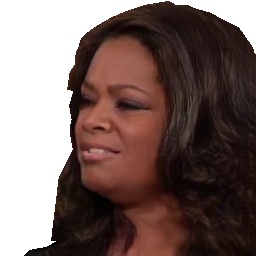} \hspace{0.1cm} & \hspace{0.1cm}
        \includegraphics[width=\wid]{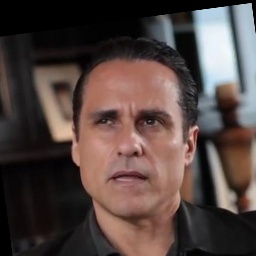} & \hspace{\mrgone}
        \includegraphics[width=\wid]{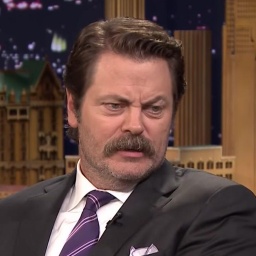} & \hspace{\mrgone}
        \includegraphics[width=\wid]{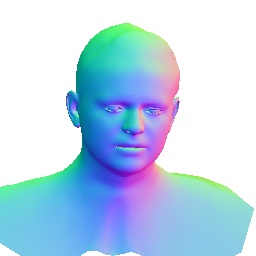} & \hspace{\mrgone}
        \includegraphics[width=\wid]{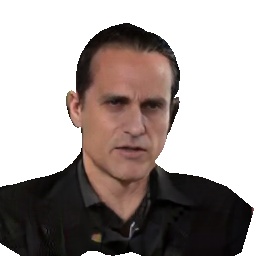}
        \\
        \textbf{Source} & \hspace{\mrgone} \textbf{Driver} & \hspace{\mrgone} \textbf{Mesh} & \hspace{\mrgone} \textbf{Render} & \hspace{\mrgone} \textbf{Source} & \hspace{\mrgone}
        \textbf{Driver} & \hspace{\mrgone} \textbf{Mesh} & \hspace{\mrgone} \textbf{Render}
    \end{tabular}
    \captionof{figure}{Our system creates realistic mesh-based avatars from a single \textbf{source} photo. These avatars are rigged, i.e., they can be driven by the animation parameters from a different \textbf{driving} frame. At the same time, our obtained \textbf{meshes} and \textbf{renderes} achieve a high degree of personalization in both appearance and geometry and are trained in an end-to-end fashion on a dataset of in-the-wild videos without any additional 3D supervision.}
    \label{fig:teaser}
\end{figure}

\section{Related work}

\topic{Parametric models of human faces.} Over the recent decades, 3D face reconstruction methods have been actively employed to tackle the problems of face tracking and alignment~\cite{Hassner2015EffectiveFF,Guo2020Towards}, face recognition~\cite{Blanz2002FaceIA,Tran2017RegressingRA}, and generative modelling~\cite{Thies2016Face,Kim2018Deep,Mildenhall2020NeRFRS,Ramon2021H3DNetFH,Lombardi2018DeepAM,Lombardi2019NeuralV}. In all these scenarios, statistical mesh-based models, aka parametric models~\cite{Blanz1999AMM}, remain one of the widely used tools~\cite{Egger20203DMF,Ploumpis2021TowardsAC}. State-of-the-art parametric models for human heads consist of rigged meshes~\cite{Li2017LearningAM} which support a diverse range of animations via disentangled shape and expression blendshapes and rigid motions for the jaw, neck, and eyeballs. However, they only provide reconstructions for the face, ears, neck, and forehead regions, limiting the range of applications. Including full head reconstruction (i.e., hair and shoulders) into these parametric models is possible, but existing approaches require significantly more training data to be gathered in the form of 3D scans. Instead, in our work, we propose to leverage existing large-scale datasets~\cite{Chung2018VoxCeleb2DS} of in-the-wild videos via the learning-by-synthesis paradigm without any additional 3D annotations.

\topic{Neural 3D human head models.} While parametric models provide sufficient reconstruction quality for many downstream applications, they are not able to depict the fine appearance details that are needed for photorealistic modelling. In recent years, the problem of representing complex geometry and appearance of humans started being addressed using high-capacity deep neural networks. Some of these works use strong human-specific priors~\cite{Ramon2021H3DNetFH,Feng2020LearningAA,Saito2020PIFuHDMP,Lombardi2021MixtureOV}, while others fit high-capacity networks to data without the use of such priors~\cite{Ma2021SCALEMC,Park2019DeepSDFLC,Mildenhall2020NeRFRS,Lombardi2019NeuralV,Lombardi2018DeepAM,Kellnhofer2021NeuralLR,Oechsle2021UNISURFUN}. The latter methods additionally differ by the type of data structure used to represent the geometry, namely, mesh-based~\cite{Feng2020LearningAA,Lombardi2019NeuralV,Lombardi2018DeepAM,grassal2021neural}, point-based~\cite{Ma2021SCALEMC,Zakharkin2021PointBasedMO}, and implicit models~\cite{Mildenhall2020NeRFRS,Park2019DeepSDFLC,Ramon2021H3DNetFH,Saito2020PIFuHDMP,Oechsle2021UNISURFUN,Lombardi2021MixtureOV,Yenamandra2021i3DMMDI}. 
Additionally, recently there have emerged the hybrid models~\cite{Yufeng2022IMAvatar,Gafni2021CVPR} where authors integrate face priors from parametric models with implicit representations to learn geometry and rendering for the specific person from the video. 

However, mesh-based models arguably represent the most convenient class of methods for downstream applications. They provide better rendering quality and better temporal stability than point-based neural rendering. Also, unlike methods based on implicit geometry, mesh-based methods preserve topology and rigging capabilities and are much faster during fitting and rendering. However, current mesh-based methods either severely limit the range of deformations~\cite{Feng2020LearningAA}, making it infeasible to learn more complex geometry like hair and clothed shoulders, operate in the multi-shot setting~\cite{grassal2021neural} or require 3D scans as training data~\cite{Lombardi2019NeuralV,Lombardi2018DeepAM}. Our proposed method is also mesh-based, but we allow the prediction of complex deformations without 3D supervision and using a single image, lifting the limitations of the previous and concurrent works.

\topic{One-shot neural head models.} Advances in neural networks also led to the development of methods that directly predict images using large ConvNets operating in the 2D image domain, with effectively no underlying 3D geometry~\cite{Siarohin2019FirstOM,Zakharov2019FewShotAL,Zakharov2020FastBN} or very coarse 3D geometry~\cite{Wang2021OneShotFN}. These methods achieve state-of-the-art realism~\cite{Wang2021OneShotFN}, use in-the-wild images or videos with no 3D annotations for training, and can create avatars from a single image. However, the lack of an explicit geometric model makes these models incompatible with many real-world applications and limits the span of camera poses that these methods can handle.

\begin{figure}[!t]
    \centering
    \includegraphics[width=\textwidth]{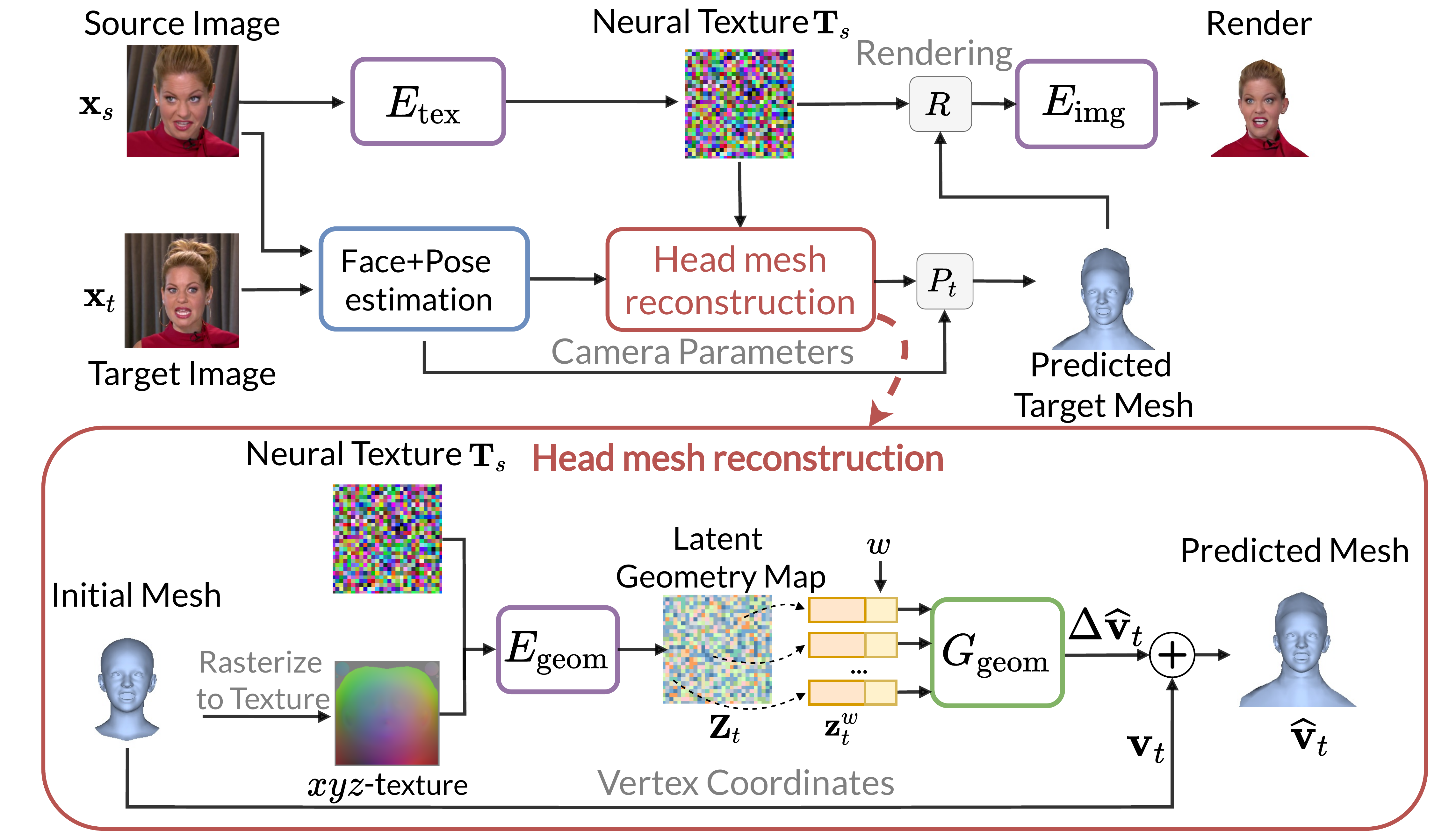}
    \caption{
    Overview of our approach and the detailed scheme of the {\color{red} head mesh reconstruction}. Given the source photo, we first estimate a \emph{neural texture} that encodes local geometric and photometric details of visible and occluded parts. We then use a pre-trained system~\cite{Feng2020LearningAA} for face reconstruction to estimate an initial mesh with a reconstructed facial part. We call this step face and 3D pose estimation. During head mesh reconstruction (bottom), using the estimated neural texture and the initial mesh, we predict the offsets for the mesh vertices, which do not correspond to a face. The offsets are predicted with a combination of a convolutional network $E_\text{geom}$ and a perceptron network $G_\text{geom}$. We then render the personalized head mesh using the camera parameters, estimated by a pre-trained regressor~\cite{Feng2020LearningAA} while superimposing the predicted neural texture. Finally, the rendering network $E_\text{img}$ estimates the RGB image and the mask from the render.
    }
    \label{fig:method}

\end{figure}

\topic{Neural mesh rendering.} Recently, approaches that combine explicit data structures (point clouds or meshes) with neural image generation have emerged. These methods gained popularity thanks to the effectiveness of the pioneering Deferred Neural Rendering system~\cite{Thies2019DeferredNR}, as well as recent advances in differentiable mesh rendering~\cite{Ravi2020Pytorch3D,Liu2019SoftRA,Laine2020diffrast}. Neural mesh rendering uses 2D convolutional networks to model complex photometric properties of surfaces. It achieves high realism of renders with fine details present even when they are missing in the underlying geometric model. In this work, we adapt these advances to human head modelling while training using a large dataset of in-the-wild videos.
\def\x{\mathbf{x}}
\def\X{\mathbf{X}}
\def\m{\mathbf{s}}
\def\v{\mathbf{v}}
\def\V{\mathbf{V}}
\def\e{\mathbf{e}}
\def\E{\mathbf{E}}
\def\T{\mathbf{T}}
\def\Z{\mathbf{Z}}
\def\z{\mathbf{z}}
\def\o{\mathbf{o}}
\def\p{\mathbf{p}}
\def\n{\mathbf{n}}

\section{Method}

Our goal is to build a system that jointly learns to produce photorealistic renders of human heads and estimate their 3D meshes using only a \emph{single image} without any 3D supervision.

To achieve that, we use a large-scale dataset~\cite{Chung2018VoxCeleb2DS} of in-the-wild videos with talking speakers. All frames in each video are assumed to depict the same person in the same environment (defined by lighting, hairstyle, and person's clothing). At each training step, we sample two random frames $\x_s$ and $\x_t$ from a random training video. Our goal is to reconstruct and render the target image $\hat\x_t$ given a) the personal details and the face shape extracted from the source image $\x_s$, as well as b) the head pose, the facial expression, and the camera pose estimated from the target image $\x_t$. The final reconstruction loss is backpropagated and used to update the parameters of the model's components.

After training, we can create a personalized head model by estimating all parameters from a single image. This model can then be \emph{animated} using face tracking parameters extracted from any talking head sequence and rendered from a range of viewpoints similar to those present in the training dataset (\fig{teaser}).

\subsection{Model overview}

In our model, we jointly train multiple neural networks that perform rendering and mesh reconstruction. The training pipeline proceeds as follows (\fig{method}):

    \textbf{Neural texture estimation.} The source image $\x_s$ is encoded into a neural texture $\T_s$, which describes both person-specific appearance and geometry. The encoding is done by a convolutional neural network $E_\text{tex}$.
    
     \textbf{Face and 3D pose estimation.} In parallel, we apply a pre-trained DECA system~\cite{Feng2020LearningAA} for face reconstruction to both the source and the target image, which estimates facial shape, expression, and head pose. Internally, it uses the FLAME parametric head model~\cite{Li2017LearningAM}, which includes mesh topology, texture mapping, and blendshapes. We use the face shape from the source image $\x_s$ as well as the facial expression and the camera pose from the target image $\x_t$ for further processing.
    
     \textbf{Head mesh reconstruction.} The vertices of the DECA mesh with personalized face region and generic non-facial parts are rendered into an $xyz$-coordinate texture using the predefined texture mapping. The $xyz$-texture and the neural texture $\T_s$ are concatenated and processed with the U-Net network~\cite{Ronneberger2015UNetCN} $E_\text{geom}$ into a new texture map $\Z_t$, called \textit{latent geometry} map. The 3D displacements for each mesh vertex are then decoded independently by the multi-layer perceptron $G_\text{geom}$ that predicts a 3D offset $\Delta \hat\v$ for each vertex. This step reconstructs the personalized model for non-face parts of the head mesh. The obtained reconstructions are compatible with the topology/connectivity of the FLAME mesh~\cite{Li2017LearningAM}.
    
     \textbf{Deferred neural rendering.} The personalized head mesh is rendered using the pose estimated by DECA for the target image and with the superimposed neural texture. The rendered neural texture and the rasterized surface normals are concatenated and processed by the decoding (rendering) U-Net network $E_\text{img}$ to predict the rendered image $\hat\x_t$ and the segmentation mask $\hat\m_t$. During training, the difference between the predictions and the ground truths is used to update all components of our system.

Below we discuss our system and its training process in more detail. We also describe a training procedure for a simplified version of our model, which represents head geometry using a linear basis of blendshapes.

\subsection{Parametric face modeling}

Our method uses a predefined head mesh with the corresponding topology, texture coordinates $w$, and rigging parameters, which remain fixed for all avatars. More specifically, we use FLAME~\cite{Li2017LearningAM} head model that has $N$ base vertices $\v_\text{base}~\in~\mathbb{R}^{3N}$, and two sets of $K$ and $L$ basis vectors (blendshapes) that encode shape $\mathcal{B}~\in~\mathbb{R}^{3N \times K}$ and expression $\mathcal{D}~\in~\mathbb{R}^{3N \times L}$. The reconstruction process is carried out in two stages. First, the basis vectors are blended using the person- and expression-specific vectors of linear coefficients $\phi$ and $\psi$. Then, the linear blend skinning~\cite{Li2017LearningAM} function $\mathcal{W}$ is applied, parameterized by the angles $\theta$, which rotates the predefined groups of vertices around linearly estimated joints. The final reconstruction in world coordinates can be expressed as follows: 
\begin{equation*}
    \v (\phi, \psi, \theta) = \mathcal{W} \Big( \v_\text{base} + \mathcal{B} \phi + \mathcal{D} \psi,\ \theta \Big).
\end{equation*}

In previous works~\cite{Thies2016Face}, a similar set of parameters for the 3DMM~\cite{Blanz1999AMM} parametric model was obtained via photometric optimization. More recently, learning-based methods~\cite{Feng2020LearningAA,Guo2020Towards} capable of feed-forward estimation started to emerge. In our work, given an input image, we use a pre-trained feed-forward DECA system~\cite{Feng2020LearningAA} to estimate $\phi, \psi$, $\theta$, and the camera parameters.

During training, we apply DECA to both source image $\x_s$ and the target image $\x_t$. The face shape parameters $\phi_s$ from the source image $\x_s$ alongside the expression $\psi_t$, head pose $\theta_t$ and camera parameters from the target image $\x_t$ are then used to reconstruct the initial FLAME vertices $\v_t = \v(\phi_s, \psi_t, \theta_t)$, as well as camera transform $\mathcal{P}_t$.

\subsection{Head mesh reconstruction}

The FLAME vertices $\v_t$ estimated by DECA provide good reconstructions for the face region but lack any person-specific details in the remaining parts of the head (hair and shoulders). To alleviate that, we predict person-specific mesh offsets for non-facial regions while preserving the face shape predicted by DECA. We additionally exclude ear regions since their geometry in the initial mesh is too complex to be learned from in-the-wild video datasets.

These mesh offsets are estimated in two steps. First, we encode both the $xyz$-coordinate texture and the neural texture $\T_s$ into the latent geometry texture map $\Z_t$ via a U-Net network $E_\text{geom}$. It allows the produced latent map to contain both positions of the initial vertices $\v_t$ and their semantics, provided by the neural texture. 

From $\Z_t$ we obtain the vectors $\z_t^w$ by bilinear interpolation at the fixed texture coordinates $w$. The vectors $\z_t^w$ and their coordinates $w$ are then concatenated and passed through a multi-layer perceptron $G_\text{geom}$ to predict the coefficients $\hat{\mathbf{m}}_t \in \mathbb{R}^{3N \times 3}$ independently for each vertex in the mesh. These coefficients are multiplied elementwise by the normals $\n_t$, calculated for each vertex in $\v_t$, to obtain the displacements: $\Delta \hat\v_t = \hat{\mathbf{m}} \odot \n_t$. These displacements are then zeroed out for face and ear regions, and the final reconstruction in world coordinates is obtained as follows: $ \hat\v_t = \v_t + \Delta \hat\v_t$.

\subsection{Deferred neural rendering}

We render the reconstructed head vertices $\hat\v_t$ using the topology and texture coordinates $w$ from the FLAME model with the superimposed neural texture $\T_s$. For that, we use a differentiable mesh renderer $\mathcal{R}$~\cite{Ravi2020Pytorch3D} with the camera transform $\mathcal{P}_t$ estimated by DECA for the target image $\x_t$.

The resulting rasterization, which includes both the neural texture and the surface normals, is processed by the rendering network $E_\text{img}$ to obtain the predicted image $\hat\x_t$ and the segmentation mask $\hat\m_t$. $E_\text{img}$ consists of two U-Nets that separately decode an image and a mask. The result of the deferred neural rendering is the reconstruction of the target image $\hat\x_t$ and its mask $\hat\m_t$, which are compared to the ground-truth image $\x_t$ and mask $\m_t$ respectively.

\subsection{Training objectives}

In our approach, we learn the geometry of hair and shoulders, which are not reconstructed by the pre-trained DECA estimator, without any ground-truth 3D supervision during training. For that we utilize two types of objectives: segmentation-based geometric losses $\mathcal{L}_\text{geom}$ and photometric losses $\mathcal{L}_\text{photo}$.

We found that explicitly assigning subsets of mesh vertices to the neck and the hair regions helps a lot with the quality of final deformations. It allows us to introduce a topological prior for the predicted offsets, which is enforced by. %

To evaluate the geometric losses, we calculate two separate occupancy masks using a soft rasterization operation~\cite{Liu2019SoftRA}. First, $\hat\o_t^\text{hair}$ is calculated with detached neck vertices, so that the gradient flows through that mask only to the offsets corresponding to the hair vertices, and then $\hat\o_t$ is calculated with detached hair vertices. We match the hair occupancy mask to the ground-truth mask $\m^\text{hair}_t$ (which covers the hair, face, and ears), and the estimated occupancy mask to the whole segmentation mask $\m_t$: $\mathcal{L}_\text{occ} = \lambda_\text{hair} \big\| \hat\o_t^\text{hair} - \m_t^{\text{hair}} \big\|_2^2 + \lambda_\text{o} \big\| \hat\o_t - \m_t \big\|_2^2.$

We also use an auxiliary Chamfer loss to ensure that the predicted mesh vertices cover the head more uniformly. Specifically, we match the 2D coordinates of the mesh vertices projected into the target image to the head segmentation mask. We denote the subset of predicted mesh vertices, visible in the target image, as $\hat\p_t = \mathcal{P}'_t(\hat\v_t)$, and the number of these vertices as $N_t$, so that $\hat\p_t \in \mathbf{R}^{N_t \times 2}$. Notice that operator $\mathcal{P}'_t$ here not only does the camera transformation but also discards the $z$ coordinate of the projected mesh vertices. To compute the loss, we then sample $N_t$ 2D points from the segmentation mask $\m_t$ and estimate the Chamfer distance between the sampled set of points $\p_t$ and the vertex projections:
\begin{equation*}
\begin{split}
    \mathcal{L}_\text{chm} = & \frac{1}{2N_t} \sum_{\hat{p}_t \in \hat\p_t} \Big\| \hat{p}_t - \arg\min_{p \in \p_t}\big\| p - \hat{p}_t \big\| \Big\| + \\
    & \frac{1}{2N_t} \sum_{p_t \in \p_t} \Big\| p_t - \arg\min_{\hat{p} \in \hat\p_t}\big\| \hat{p} - p_t \big\| \Big\|.
\end{split}
\end{equation*}

Lastly, we regularize the learned geometry using the Laplacian penalty~\cite{SorkineHornung2005LaplacianMP}. Initially, we found that regularizing offsets $\Delta \hat\v$ worked better than regularizing full coordinates $\hat\v$ and stuck to that approach for all experiments. Our version of the Laplacian loss can be written as:
\begin{equation*}
    \mathcal{L}_{\text{lap}} = \dfrac{1}{V}\sum_{i=1}^V \Big\| \Delta \hat\v_i - \dfrac{1}{\mathcal{N}(i)}\sum\limits_{j \in \mathcal{N}(i)}  \Delta \hat\v_j \Big\|_1,
\end{equation*}
where $\mathcal{N}(i)$ denotes a set indices for vertices adjacent to the $i$-th vertex in the mesh.

We also use photometric optimization that matches the predicted and the ground truth images. This allows us to obtain photorealistic renders and aid in learning proper geometric reconstructions. We utilize perceptual loss $\mathcal{L}_\text{per}$~\cite{Johnson2016PerceptualLF}, the face recognition loss $\mathcal{L}_\text{idt}$~\cite{Cao2018VGGFace2AD} and adversarial loss $\mathcal{L}_\text{adv}$~\cite{Goodfellow2014GenerativeAN,Wang2018HighResolutionIS}. We also use the Dice loss $\mathcal{L}_\text{seg}$~\cite{Milletari2016VNetFC} to match the predicted segmentation masks.

The final objective is weighted sum of the geometric and the photometric losses described above.%

\subsection{Linear deformation model}

In addition to the full non-linear model introduced above, we consider a simplified parametric model with a linear basis of offsets $\Delta \hat\v$. While this model is similar to parametric models~\cite{Li2017LearningAM,zuffi2019threed}, we still do not use 3D scans for training and instead obtain our linear model by ``distilling'' the non-linear model. Additionally, we train a feed-forward estimator that predicts the linear coefficients from the input image.

The motivation for training this additional model is to show that the deformations learned by our method can be approximated using a system with a significantly lower capacity. Such a simple regression model can be easier to apply for inference on low-performance devices.

To train the linear model, we first obtain the basis of offsets $\mathcal{F}\in \mathbb{R}^{3 N \times K}$, which is similar to the blendshapes used in the FLAME parametric model. This basis is obtained by applying a low-rank PCA~\cite{Halko2011FindingSW} to the matrix of offsets $\Delta \hat\V \in  \mathbb{R}^{3 N \times M}$, calculated using $M$ images from the dataset. Following~\cite{Li2017LearningAM}, we discard most of the basis vectors and only keep $K$ components corresponding to maximal singular values. The approximated vertex offsets $\tilde{\v}$ for each image can then be estimated as following $\tilde{\v} = \mathcal{F} \eta$,
where $\eta$ is obtained by applying the pseudo-inverse of a basis matrix $\mathcal{F}$ to the corresponding offsets $\Delta \hat\v$: $\eta = \big( \mathcal{F}^{T}\mathcal{F} \big)^{-1} \mathcal{F}^{T} \Delta\hat\v$

We then train the regression network by estimating a vector of basis coefficients $\eta_t$, given an image $\x_t$. For that, we minimize the mean squared error (MSE) loss $\| \hat\eta_t - \eta_t \|_2^2$ between the estimated coefficients and the ground truth, as well as the segmentation loss $\mathcal{L}_\text{occ}$ and a Chamfer distance between predicted and ground truth meshes.
\begin{figure}[t!]
    \centering    
    \setlength{\wid}{0.156\textwidth}
    \begin{tabular}{cccccccc}
        \vspace{-0.2cm}
        \includegraphics[width=\wid]{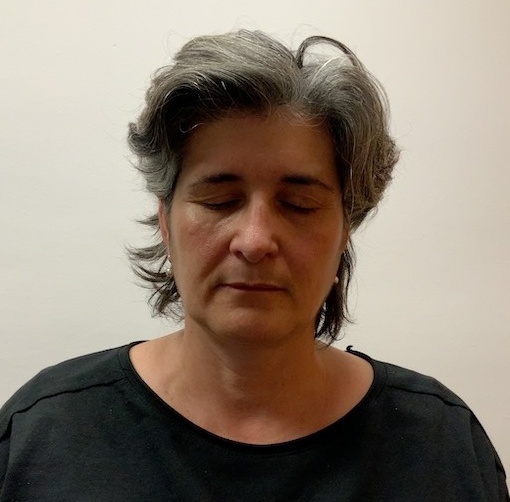} & \hspace{-0.1cm}
        \includegraphics[width=\wid]{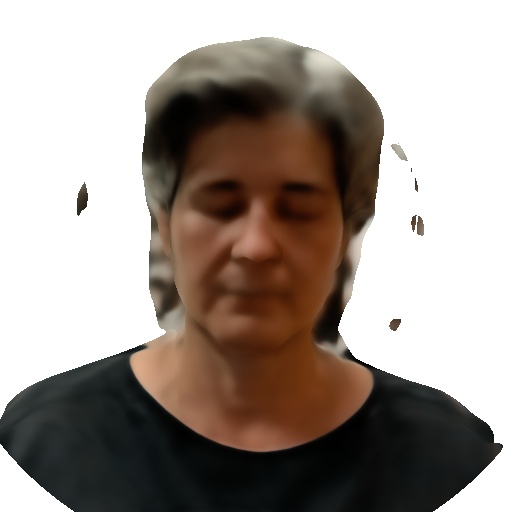} & \hspace{-0.1cm}
        \includegraphics[width=\wid]{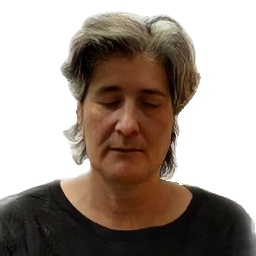} & \hspace{-0.1cm}
        \includegraphics[width=\wid]{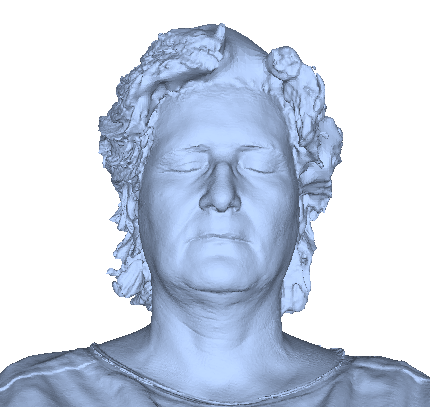} & \hspace{-0.1cm}
        \includegraphics[width=\wid]{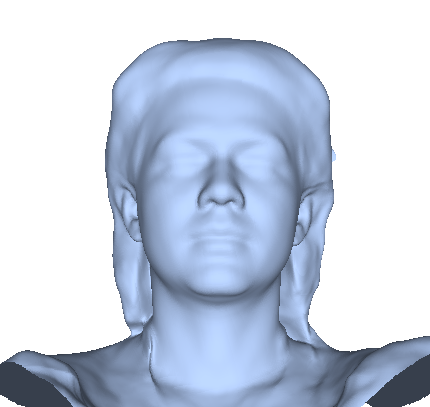} & \hspace{-0.1cm}
        \includegraphics[width=\wid]{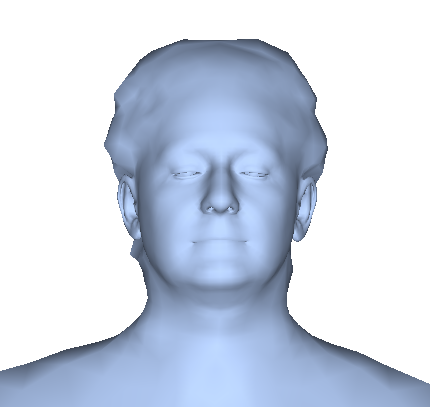} \\ \vspace{-0.07cm}
        \includegraphics[width=\wid]{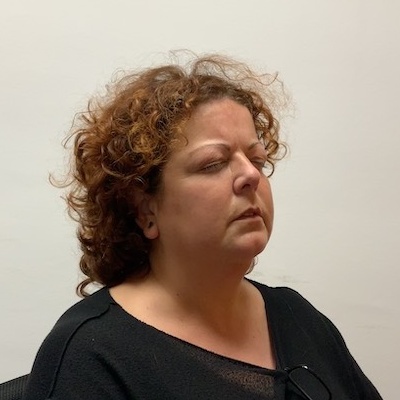} & \hspace{-0.1cm}
        \includegraphics[width=\wid]{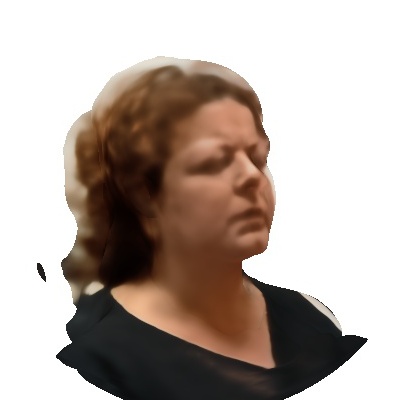} & \hspace{-0.1cm}
        \includegraphics[width=\wid]{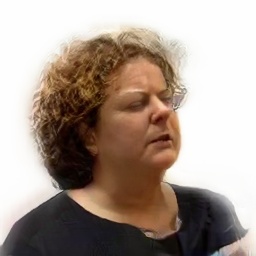} & \hspace{-0.1cm}
        \includegraphics[width=\wid]{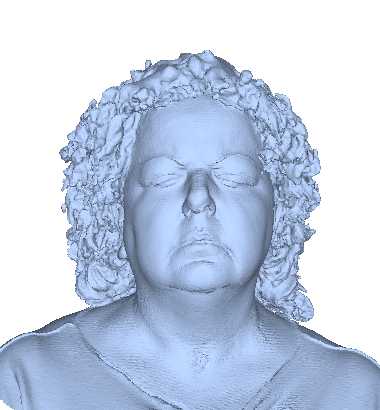} & \hspace{-0.1cm}
        \includegraphics[width=\wid]{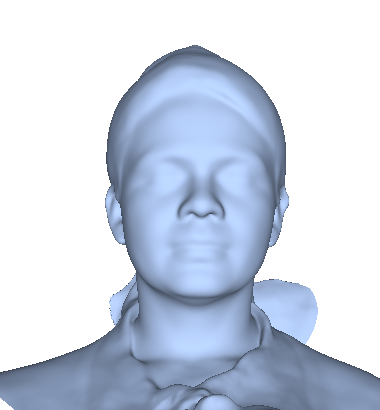} & \hspace{-0.1cm}
        \includegraphics[width=\wid]{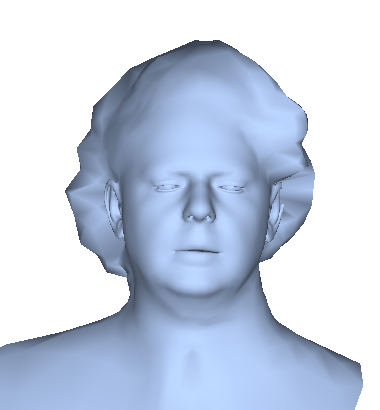} \\
        \includegraphics[width=\wid]{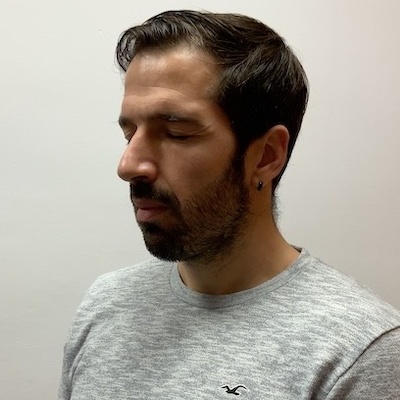} & \hspace{-0.1cm}
        \includegraphics[width=\wid]{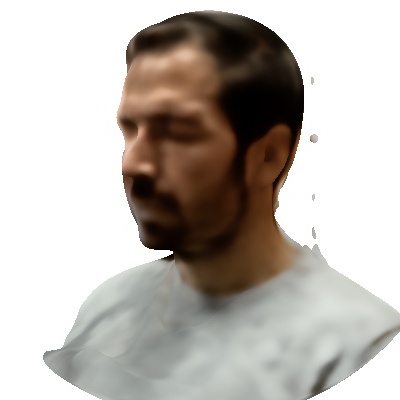} & \hspace{-0.1cm}
        \includegraphics[width=\wid]{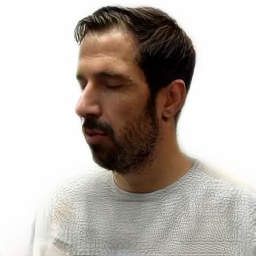} & \hspace{-0.1cm}
        \includegraphics[width=\wid]{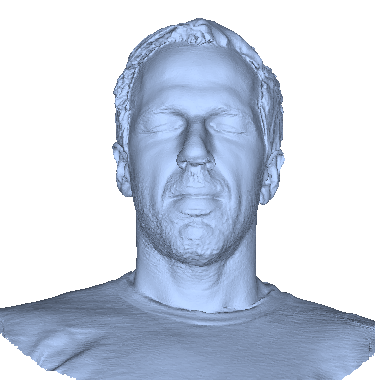} & \hspace{-0.1cm}
        \includegraphics[width=\wid]{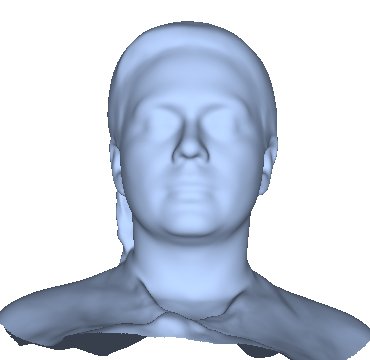} & \hspace{-0.1cm}
        \includegraphics[width=\wid]{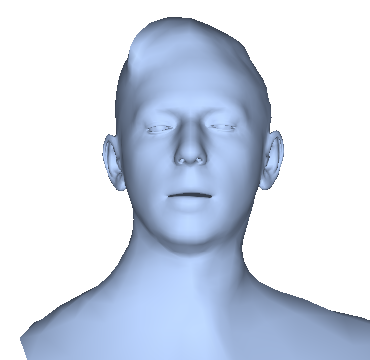} &\hspace{-0.1cm}
        \\
        \textbf{Source}& \hspace{-0.1cm} \textbf{H3D-Net} &\hspace{-0.1cm} \textbf{ROME} & \hspace{-0.1cm} \textbf{GT}  &\hspace{-0.1cm} \textbf{H3D-Net}  &\hspace{-0.1cm} \textbf{ROME}
    \end{tabular}
    \caption{Qualitative comparison of the representative cases from the H3DS dataset. While neither of the two methods achieves perfect results, arguably, ROME achieves more realistic renders and better matches the head geometry than H3D-Net in the single-shot mode. Furthermore, an important advantage of ROME is that the resulting avatars are ready for animation and are obtained in a feed-forward manner without the lengthy fine-tuning process employed by H3D-Net.}
    \label{fig:h3ds_compare}
    \vspace{-0.5cm}
\end{figure}

\section{Experiments}

We train our models on the VoxCeleb2~\cite{Chung2018VoxCeleb2DS} dataset of videos. This large-scale dataset  contains an order of $10^5$ videos of $10^3$ different speakers. It is widely used~\cite{Doukas2020HeadGANVT,Wang2021OneShotFN,Zakharov2020FastBN} to train talking head models. However, the main drawback of this dataset is the mixed quality of videos and the heavy bias towards frontal poses.

To address these well-known limitations, we process this dataset using an off-the-shelf image quality analysis model~\cite{Su2020BlindlyAI} and a 3D face-alignment network~\cite{bulat2017far}. We then filter out the data which has poor quality and non-diverse head rotations. Our final training dataset has $\approx{}15000$ sequences. We note that filtering/pruning does not fully solve the problem of head rotation bias, and our method still works best in frontal views. For more details, please refer to the supplementary materials.

We also use the H3DS~\cite{Ramon2021H3DNetFH} dataset of photos with associated 3D scans to evaluate the quality of head reconstructions.

\subsection{Implementation details}

In the experiments, unless noted otherwise, we train all architectures jointly and end-to-end. We use the following weights: $\lambda_\text{hair}=10$, $\lambda_\text{per}=1$, $\lambda_\text{idt}=0.1$, $\lambda_\text{adv}=0.1$, $\lambda_\text{seg}=10$, and enable the neck and the 2D Chamfer loss $\lambda_\text{chm} = 0.01$) and $\lambda_\text{lap} = 10$. We ablate all geometry losses and method parts below.

We train our models at $256 \times 256$ resolution using ADAM~\cite{Kingma2015AdamAM} with the fixed learning rate of $10^{-4}$, $\beta_1 = 0$, $\beta_2 = 0.999$, and a batch size of $32$. For more details, please refer to the supplementary materials.

\subsection{Evaluation}

\begin{table}[t!]
\centering
    \caption{Evaluation results on the H3DS dataset in the one-shot scenario for our models, H3D-Net, and DECA. We compute Chamfer distance (lower is better) across all available scans, reconstructed from three different viewpoints. Both of the ROME variants significantly exceed H3D-Net in the one-shot reconstruction quality.}\label{tab:h3d_compr}
    \vspace{0.3cm}
\begin{minipage}{0.7\textwidth}\resizebox{\textwidth}{!}{
        \centering
        \begin{tabular}{ lcccc} 
            Method & DECA & H3D-Net & ROME & LinearROME  \\ 
            \hline 
            Chamfer Distance & $15.0$ & $15.1$  & $12.6$ & $12.5$ \\
        \end{tabular}
    }\end{minipage}
    \label{tab:h3d_q}
    \vspace{-0.5cm}
\end{table}

\topic{3D reconstruction.}
We evaluate our head reconstruction quality using a novel H3DS dataset~\cite{Ramon2021H3DNetFH}. We compare against the state-of-the-art head reconstruction method H3D-Net~\cite{Ramon2021H3DNetFH}, which uses signed distance functions to represent the geometry. While providing great reconstruction quality in the sparse-view scenario, their approach has several limitations. For example, H3D-Net requires a dataset of full head scans to learn the prior on head shapes. Additionally, its results do not have fixed topology or rigging and their method requires fine-tuning per scene, while our method works in a feed-forward way.

We carry out the comparison with H3D-Net in a single-view scenario, which is native for our method but is beyond the capabilities stated by the authors in the original publication~\cite{Ramon2021H3DNetFH}. However, to the best of our knowledge, H3D-Net is the closest system to ours in single-view reconstruction capabilities (out of all systems with either their code or results available). Additionaly, we tried to compare our system with PIFuHD~\cite{saito2020pifuhd}, which unfortunately failed to work with heads images without body (see supplementary).

We show qualitative comparison in~\fig{h3ds_compare}. We evaluate our method and H3D-Net both for frontal- and side-view reconstruction. We note the significant overfitting of H3D-Net to the visible hair geometry, while our model provides reconstructions more robust to the change of viewpoint.

In total, we compared our models on all scans available in the test set of the H3DS dataset, and each scan was reconstructed from three different viewpoints. We provide the measured mean Chamfer distance both for our method and baselines across all scans in Tab.~\ref{tab:h3d_compr}. 

\begin{figure}[t!]
    \centering    
    \setlength{\wid}{0.155\textwidth}
    \begin{tabular}{cccccc}
        \includegraphics[width=\wid]{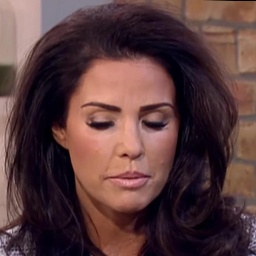} &
        \includegraphics[width=\wid]{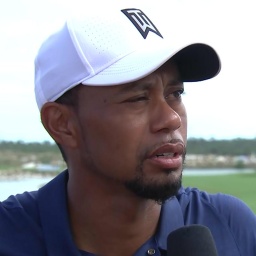} & 
        \includegraphics[width=\wid]{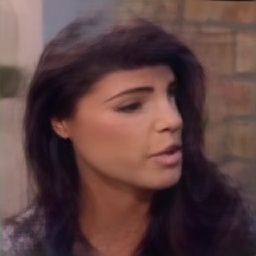} & 
        \includegraphics[width=\wid]{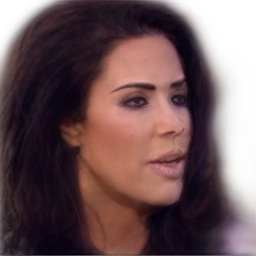} & 
        \includegraphics[width=\wid]{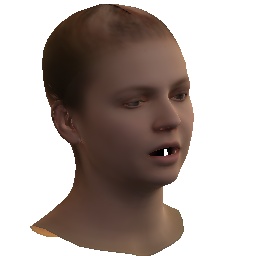} & 
        \includegraphics[width=\wid]{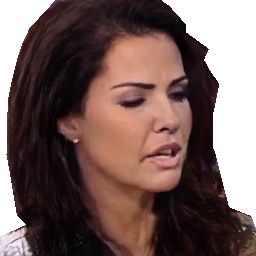} \\ %
        \includegraphics[width=\wid]{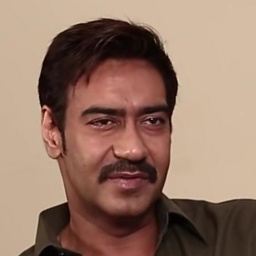} & 
        \includegraphics[width=\wid]{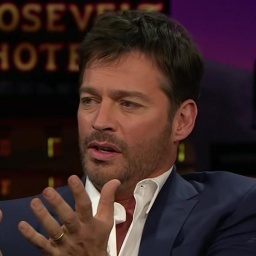} & 
        \includegraphics[width=\wid]{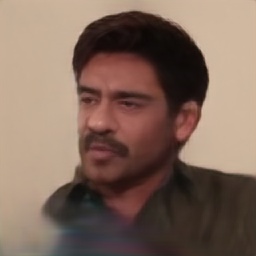} & 
        \includegraphics[width=\wid]{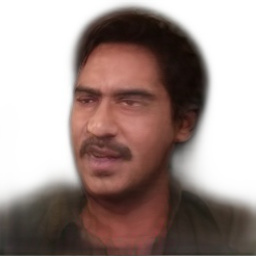} & 
        \includegraphics[width=\wid]{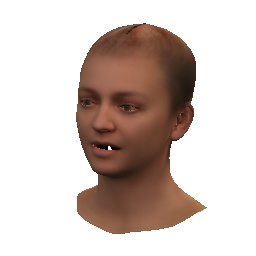} & 
        \includegraphics[width=\wid]{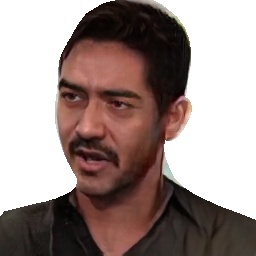} \\ %
        \includegraphics[width=\wid]{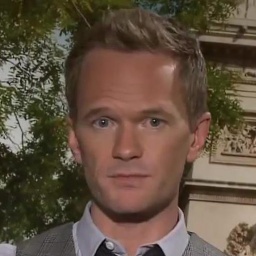} & 
        \includegraphics[width=\wid]{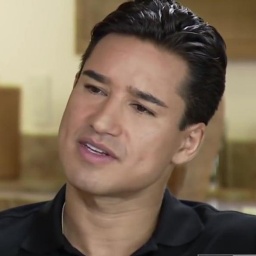} & 
        \includegraphics[width=\wid]{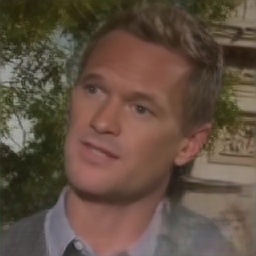} & 
        \includegraphics[width=\wid]{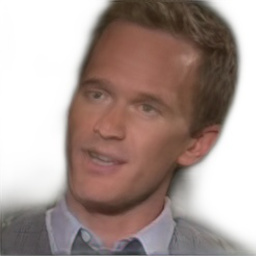} & 
        \includegraphics[width=\wid]{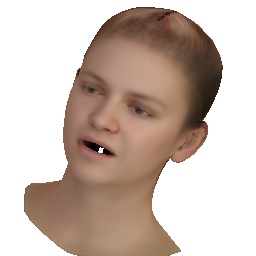} & 
        \includegraphics[width=\wid]{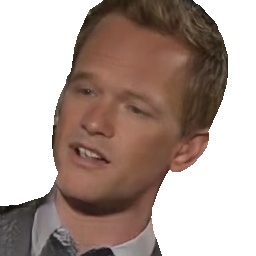} \\ 
        \textbf{Source} & \textbf{Driver} & \textbf{FOMM}  & \textbf{Bi-Layer} & \textbf{FLAMETex} & \textbf{ROME}
    \end{tabular}
    \caption{Comparison of renders on a VoxCeleb2 dataset. The task is to reenact the \textbf{source} image with the expression and pose of the \textbf{driver} image. Here, we picked diverse examples in terms of pose variation to highlight the differences in performance of compared methods. We observe that for the large head pose rotations, purely neural-based methods (\textbf{FOMM}, \textbf{Bi-Layer}) struggle to maintain consistent quality. In contrast, our rendering method produces images that are more robust to pose changes. Admittedly, for small pose changes, neural-based methods exhibit a smaller identity gap than ROME (bottom row) and overall outperform our method in terms of rendering quality. As a reference, we also include a non-neural \textbf{FLAMETex} rendering method, which is employed in state-of-the-art one-shot face reconstruction systems~\cite{Feng2020LearningAA} but is not able to personalize the avatar at the head level.}
    \label{fig:vc2_render_compare}
    \vspace{-0.5cm}
\end{figure}

\topic{Rendering.} We evaluate the quality of our renders on the hold-out subset VoxCeleb2 dataset. We use a cross-driving comparison scenario for qualitative comparison to highlight the animation capabilities of our method, and self-driving scenario for quantitative comparison.

First, we compare with a FLAMETex~\cite{Li2017LearningAM} rendering system, which works explicitly with mesh rendering.
From the source image, FLAMETex estimates the albedo via a basis of RGB textures, and then combines it with predicted scene-specific shading. In contrast, our method predicts a rendered image directly and avoids the complexity of explicit albedo-shading decomposition.

We then compare with publicly available geometry-free rendering methods, which were trained on the same dataset. For that, we use the First-Order Motion Model (FOMM)~\cite{Siarohin2019FirstOM}, the Bi-Layer Avatar Model~\cite{Zakharov2020FastBN} and recently proposed Thin-Plate-Spline-Motion-Mode (TPSMM)~\cite{TPS22}. Both these systems bypass explicit 3D geometry estimation and rely only on learning the scene structure via the parameters of generative ConvNets. Other methods~\cite{Wang2021OneShotFN,Doukas2020HeadGANVT}, which internally utilize some 3D structures, like camera rotations, were out of the scope of our comparison due to the unavailability of pre-trained models.

\begin{table}[t]
    \caption{Here we present the quantitative results on the VoxCeleb2-HQ dataset in the self-reenactment and cross-reenactment modes. Our ROME system performs on par with FOMM and TPSMM in self-reenactment, notably outperforming them in the most perceptually-plausible LPIPS metrics. On the contrary, in the cross-driving scenario, when the task is complex for pure neural-based systems, our method obtains better results.}\label{tab:vc2_quant_comp}
    \vspace{0.3cm}
    \begin{subtable}{\textwidth}
        \centering
        \begin{tabular}{ lcccccc cc cccccc} %
                         & \multicolumn{6}{c}{\text{self-reenactment}} 
            && \multicolumn{6}{c}{\text{cross-reenactment}} \\
         \cmidrule{2-7} \cmidrule{9-14}  
            Method & \multicolumn{2}{c}{LPIPS$\downarrow$} & \multicolumn{2}{c}{SSIM$\uparrow$} & \multicolumn{2}{c}{PSNR$\uparrow$} & &
            \multicolumn{2}{c}{FID$\downarrow$} &
            \multicolumn{2}{c}{CSIM$\uparrow$} &
            \multicolumn{2}{c}{IQA$\uparrow$}
            \\
            \hline
            FOMM     & \multicolumn{2}{c}{$0.09$} & \multicolumn{2}{c}{$0.87$} & \multicolumn{2}{c}{$25.8$} & &
            \multicolumn{2}{c}{$52.95$} & \multicolumn{2}{c}{$0.53$} & \multicolumn{2}{c}{$55.9$} 
            \\
            Bi-Layer & \multicolumn{2}{c}{$0.08$} & \multicolumn{2}{c}{$0.83$} & \multicolumn{2}{c}{$23.7$} &&
             \multicolumn{2}{c}{$51.4$} & \multicolumn{2}{c}{$0.56$} & \multicolumn{2}{c}{$50.48$}\\
            TPSMM     & \multicolumn{2}{c}{$0.09$} & \multicolumn{2}{c}{$0.85$} & \multicolumn{2}{c}{$26.1$} & &
            \multicolumn{2}{c}{$49.27$} & \multicolumn{2}{c}{$0.57$} & \multicolumn{2}{c}{$59.5$} \\
            ROME     & \multicolumn{2}{c}{$0.08$} & \multicolumn{2}{c}{$0.86$} & \multicolumn{2}{c}{$26.2$} & & \multicolumn{2}{c}{$45.32$} & \multicolumn{2}{c}{$0.62$} & \multicolumn{2}{c}{$66.3$}\\
            
        \end{tabular}
    \end{subtable}
    \vspace{-0.5cm}
\end{table}

We present the qualitative comparison in \fig{vc2_render_compare}, and a quantitative comparison across a randomly sampled hold-out VoxCeleb2 subset in~\tab{vc2_quant_comp}. 
We restrict the comparison to the face and hair region as the shoulder pose is not controlled by our method (driven by DECA parameters), which is admittedly a limitation of our system. We thus mask the results according to the face and hair mask estimated from the ground truth image.

Generally, we observe that over the entire test set, the quality of ROME avatars in the self-reenactment mode is similar to FOMM and better than the Bi-layer model. 
For the cross-reenactment scenario, our model is clearly better than both alternatives according to three metrics, that help to asses unsupervised quality of the images in three aspects: realism, identity preservation and blind quality of the image. The huge gap for IQA~\cite{Su2020BlindlyAI} and FID~\cite{fid} is also noticeable in the qualitative comparison, especially for strong pose change (see CSIM~\cite{Zakharov2019FewShotAL} column in Tab.~\ref{tab:vc2_quant_comp}).
The PSNR and SSIM metrics penalize slight misalignments between the sharp ground truth and our renderings much stronger than the blurriness in FOMM reconstructions. The advatage of ROME avatar is noticable even for self-driving case according to LPIPS. We provide a more extensive qualitative evaluation in the supplementary materials.

\subsection{Linear basis experiments}

\newlength{\mrg}
\newlength{\bmrg}
\begin{figure}[t!]
    \centering
    \setlength{\wid}{0.12\textwidth}
    \setlength{\mrg}{0.0cm}
    \setlength{\bmrg}{0.01cm}
        \begin{tabular}{cc cc ccc}
            \includegraphics[width=\wid]{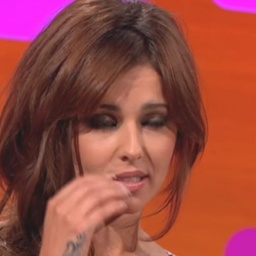} \hspace{\mrg} & 
            \includegraphics[width=\wid]{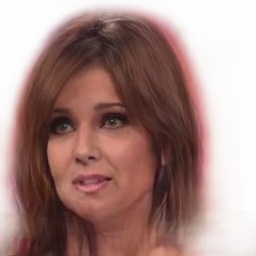} \hspace{\mrg} & 
            \includegraphics[width=\wid]{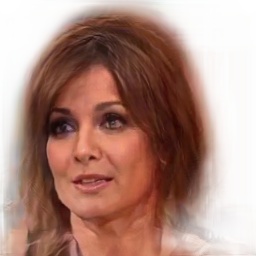} \hspace{\mrg} & 
           \includegraphics[width=\wid]{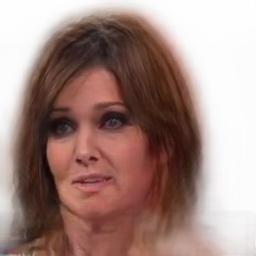} \hspace{\mrg} & 
           \includegraphics[width=\wid]{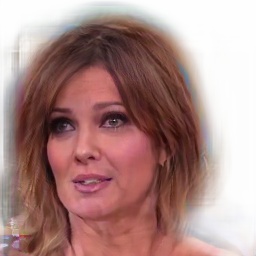} \hspace{\mrg} & 
           \includegraphics[width=\wid]{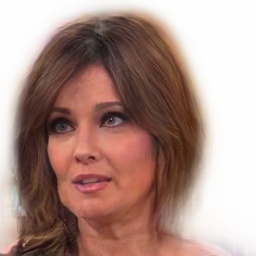} \hspace{\mrg} & 
             \includegraphics[width=\wid]{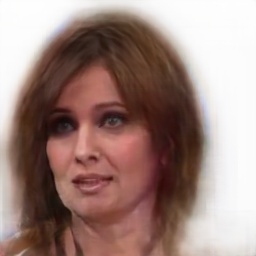} \hspace{\mrg} \\
             
             \includegraphics[width=\wid]{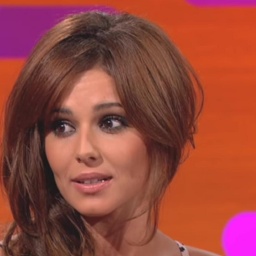} \hspace{\mrg} & 
            \includegraphics[width=\wid]{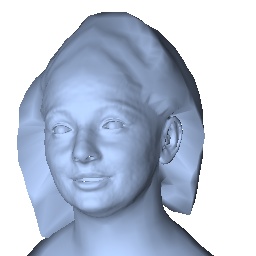} \hspace{\mrg} & 
            \includegraphics[width=\wid]{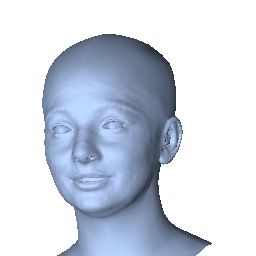} \hspace{\mrg} & 
           \includegraphics[width=\wid]{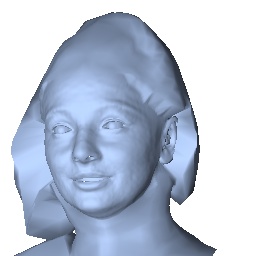} \hspace{\mrg} & 
           \includegraphics[width=\wid]{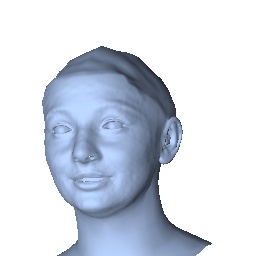} \hspace{\mrg} & 
           \includegraphics[width=\wid]{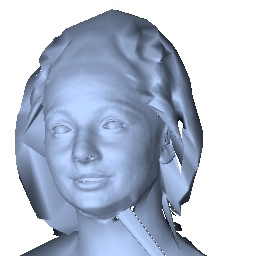} \hspace{\mrg} & 
             \includegraphics[width=\wid]{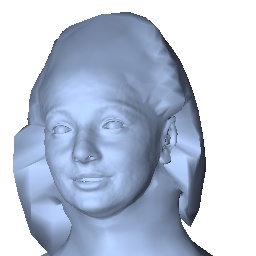} \hspace{\mrg} \vspace{5pt} \\
             
            \textbf{Input} \hspace{\mrg} &
              \textbf{Full} \hspace{\mrg}  &
              \textbf{w/o $\Delta\hat\v$} \hspace{\mrg}   &
                \textbf{w/o $\vec{n}$} \hspace{\mrg} &
            \textbf{w/o $\mathcal{L}_\text{occ}$} \hspace{\mrg} &
            \textbf{w/o $\mathcal{L}_\text{lap}$} \hspace{\mrg} &
            \textbf{w/o $\mathcal{L}_\text{chm}$} \hspace{\mrg}
        \end{tabular}
    \caption{Ablation study. We qualitatively evaluate the individual components of our \emph{full} model. w/o $\Delta \hat\v$: without the per-vertex displacements, we obtain a significantly worse render quality. w/o $\vec{n}$: when we apply per-vertex deformations instead of per-vertex displacements (i.e., deformations alongside the normals), we obtain noisy reconstructions in neck area and worse renders. w/o $\mathcal{L}_\text{occ}$: without silhouette-based losses, our model fails to learn proper reconstructions. w/o $\mathcal{L}_\text{lap}$: Laplacian regularization smooths the reconstructions.  w/o $\mathcal{L}_\text{chm}$: chamfer loss allows us to constrain the displaced vertices to lie inside the scene boundaries, which positively affects the smoothness of the visible part of the reconstruction.}
    \label{fig:ablation}
    \vspace{-0.5cm}
\end{figure}

\begin{figure}[h!]
    \begin{subfigure}{.48\textwidth}
    \setlength{\wid}{0.3\textwidth}
        \centering    

    \begin{tabular}{ccc}
        \hspace{-0.1cm}\includegraphics[width=\wid]{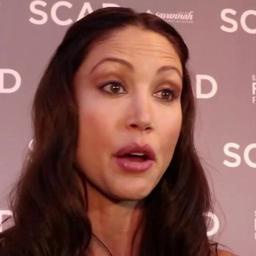} &
        \hspace{-0.1cm} \includegraphics[width=\wid]{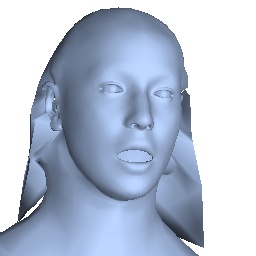} & \hspace{-0.1cm}
        \includegraphics[width=\wid]{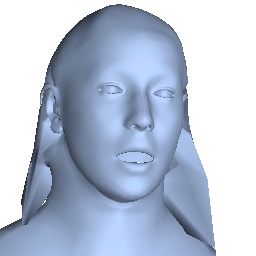}
        \\
        \hspace{-0.2cm}\textbf{Image} &
        \hspace{-0.2cm} \textbf{ROME} &  
        \hspace{-0.2cm} \textbf{Distilled}
    
    \end{tabular}
    \caption{Linear model}
        \label{fig:linear_model}
    \end{subfigure}
    \begin{subfigure}{.48\textwidth}
    \centering    
    \setlength{\wid}{0.3\textwidth}
    \begin{tabular}{ccc}
        \includegraphics[width=\wid]{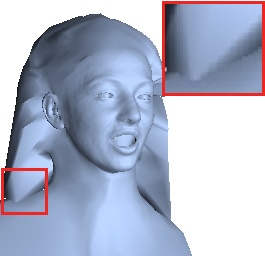} \hspace{-0.2cm} &
        \includegraphics[width=\wid]{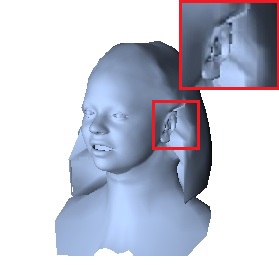} \hspace{-0.2cm} & 
        \includegraphics[width=\wid]{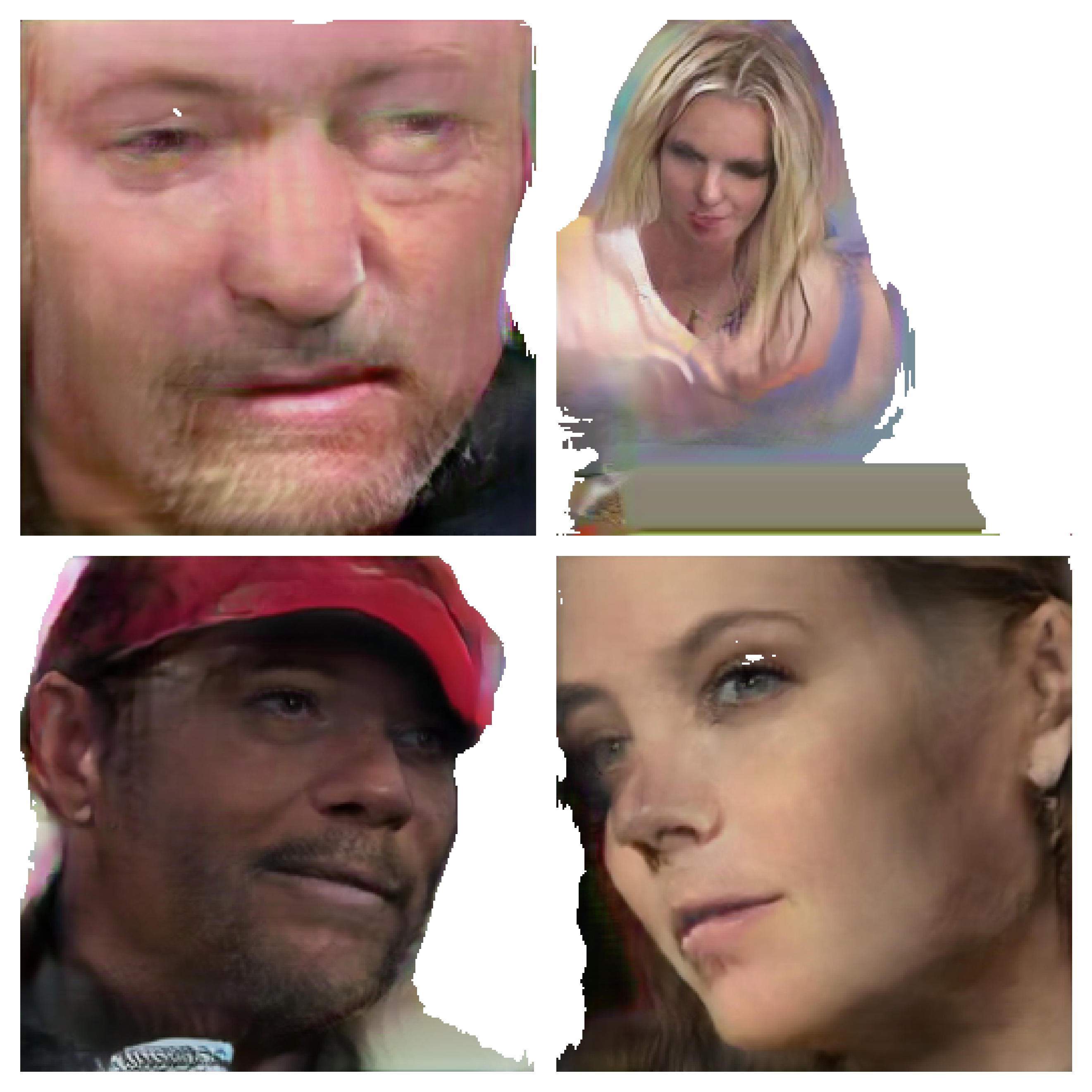}  \\
        \textbf{Long hair} & \textbf{Ear cover} &\textbf{Failed renders}
        \end{tabular}
    \caption{Limitations}
    \label{limitations}
    \end{subfigure}

    \caption{Linear model results and the examples of limitations. On the left, we show how reconstructions learned by our method, \textbf{ROME}, could be \textbf{distilled} using a linear parametric model. We are able to compress the learned offsets into a small basis, reducing the degrees of freedom by two orders of magnitude. We can then \textbf{distill} these offsets using a much faster regression network with a small gap in terms of the reconstruction quality. On the right, we highlight the main limitations of our method, which include the failure related to \textbf{long hair} modelling, caused by an incorrect topological prior, no \textbf{coverage of ears} and \textbf{unrealistic renders} under a significant change of scales.
    }
    \label{fig:linear_model_lim}
\end{figure}

As discussed above, we distill our ROME head reconstruction model into a linear parametric model. To do that, we set the number of basis vectors to $50$ for the hair and $10$ for the neck offsets and run low-rank Principle Component Analysis (PCA) to estimate them. The number of components is chosen to obtain a low enough approximation error. Interestingly, the offsets learned by our model can be compressed by almost two orders of magnitude in terms of degrees of freedom without any practical loss in quality~(\fig{linear_model_lim}\subref{fig:linear_model}), which suggests that the capacity of the offset generator is underused in our model. We combine estimated basis vectors with the original basis of the FLAME.

After that, we train feed-forward encoders that directly predict the coefficients of the two basis from the source image. The prediction is performed in two stages. First, face expression, pose and camera parameters are predicted with a MobileNetV2~\cite{Sandler2018MobileNetV2IR} encoder. 
Then a slower ResNet-50 encoder~\cite{He2016DeepRL} is used to predict hair, neck and shape coefficients. The choice of architectures are motivated by the fact that in many practical scenarios only the first encoder needs to be invoked frequently (per-frame), while the second can run at much lower rate or even only at the model creation time.

\subsection{Ablation study}
We demonstrate results of ablation study at \fig{ablation}. As expected, predicting more  accurate geometry affect the renders (first row). Also, we verify the necessity of all terms of geometry loss.
We observe significant improvement in quality of renders with additional geometry (see Supp~\ref{tab:supp_vc2_abl}), which leads us to an optimistic conclusion that our learned coarse mesh may be integrated into other neural rendering systems~\cite{Wang2021OneShotFN} to improve their quality. Additionally, we observe that geometry losses allows to correctly model coarse details on the hair without noise and reconstruct the hair without sticking with neck. Similar artifacts are removed by adding shifts along the normals.

\subsection{Limitations}

In some cases, the proposed system often produces somewhat oversmoothed geometry without person-specific attributes. Most of those attributes simply cannot be presented with the limited set of vertices. Our preliminary experiments with subdivision during training did not help to alleviate this problem. It may be interesting to study this problem in a few-shot scenario (given different views) or to use images and proxy geometry in a higher resolution.
Another frequent geometry artifact is the roughness of the clothing approximation.  As we know, modeling any clothes without 3D data is an extremely difficult task and the fact that our models is not able to do so from videos is not surprising. A principled solution to this problem will likely require learning from datasets that cover these areas (e.g., \ circular 3D scans of human heads)~\cite{Alldieck2019}.

Our current model is trained at roughly fixed scale, though explicit geometry modeling allows it to generalize to adjacent scale reasonably well. Still,  strong changes of scale lead to poor performance (\fig{linear_model_lim}\subref{limitations}). More examples are provided in the supplementary materials. Addressing this issue via mip-mapping and multi-scale GAN training techniques remains future work.

Lastly, our model can have artifacts with long hair (\fig{linear_model_lim}\subref{limitations}, left) or ears (\fig{linear_model_lim}\subref{limitations}, middle). Handling such cases gracefully are likely to require a departure from the predefined FLAME mesh connectivity to new person-specific mesh topology. Handling such issues using a limited set of pre-designed hair meshes is an interesting direction for future research.

\section{Summary}

We have presented ROME avatars: a system for creating realistic one-shot mesh-based human head models that can be animated and are compatible with FLAME head models. We compare our model with representative state-of-the-art models from different classes, and show that it is highly competitive both in terms of geometry estimation and the quality of rendering.

Crucially, our system can learn to model head geometry without direct supervision in the form of 3D scans. Despite that, we have observed it to achieve state-of-the-art results in head geometry recovery from a single photograph. At the same time, it also performs better than previous one-shot neural rendering approaches in the cross- and self-driving scenario. We have thus verified that the resulting geometry could be used to improve the rendering quality.

As neural rendering becomes more widespread within graphics systems, ROME avatars and similar systems can become directly applicable, while their one-shot capability and the simplicity of rigging derived from DECA and FLAME could become especially important in practical applications.%

\clearpage

\section*{Acknowledgements}

We sincerely thank Eduard Ramon for providing us the one-shot H3D-Net reconstructions. We also thank Arsenii Ashukha for their comments and suggestions regarding the text contents and clarity, as well as Julia Churkina for helping us with proof-reading. The computational resources for this work were mainly provided by Samsung ML Platform.

\bibliographystyle{unsrt}
\bibliography{refs}

\newpage
\appendix
\def\x{\mathbf{x}}
\def\X{\mathbf{X}}
\def\s{\mathbf{s}}
\def\v{\mathbf{v}}
\def\V{\mathbf{V}}
\def\e{\mathbf{e}}
\def\E{\mathbf{E}}
\def\T{\mathbf{T}}
\def\Z{\mathbf{Z}}
\def\z{\mathbf{z}}
\def\o{\mathbf{o}}
\def\p{\mathbf{p}}
\section{Supplementary material}

\subsection{Implementation details}

\paragraph{Photometric training objectives.}

During training, we use the photometric loss $\mathcal{L}_\text{photo}$ to aid in learning the geometry, as well as to train the rendering. Our photometric loss is the combination of the perceptual, the identity, the adversarial, and the segmentation losses:
\begin{equation}
    \mathcal{L}_\text{photo} = \lambda_\text{per} \mathcal{L}_\text{per} + \lambda_\text{idt} \mathcal{L}_\text{idt} + \lambda_\text{adv} \mathcal{L}_\text{adv} + \lambda_\text{seg} \mathcal{L}_\text{seg}.
\end{equation}

For the perceptual loss $\mathcal{L}_\text{per}$, we use a weighted combination of distances between features of predicted and target images. These features are taken from two pre-trained convolutional neural networks. We use features from a VGG19~\cite{Simonyan2015VeryDC} network pre-trained on ImageNet~\cite{Russakovsky2015ImageNetLS} to match the general content and features from a VGG16-based gaze detection network~\cite{Cortacero2019RTBENEAD} to match the gaze direction. This loss, therefore, can be expressed as follows:
\begin{equation}
    \mathcal{L}_\text{per} = \mathcal{L}_\text{vgg} + \mathcal{L}_\text{gaze}.
\end{equation}

In both of these losses, we measure the L1 distance between conv1\_1, conv2\_1, conv3\_1, conv4\_1, and conv5\_1 features after ReLU activations. We then sum these distances with the following weights: $\frac{1}{32}, \frac{1}{16}, \frac{1}{8}, \frac{1}{4},$ and $1$ to obtain the loss value.

In the gaze detection network, we independently process both eyes and average the losses corresponding to each one of them. Before feature extraction, we additionally perform alignment of both eyes using the keypoints extracted from the image via a differentiable interpolation. For more details, please refer to~\cite{Cortacero2019RTBENEAD}.

The face identity loss $\mathcal{L}_\text{idt}$ is the cosine distance between the embeddings of a pre-trained face recognition network~\cite{Cao2018VGGFace2AD}, evaluated for the predicted and the target images. We use this loss to better preserve the person's identity in the renders. We calculate the cosine distance as minus the cosine similarity.

To calculate the adversarial loss $\mathcal{L}_\text{adv}$ we jointly train a discriminator network alongside our reconstruction and rendering networks. We use a weighted combination of the hinge-loss discriminator loss as well as the feature-matching loss~\cite{Wang2018HighResolutionIS}. We additionally apply spectral normalization~\cite{Miyato2018SpectralNF} to the discriminator network. The configuration of the adversarial training closely follows the related works on image-to-image translation~\cite{Wang2019FewshotVS}. In particular, we use the PatchGAN~\cite{Isola2017ImagetoImageTW} architecture for the discriminator.

Lastly, we use Dice loss to match the predicted segmentation $\hat\s_t$ to the ground-truth $\s_t$:
\begin{equation}
    \mathcal{L}_\text{seg} = 1 - 2 \frac{\hat\s_t \cdot \s_t}{ \|\hat\s_t\|^2_2 + \|\s_t\|^2_2 },
\end{equation}
where $\cdot$ denotes a scalar product. The segmentation loss is calculated using each batch element separately and then averaged across the batch.

\paragraph{Architectures of the neural networks.} We use group normalization~\cite{Wu2018GroupN} paired with weight standardization~\cite{Qiao2019WeightS} in all networks to facilitate training with smaller batch sizes. Empirically we found this combination to perform better than standard instance normalization~\cite{Ulyanov2016InstanceNT}. 

In the autoencoder $E_\text{tex}$, which encodes the input image $\x_s$ into the neural texture $\T_s$, we use pre-activation residual blocks~\cite{He2016IdentityMI}. We set the number of channels in the neural texture $\T_s$ to eight (we observed that values between $8$ and $16$ result in similar performance). We additionally align the input image using the transformation similar to the one used in the FFHQ dataset~\cite{Karras2019ASG}. We only modify the zoom-out factor to $1.25$ so that the aligned image contains more hair and upper-body regions.

For the networks $E_\text{img}$ and $E_\text{geom}$ we use a standard U-Net architecture. The network $E_\text{img}$ is U-Net, which predict an output image and the segmentation mask. Besides the neural texture, we additionally condition these two networks on the rendered mesh normals. Specifically, we process each input dimension of the normal vectors using $\{ \sin(kx) \}_{k=1}^K$ and $\{ \cos(kx) \}_{k=1}^K$ functions. In our experiments, we set $K=6$. We then concatenate the resulting encodings to the neural texture. We use max-pooling layers for downsampling and nearest neighbors upsampling in the network, which decodes an image, and average pooling with bilinear upsampling in the network, which decodes segmentations. Additionally, the segmentation U-Net network has two times fewer channels than the image network.

The architecture of $E_\text{geom}$ is the same as the image encoding U-Net, albeit with a different number of input and output channels. We set the dimensionality of a latent geometry map $\Z_t$, which is an output of $E_\text{geom}$, to 32. Additionally, we encode the $xyz$-texture using harmonic functions via the same process described before. We then concatenate the obtained embeddings to the initial $xyz$-texture. The resulting feature map has $3 + 3 \cdot 6 \cdot 2 = 39$ channels. In total, $E_\text{geom}$ has $8 + 39 = 47$ input channels, since we also concatenate the embeddings of the $xyz$-texture to the neural texture.

Finally, $G_\text{geom}$ consists of an MLP network which we apply separately for each vertex to predict its offsets. To obtain its inputs, we first resample the latent geometry map $\Z_t$ using an irregular grid specified by the texture coordinates $w$. The obtained features for each vertex are denoted as $\z_t^w$. These features are concatenated to the harmonic embeddings of $w$ in the same way as the $xyz$-texture. Specifically, each vector $w$ has two dimensions, therefore we encode it into $2 + 2 \cdot 6 \cdot 2 = 26$ features, and then concatenate with $32$ channels of $\z_t^w$ to obtain a vector with the dimensionality of $58$ --- the number of $G_\text{geom}$ input dimensions.

\paragraph{Training details.} We use the ADAM optimizer to train all networks in our model. We set the learning rate to $1 \cdot 10^{-4}$ for all the networks except for the discriminator. For it, we set the learning rate to $4 \cdot 10^{-4}$. We also set $\beta_1 = 0$, and $\beta_2 = 0.999$. We train using eight P40 NVIDIA GPUs to facilitate the batch size of 32. We observe that decreasing the batch size leads to visible degradation of both rendering and reconstructions, but we did no ablations to measure this effect.

\subsection{Evaluation}

\begin{figure*}[h!]
    \centering    
    \setlength{\wid}{0.154\textwidth}
    \setlength{\mrgone}{-0.1cm}
    \begin{tabular}{cccccc}
        \includegraphics[width=\wid]{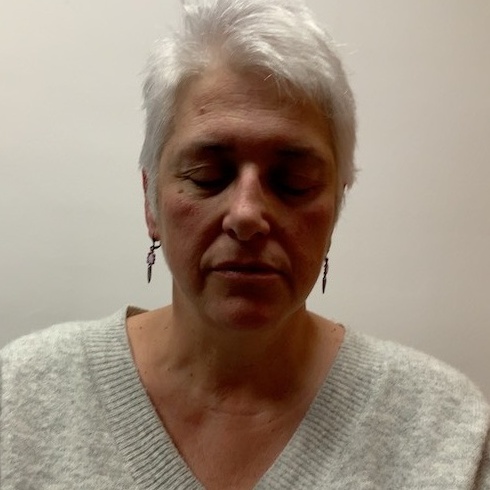} & \hspace{\mrgone}
        \includegraphics[width=\wid]{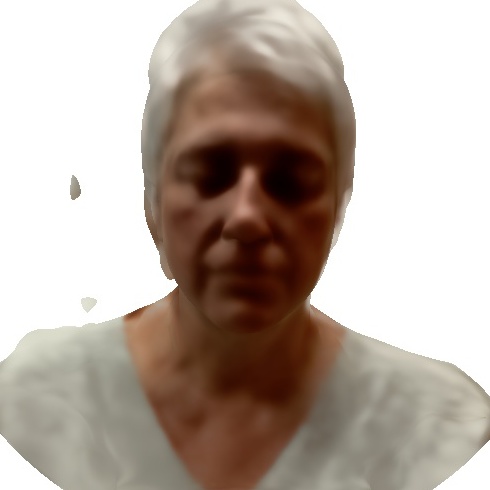} & \hspace{\mrgone}
        \includegraphics[width=\wid]{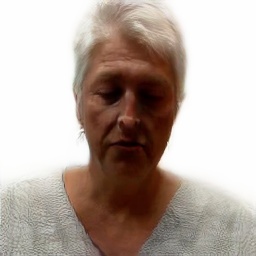} & \hspace{\mrgone}
        \includegraphics[width=\wid]{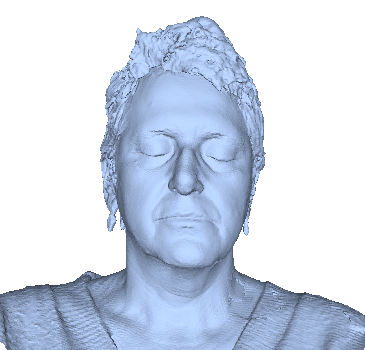} & \hspace{\mrgone}
        \includegraphics[width=\wid]{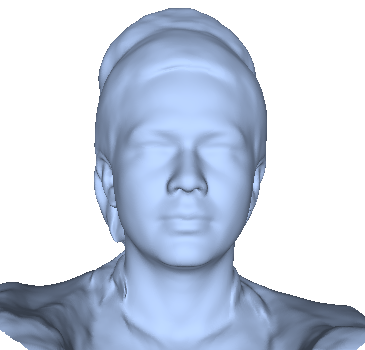} & \hspace{\mrgone}
        \includegraphics[width=\wid]{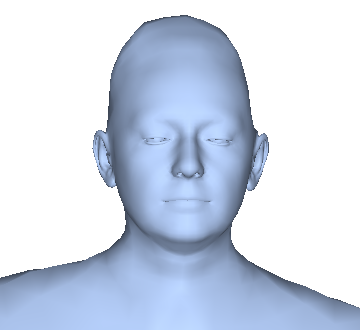} \\ %
        \includegraphics[width=\wid]{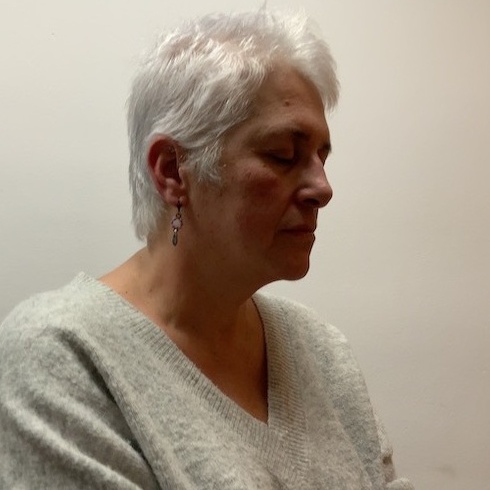} & \hspace{\mrgone}
        \includegraphics[width=\wid]{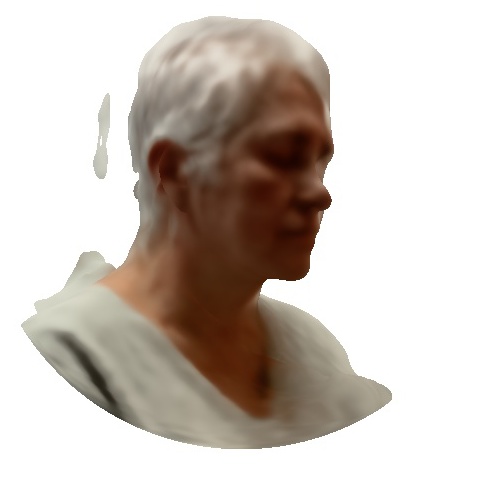} & \hspace{\mrgone}
        \includegraphics[width=\wid]{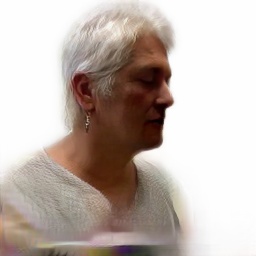} & \hspace{\mrgone}
        \includegraphics[width=\wid]{figures_suppmat_eccv/h3dnet/crop_gt.png} & \hspace{\mrgone}
        \includegraphics[width=\wid]{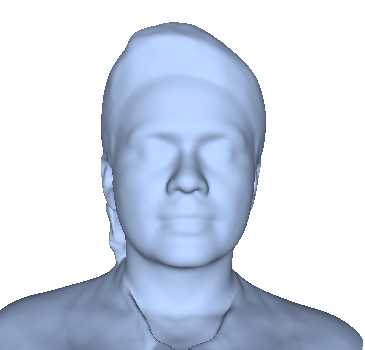} & \hspace{\mrgone}
        \includegraphics[width=\wid]{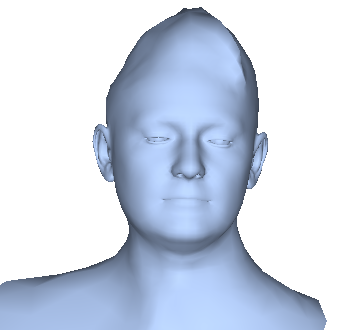} \\ %
        \includegraphics[width=\wid]{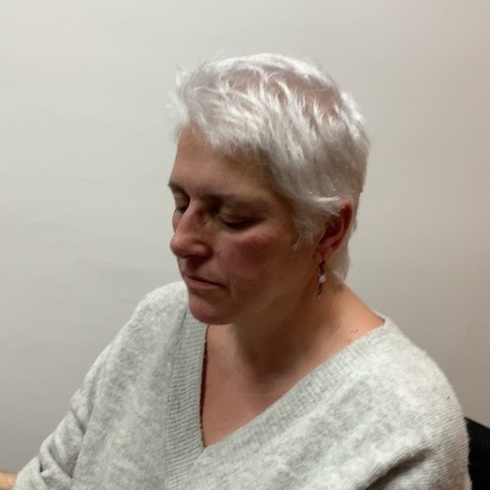} & \hspace{\mrgone}
        \includegraphics[width=\wid]{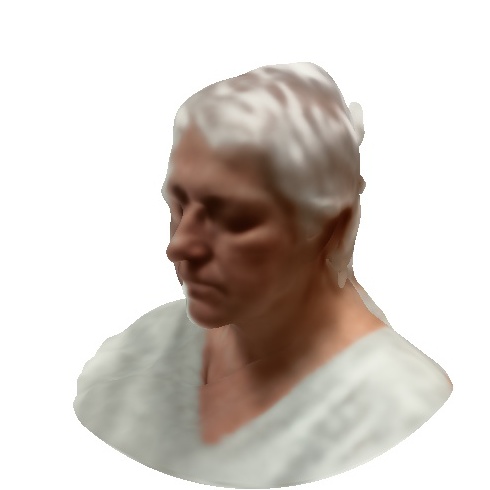} & \hspace{\mrgone}
        \includegraphics[width=\wid]{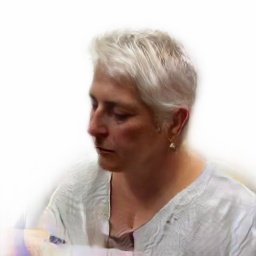} & \hspace{\mrgone}
        \includegraphics[width=\wid]{figures_suppmat_eccv/h3dnet/crop_gt.png} & \hspace{\mrgone}
        \includegraphics[width=\wid]{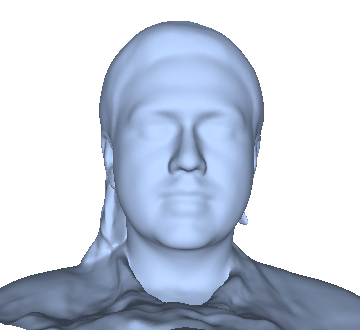} & \hspace{\mrgone}
        \includegraphics[width=\wid]{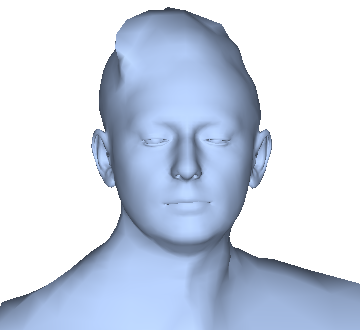} \\ %
        \includegraphics[width=\wid]{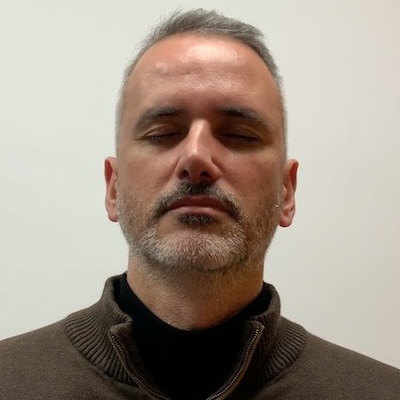} & \hspace{\mrgone}
        \includegraphics[width=\wid]{figures_suppmat/h3ds/2_1_gt.jpg} & \hspace{\mrgone}
        \includegraphics[width=\wid]{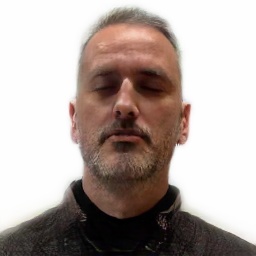} & \hspace{\mrgone}
        \includegraphics[width=\wid]{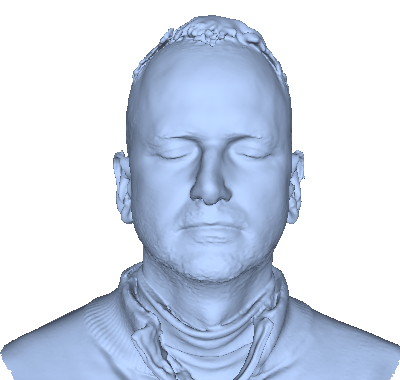} & \hspace{\mrgone}
        \includegraphics[width=\wid]{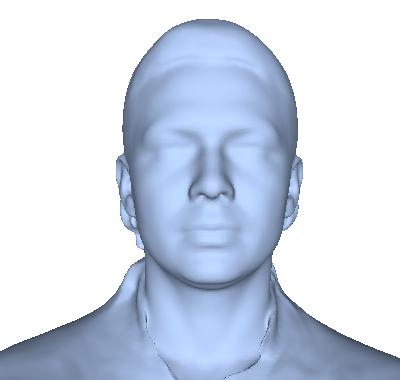} & \hspace{\mrgone}
        \includegraphics[width=\wid]{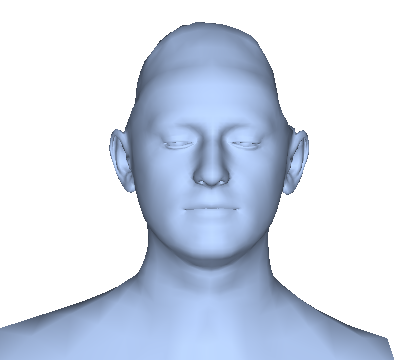} \\ %
        \includegraphics[width=\wid]{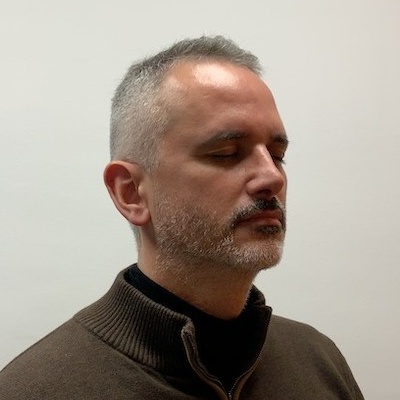} & \hspace{\mrgone}
        \includegraphics[width=\wid]{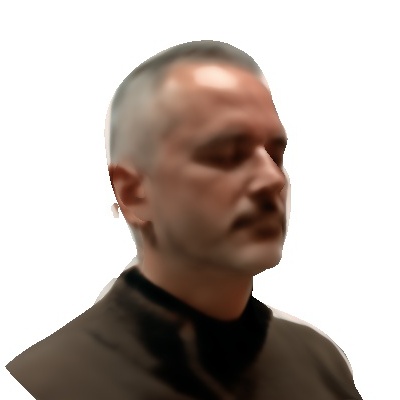} & \hspace{\mrgone}
        \includegraphics[width=\wid]{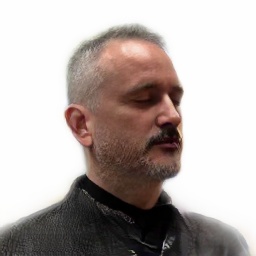} & \hspace{\mrgone}
        \includegraphics[width=\wid]{figures_suppmat_eccv/h3dnet/crop_gt_.png} & \hspace{\mrgone}
        \includegraphics[width=\wid]{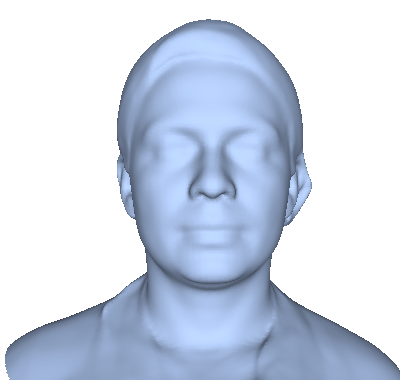} & \hspace{\mrgone}
        \includegraphics[width=\wid]{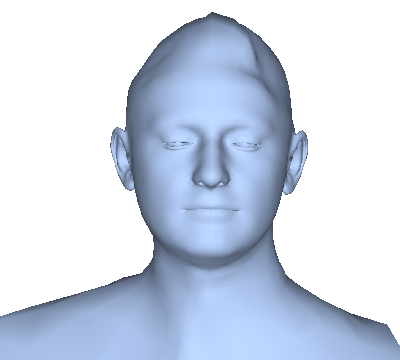} \\ %
        \includegraphics[width=\wid]{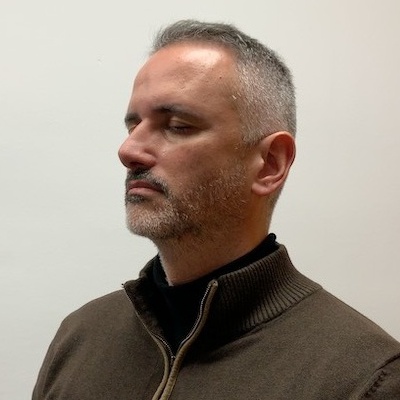} & \hspace{\mrgone}
        \includegraphics[width=\wid]{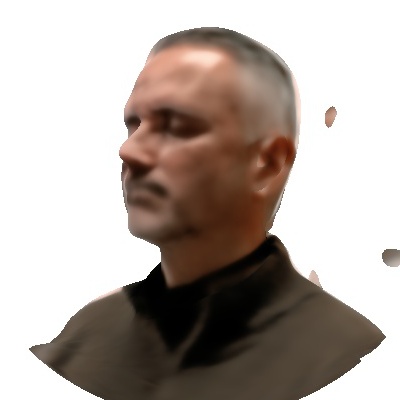} & \hspace{\mrgone}
        \includegraphics[width=\wid]{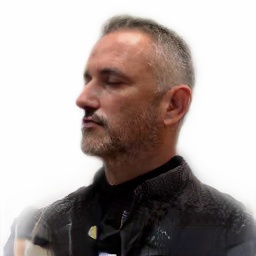} & \hspace{\mrgone}
        \includegraphics[width=\wid]{figures_suppmat_eccv/h3dnet/crop_gt_.png} & \hspace{\mrgone}
        \includegraphics[width=\wid]{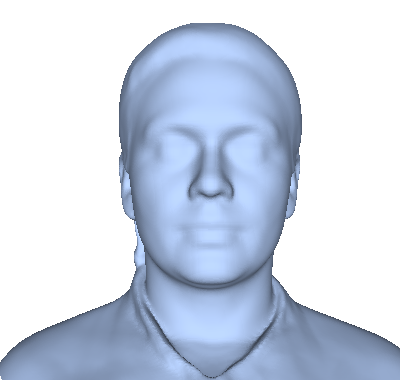} & \hspace{\mrgone}
        \includegraphics[width=\wid]{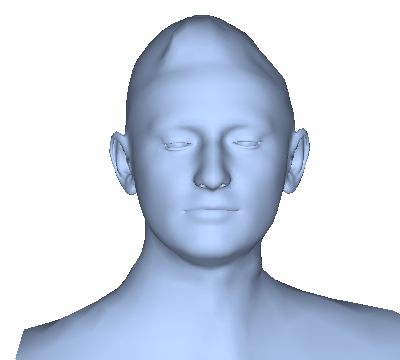} \\ %
        \includegraphics[width=\wid]{figures_suppmat/h3ds/2_1_gt.jpg} & \hspace{\mrgone}
        \includegraphics[width=\wid]{figures_suppmat/h3ds/2_1_gt.jpg} & \hspace{\mrgone}
        \includegraphics[width=\wid]{figures_suppmat_eccv/h3dnet/1.jpg} & \hspace{\mrgone}
        \includegraphics[width=\wid]{figures_suppmat_eccv/h3dnet/crop_gt_.png} & \hspace{\mrgone}
        \includegraphics[width=\wid]{figures_suppmat_eccv/h3dnet/crop_h_1_.png} & \hspace{\mrgone}
        \includegraphics[width=\wid]{figures_suppmat_eccv/h3dnet/crop_o_1_.png} \\ %
        \includegraphics[width=\wid]{figures_suppmat/h3ds/2_15_gt.jpg} & \hspace{\mrgone}
        \includegraphics[width=\wid]{figures_suppmat/h3ds/2_15_h.jpg} & \hspace{\mrgone}
        \includegraphics[width=\wid]{figures_suppmat_eccv/h3dnet/15.jpg} & \hspace{\mrgone}
        \includegraphics[width=\wid]{figures_suppmat_eccv/h3dnet/crop_gt_.png} & \hspace{\mrgone}
        \includegraphics[width=\wid]{figures_suppmat_eccv/h3dnet/crop_h_15_.png} & \hspace{\mrgone}
        \includegraphics[width=\wid]{figures_suppmat_eccv/h3dnet/crop_o_15_.png} \\ %
        \includegraphics[width=\wid]{figures_suppmat/h3ds/2_62_gt.jpg} & \hspace{\mrgone}
        \includegraphics[width=\wid]{figures_suppmat/h3ds/2_62_h.jpg} & \hspace{\mrgone}
        \includegraphics[width=\wid]{figures_suppmat_eccv/h3dnet/62.jpg} & \hspace{\mrgone}
        \includegraphics[width=\wid]{figures_suppmat_eccv/h3dnet/crop_gt_.png} & \hspace{\mrgone}
        \includegraphics[width=\wid]{figures_suppmat_eccv/h3dnet/crop_h_62_.png} & \hspace{\mrgone}
        \includegraphics[width=\wid]{figures_suppmat_eccv/h3dnet/crop_o_62_.png} \\ %
        \textbf{Source}&\hspace{\mrgone} \textbf{H3D-Net} &\hspace{\mrgone} \textbf{ROME}&\hspace{\mrgone} \textbf{Target}  &\hspace{\mrgone} \textbf{H3D-Net}  &\hspace{\mrgone} \textbf{ROME} 
    \end{tabular}
    \vspace{-0.25cm}
    \caption{Extended qualitative comparison on the H3DS dataset. We compare 3D reconstructions and renders obtained using a single \textbf{source} image.%
    }\label{fig:supp_h3ds_comp}
\end{figure*}

\begin{figure*}[h!]
    \centering    
    \setlength{\wid}{0.120\textwidth}
    \def\bmarg{0.12cm}
    \def\sep{-0.2cm}
    \def\halfbigsep{0.2cm}
    \begin{tabular}{cccc cccc}
    
        \includegraphics[align=c,bmargin=\bmarg,width=\wid]{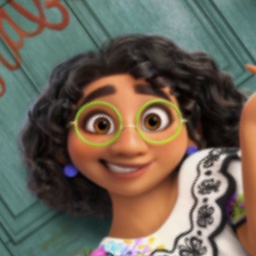} & \hspace{\sep}
        \includegraphics[align=c,width=\wid]{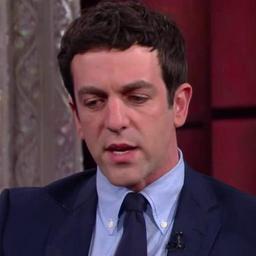} & \hspace{\sep}
        \includegraphics[align=c,width=\wid]{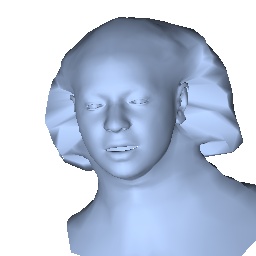} & \hspace{\sep}
        \includegraphics[align=c,width=\wid]{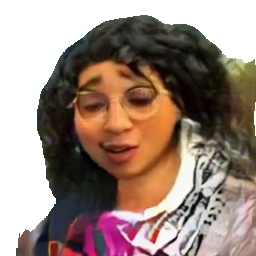} \hspace{\sep}
        \hspace{\halfbigsep} & \hspace{\halfbigsep}
        \includegraphics[align=c,width=\wid]{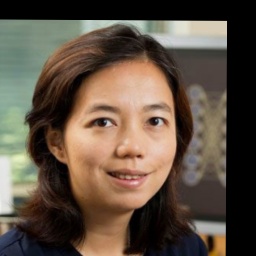} & \hspace{\sep}
        \includegraphics[align=c,width=\wid]{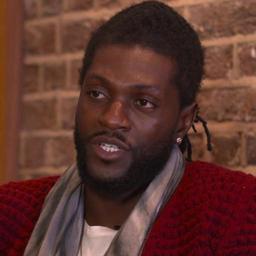} & \hspace{\sep}
        \includegraphics[align=c,width=\wid]{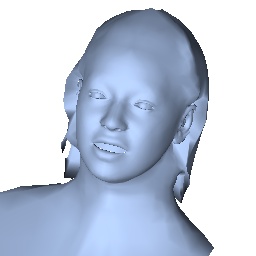} & \hspace{\sep}
        \includegraphics[align=c,width=\wid]{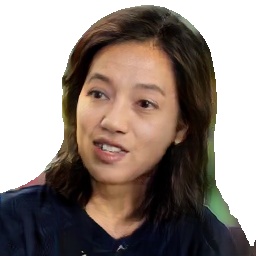} \\
  
         \includegraphics[align=c,bmargin=\bmarg,width=\wid]{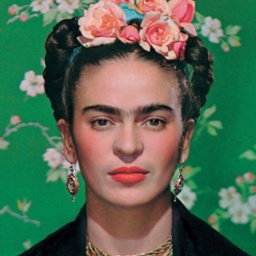} & \hspace{\sep}
        \includegraphics[align=c,width=\wid]{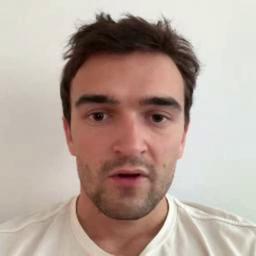} & \hspace{\sep}
        \includegraphics[align=c,width=\wid]{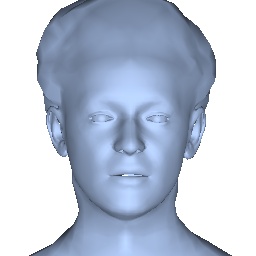} & \hspace{\sep}
        \includegraphics[align=c,width=\wid]{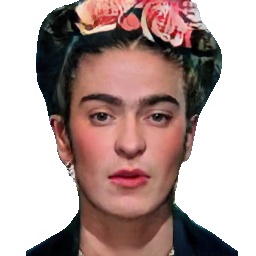} \hspace{\sep}
        \hspace{\halfbigsep} & \hspace{\halfbigsep}
        \includegraphics[align=c,width=\wid]{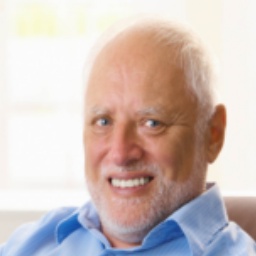} & \hspace{\sep}
        \includegraphics[align=c,width=\wid]{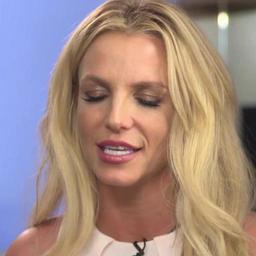} & \hspace{\sep}
        \includegraphics[align=c,width=\wid]{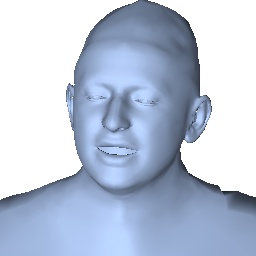} & \hspace{\sep}
        \includegraphics[align=c,width=\wid]{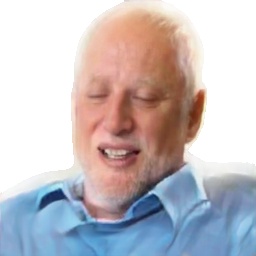} \\      
        
        \includegraphics[align=c,bmargin=\bmarg,width=\wid]{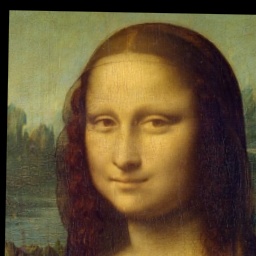} & \hspace{\sep}
        \includegraphics[align=c,width=\wid]{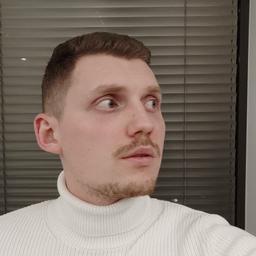} & \hspace{\sep}
        \includegraphics[align=c,width=\wid]{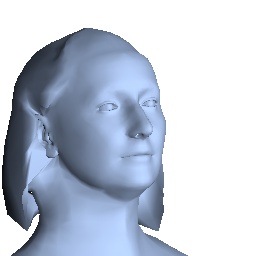} & \hspace{\sep}
        \includegraphics[align=c,width=\wid]{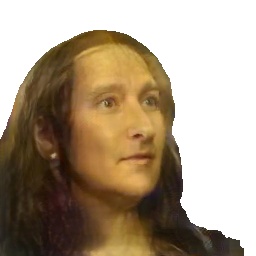} \hspace{\sep}
        \hspace{\halfbigsep} & \hspace{\halfbigsep}
        \includegraphics[align=c,width=\wid]{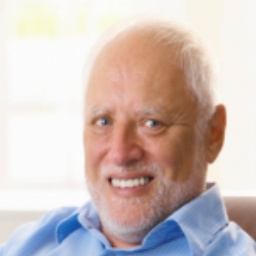} & \hspace{\sep}
        \includegraphics[align=c,width=\wid]{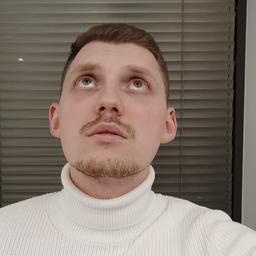} & \hspace{\sep}
        \includegraphics[align=c,width=\wid]{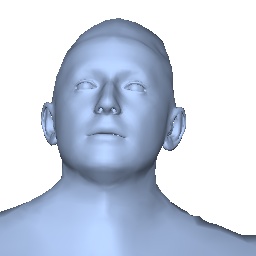} & \hspace{\sep}
        \includegraphics[align=c,width=\wid]{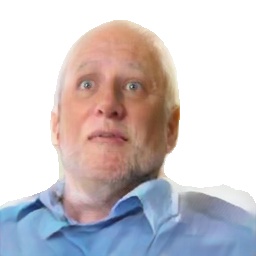} \\

        \includegraphics[align=c,bmargin=\bmarg,width=\wid]{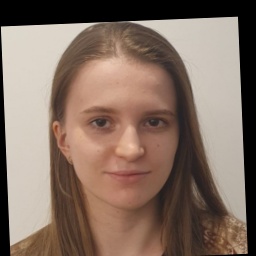} & \hspace{\sep}
        \includegraphics[align=c,width=\wid]{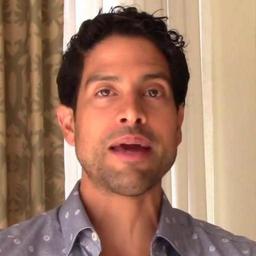} & \hspace{\sep}
        \includegraphics[align=c,width=\wid]{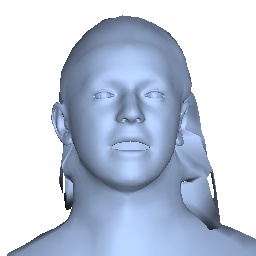} & \hspace{\sep}
        \includegraphics[align=c,width=\wid]{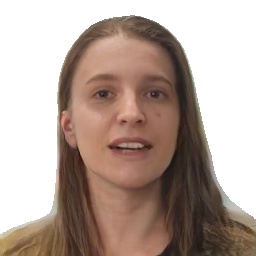} \hspace{\sep}
        \hspace{\halfbigsep} & \hspace{\halfbigsep}
        \includegraphics[align=c,width=\wid]{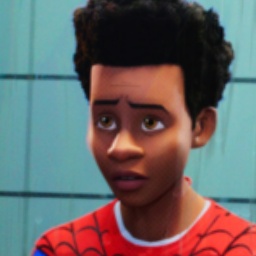} & \hspace{\sep}
        \includegraphics[align=c,width=\wid]{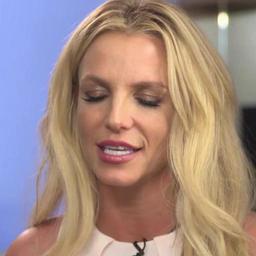} & \hspace{\sep}
        \includegraphics[align=c,width=\wid]{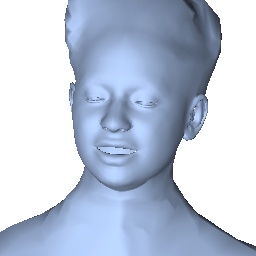} & \hspace{\sep}
        \includegraphics[align=c,width=\wid]{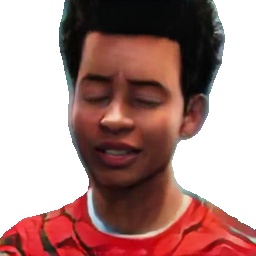} \\
        
        \includegraphics[align=c,bmargin=\bmarg,width=\wid]{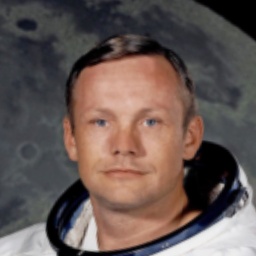} & \hspace{\sep}
        \includegraphics[align=c,width=\wid]{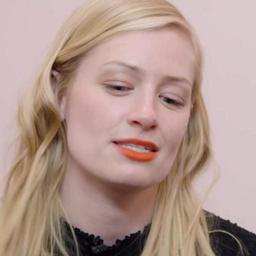} & \hspace{\sep}
        \includegraphics[align=c,width=\wid]{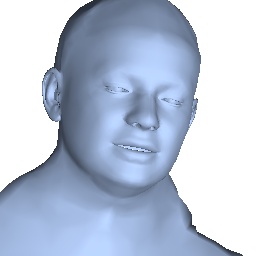} & \hspace{\sep}
        \includegraphics[align=c,width=\wid]{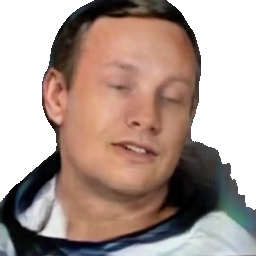} \hspace{\sep}
        \hspace{\halfbigsep} & \hspace{\halfbigsep}
        \includegraphics[align=c,width=\wid]{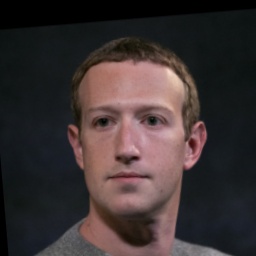} & \hspace{\sep}
        \includegraphics[align=c,width=\wid]{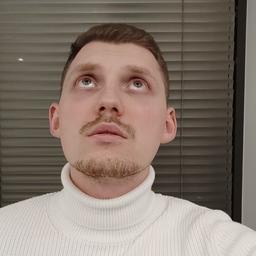} & \hspace{\sep}
        \includegraphics[align=c,width=\wid]{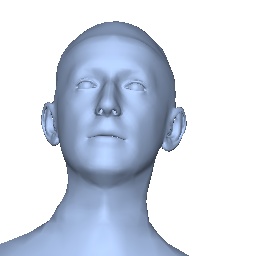} & \hspace{\sep}
        \includegraphics[align=c,width=\wid]{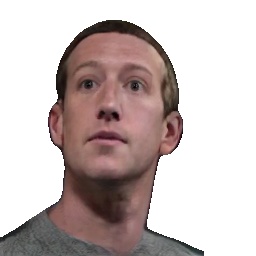} \\
        
        \includegraphics[align=c,bmargin=\bmarg,width=\wid]{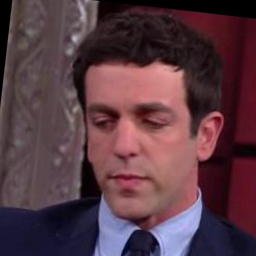} & \hspace{\sep}
        \includegraphics[align=c,width=\wid]{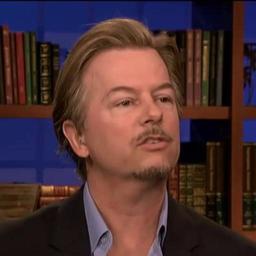} & \hspace{\sep}
        \includegraphics[align=c,width=\wid]{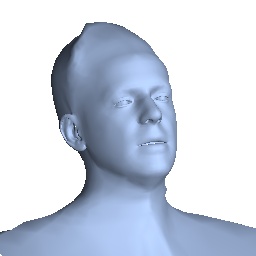} & \hspace{\sep}
        \includegraphics[align=c,width=\wid]{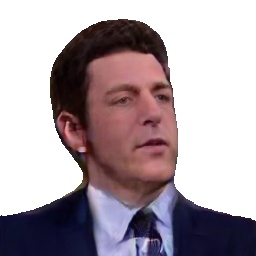} \hspace{\sep}
        \hspace{\halfbigsep} & \hspace{\halfbigsep}
        \includegraphics[align=c,width=\wid]{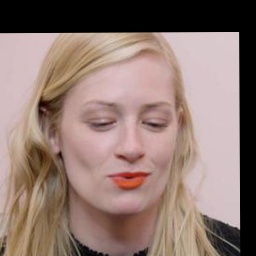} & \hspace{\sep}
        \includegraphics[align=c,width=\wid]{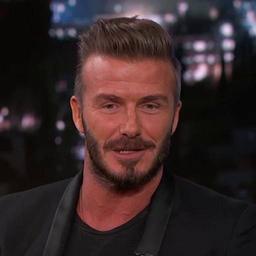} & \hspace{\sep}
        \includegraphics[align=c,width=\wid]{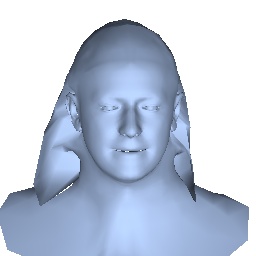} & \hspace{\sep}
        \includegraphics[align=c,width=\wid]{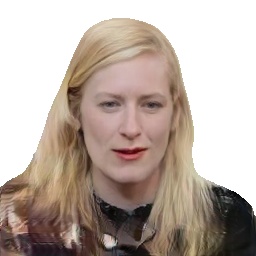} \\
        
        \includegraphics[align=c,width=\wid]{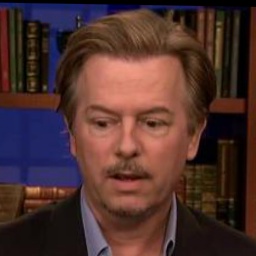} & \hspace{\sep}
        \includegraphics[align=c,width=\wid]{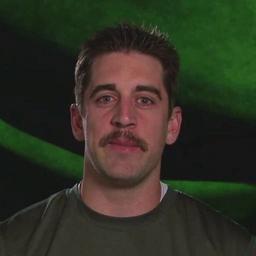} & \hspace{\sep}
        \includegraphics[align=c,width=\wid]{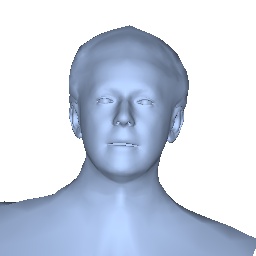} & \hspace{\sep}
        \includegraphics[align=c,width=\wid]{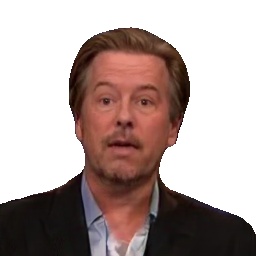}
        \hspace{\sep}
        \hspace{\halfbigsep} & \hspace{\halfbigsep}
        
          \includegraphics[align=c,width=\wid]{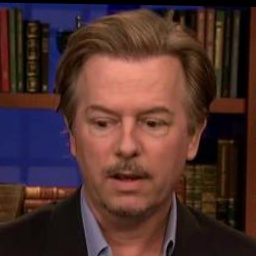} & \hspace{\sep}
        \includegraphics[align=c,width=\wid]{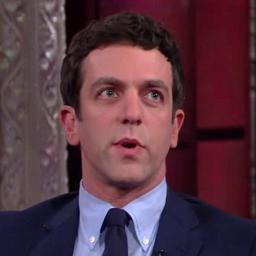} & \hspace{\sep}
        \includegraphics[align=c,width=\wid]{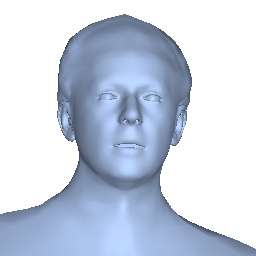} & \hspace{\sep}
        \includegraphics[align=c,width=\wid]{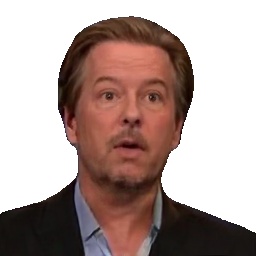}\\
        
      \includegraphics[align=c,bmargin=\bmarg,width=\wid]{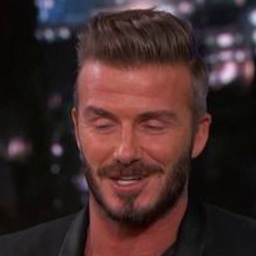} & \hspace{\sep}
        \includegraphics[align=c,width=\wid]{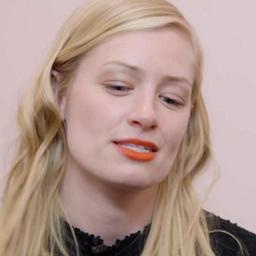} & \hspace{\sep}
        \includegraphics[align=c,width=\wid]{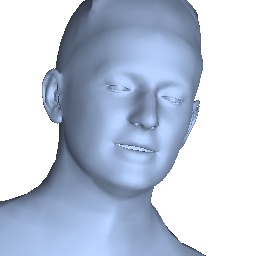} & \hspace{\sep}
        \includegraphics[align=c,width=\wid]{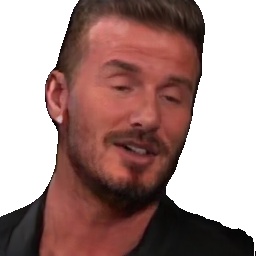}
         \hspace{\sep}
        \hspace{\halfbigsep} & \hspace{\halfbigsep}
        \includegraphics[align=c,bmargin=\bmarg,width=\wid]{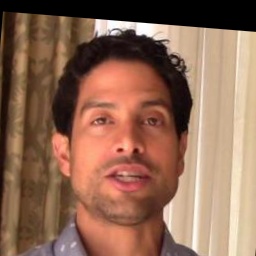} & \hspace{\sep}
        \includegraphics[align=c,width=\wid]{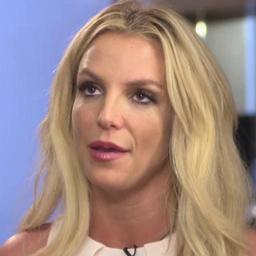} & \hspace{\sep}
        \includegraphics[align=c,width=\wid]{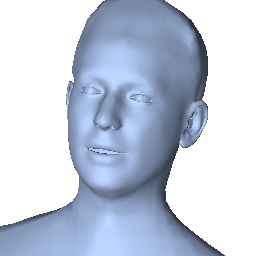} & \hspace{\sep}
        \includegraphics[align=c,width=\wid]{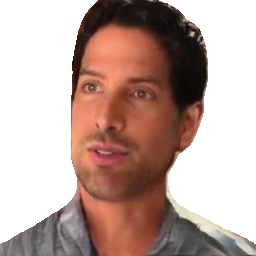} \\

        \textbf{Source} & \hspace{\sep} \textbf{Driver} & \multicolumn{2}{c}{\hspace{\sep} \textbf{ROME}} \hspace{\halfbigsep} & \hspace{\halfbigsep} \textbf{Source} & \hspace{\sep} \textbf{Driver} &  \multicolumn{2}{c}{\hspace{\sep} \textbf{ROME}} \\
    \end{tabular}
    \caption{Additional examples of mesh-based avatars creating using a single \textbf{source} image, and animated using the camera pose and the expression parameters estimated from the \textbf{driver} image. First five rows contains results for samples out of the train distribution.}\label{fig:supp_teaser}
\end{figure*}

We present additional results for the 3D reconstruction in both self-driving (reconstruction using frames from the same video), and cross-driving (reconstruction from a photo of one person and animation from a video of a different person) scenarios. In~\fig{supp_h3ds_comp}, we present more self-driving results on the H3DS dataset, as well as a side-by-side comparison with H3D-Net. In~\fig{supp_teaser}, we present cross-driving evaluation results using the hold-out samples from the VoxCeleb2 dataset, as well as paintings, the latter allowing us to evaluate the susceptibility of our approach to domain shifts.

\begin{figure}[ht!]
\begin{subfigure}{0.52\linewidth}
\centering
    \setlength{\wid}{1.8cm}
\centering
\begin{tabular}{ccc}
    \includegraphics[width=\wid]{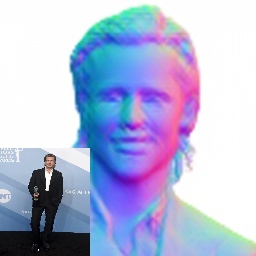}
    & \hspace{\mrg}
    \includegraphics[width=\wid]{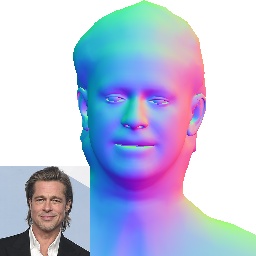}
    & \hspace{\mrg}
    \includegraphics[width=0.5\wid]{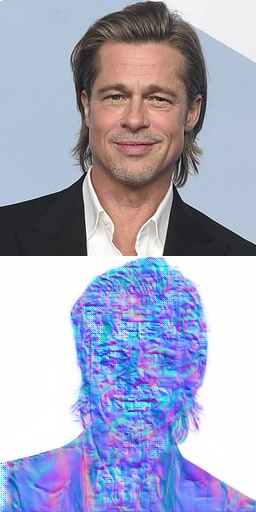} 
    \\
    \textbf{\scriptsize  \begin{tabular}{c}PIFuHD \\  Full body \end{tabular}  } 
    & \hspace{\mrg}
    \textbf{\scriptsize \begin{tabular}{c} Ours \\  Head \end{tabular} } 
    & \hspace{\mrg}
    \textbf{\scriptsize \begin{tabular}{c} PIFuHD \\ Head \end{tabular} }
\end{tabular}
\subcaption{Comparison with PIFuHD.}
\label{fig:ablation_pifu}

\end{subfigure}
\begin{subfigure}{0.44\linewidth}
    \setlength{\wid}{1.1cm}
    \centering
    \begin{tabular}{ccc}
        \includegraphics[width=\wid]{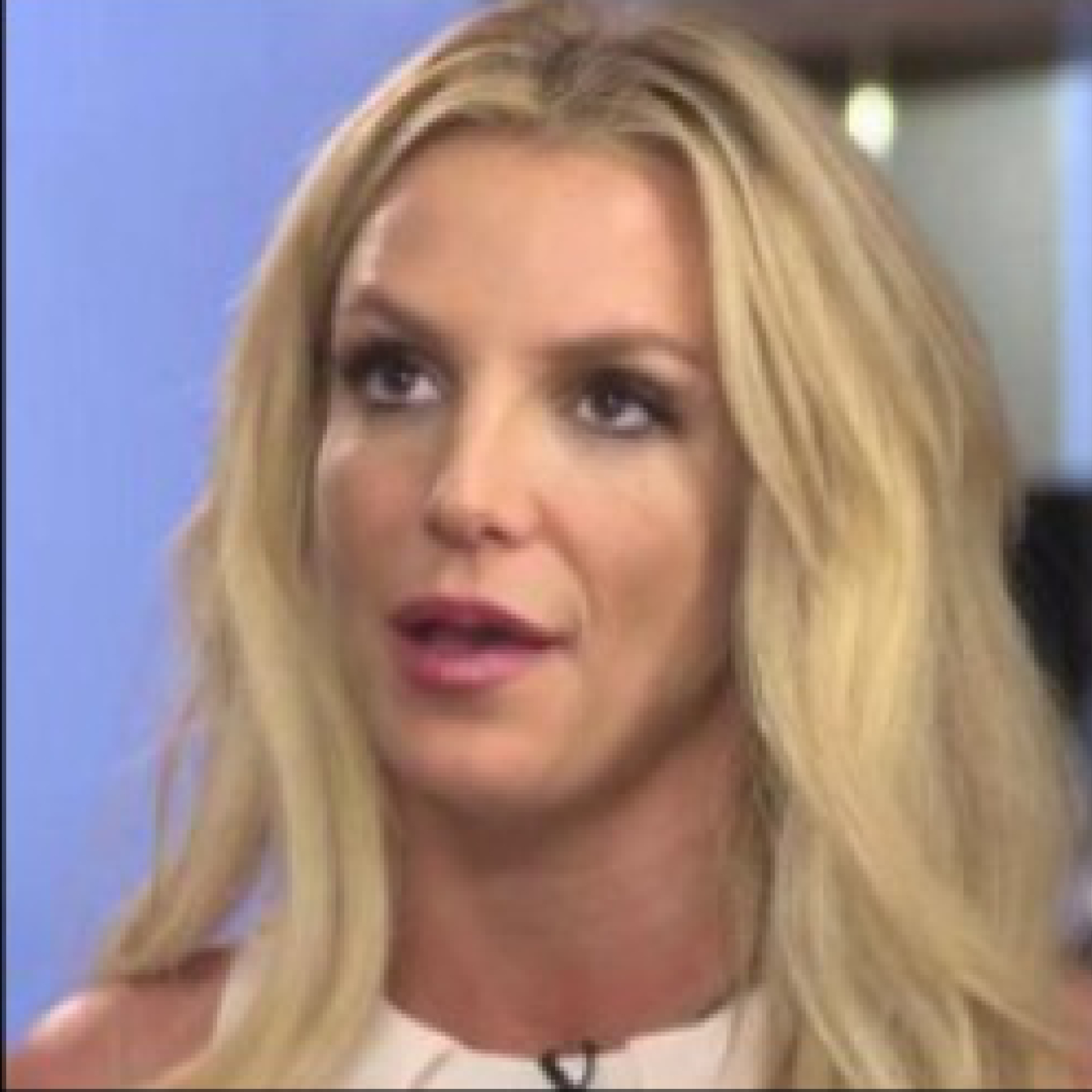} &
        \hspace{\mrg}
        \includegraphics[width=\wid]{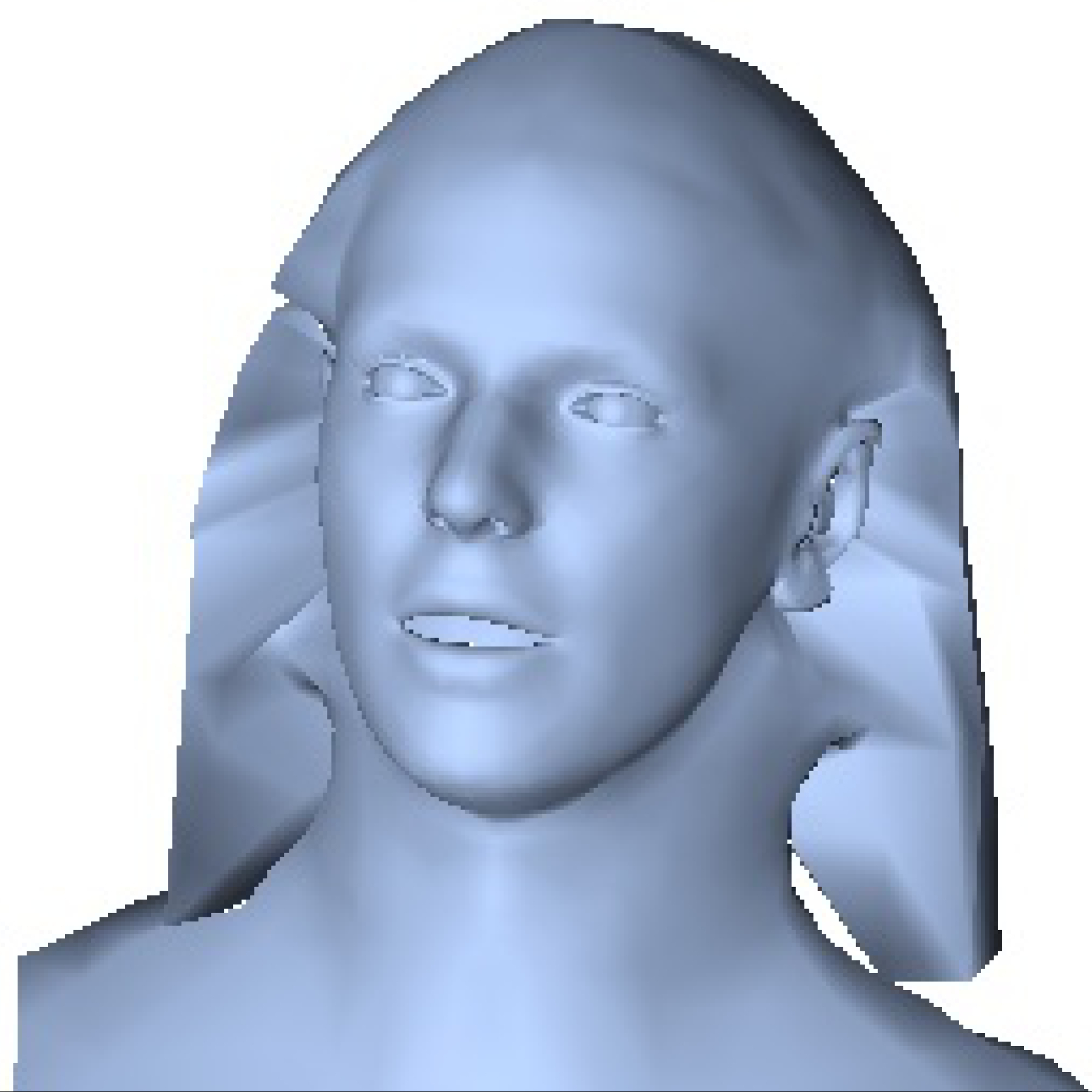} &
        \hspace{\mrg}\hspace{-0.1cm}
        \includegraphics[width=\wid]{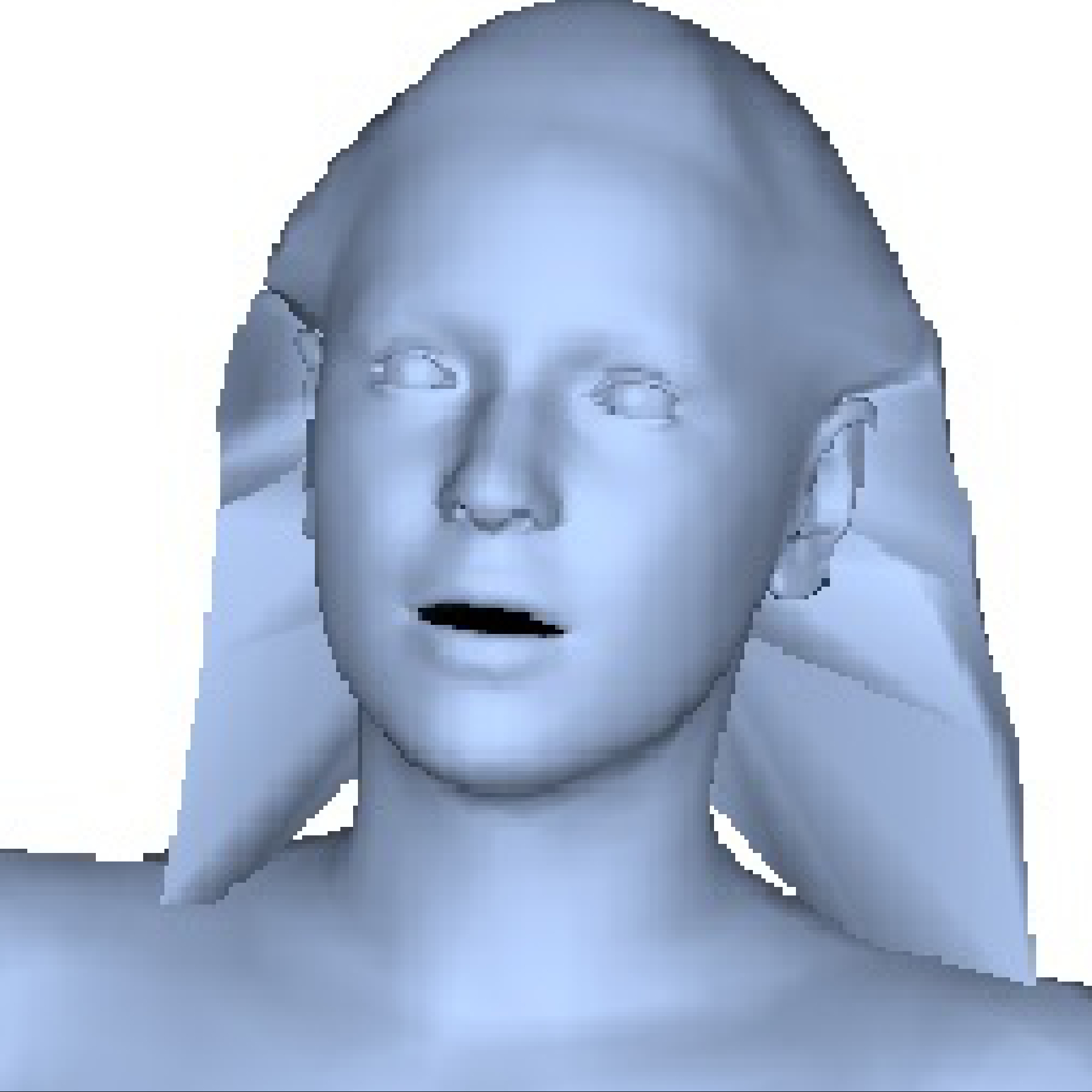}
        \\
        \includegraphics[width=\wid]{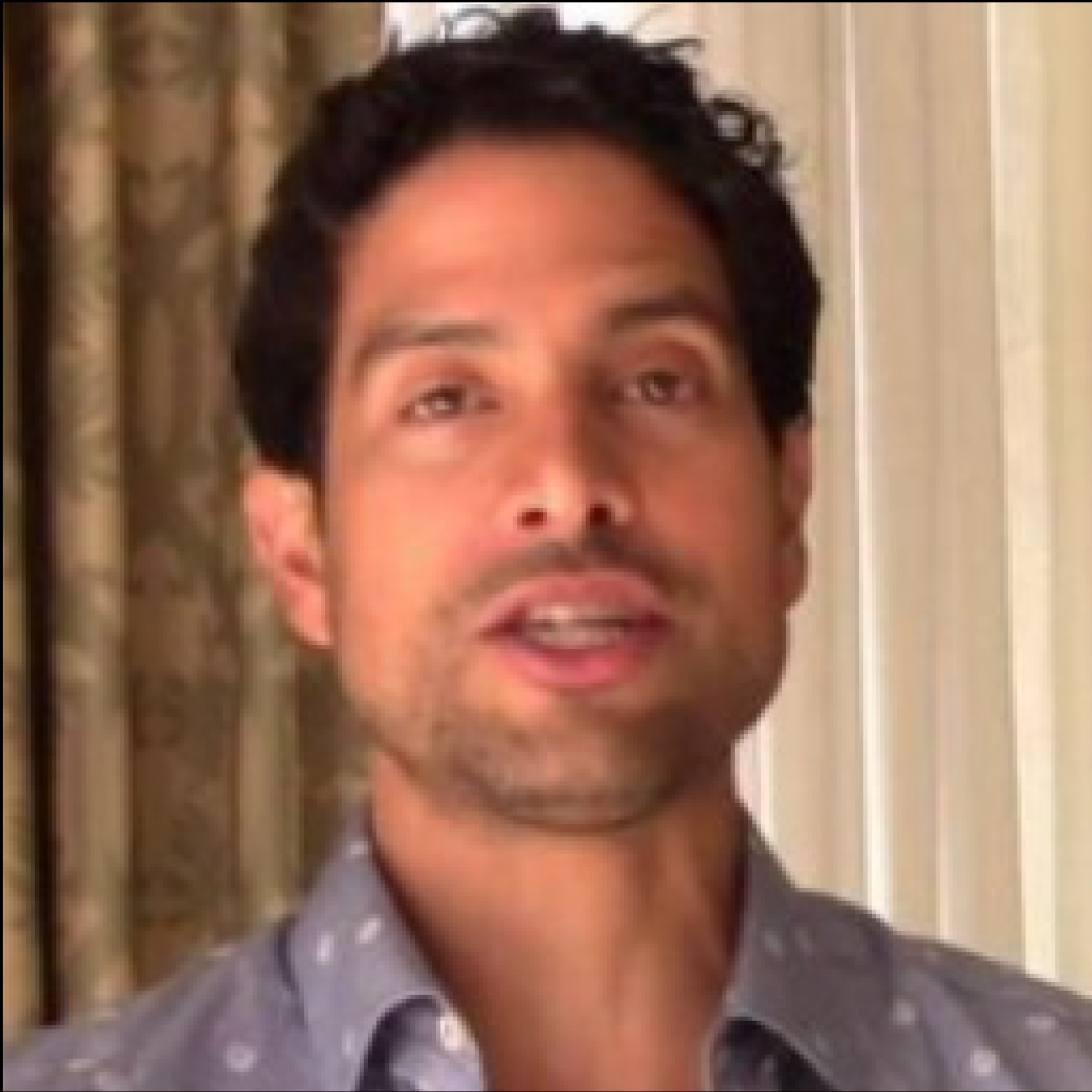} &
        \hspace{\mrg}
        \includegraphics[width=\wid]{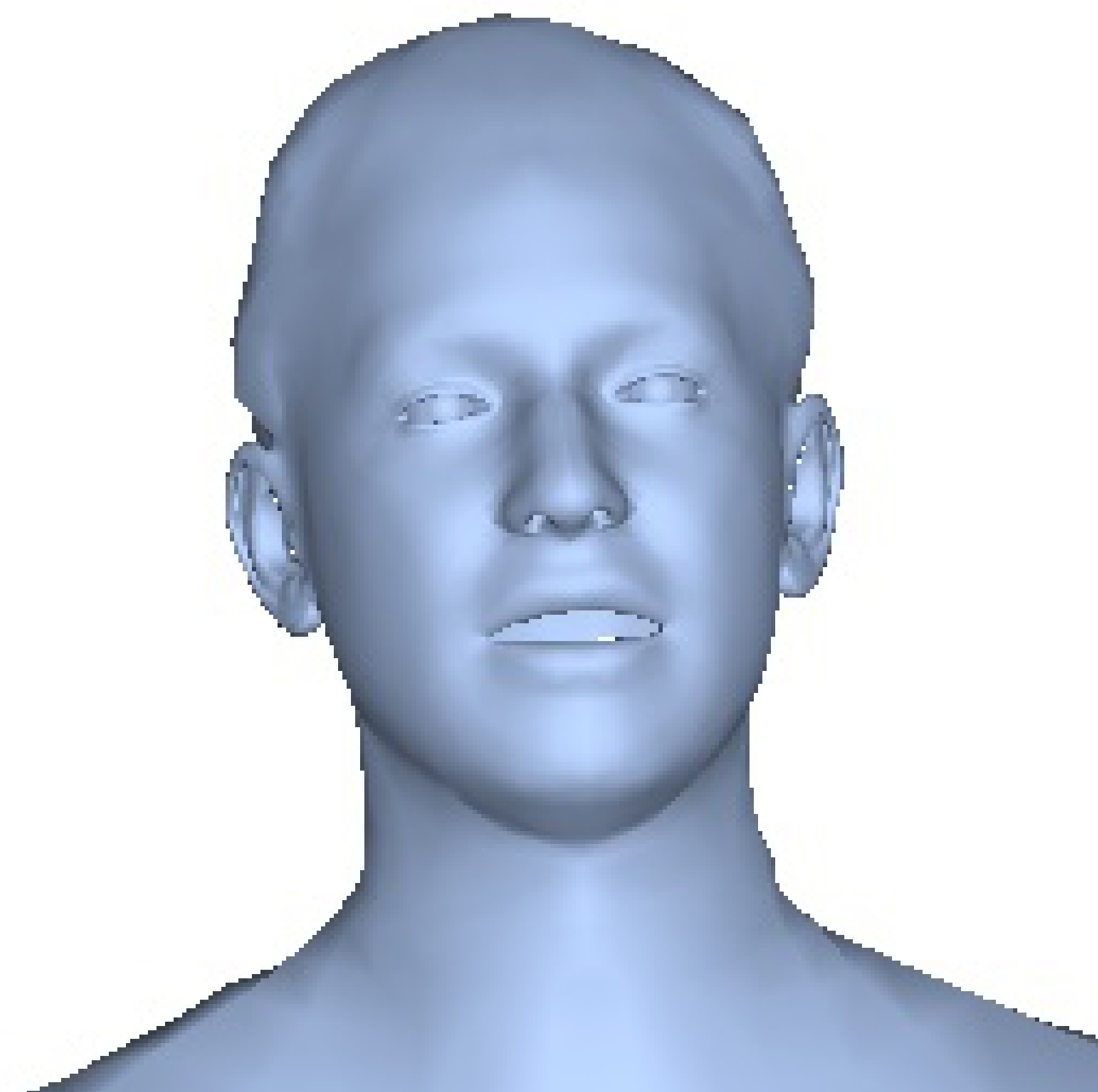} &
        \hspace{\mrg}\hspace{-0.1cm}
        \includegraphics[width=\wid]{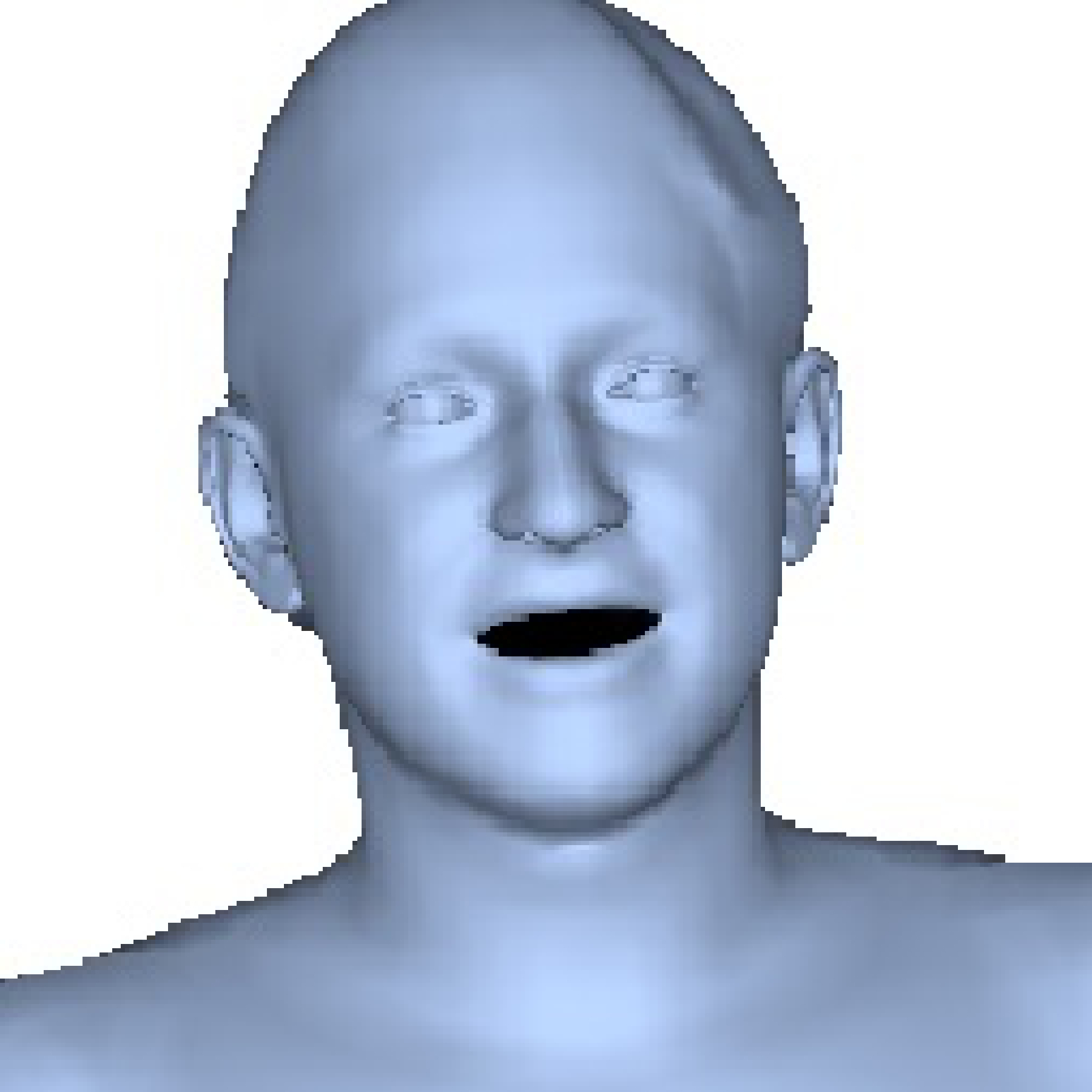}
        \\
        \textbf{\scriptsize \begin{tabular}{c} Source \\ Image \end{tabular}} &
        \hspace{\mrg}
        \textbf{\scriptsize \begin{tabular}{c} ROME \\ w/ DECA \end{tabular}} &
        \hspace{\mrg}\hspace{-0.1cm}
        \textbf{\scriptsize \begin{tabular}{c} ROME(L) \\ w/ PIXIE \end{tabular}}
    \end{tabular}
    \subcaption{ROME integrated with PIXIE.}
    \label{fig:rome_pixie}
\end{subfigure}
\end{figure}

Additionally, we show the results in Fig.~\ref{fig:ablation_pifu}. 
PIFuHD can only recover head geometry from the full-body images, which prevents us from doing a comparison with this method using existing head reconstruction and reenactment benchmarks.
While having more detailed reconstructions, PIFuHD requires training with 3D supervision and does not have animation capabilities. Currently, these limitations can be lifted only at the cost of lengthy multi-shot training per each avatar (IMAvatars), or by sacrificing some detalization of reconstructions and retaining both real-time and one-shot capabilities, which is done in our method.

To demonstrate the the ease of integration with existing SMPL-X based models we predict the vertices corresponded to hair using distilled version of the ROME with PIXIE~\cite{PIXIE:3DV:2021}. The resulted mesh contains the hair from ROME basis and shoulders from SMPL-X.

\begin{figure*}[h!]
    \centering    
    \setlength{\wid}{0.15\textwidth}
    \begin{tabular}{cccccc}

        \includegraphics[align=c,bmargin=0.13cm,width=\wid]{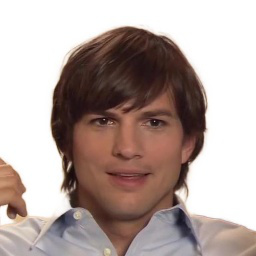} & \hspace{-0.0cm}
        \includegraphics[align=c,width=\wid]{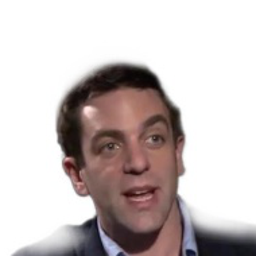} & \hspace{-0.0cm}
        \includegraphics[align=c,width=\wid]{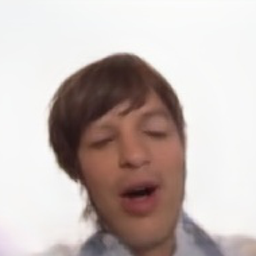} & \hspace{-0.0cm}
        \includegraphics[align=c,width=\wid]{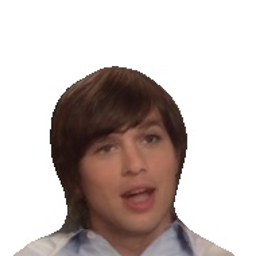} & \hspace{-0.0cm}
        \includegraphics[align=c,width=\wid]{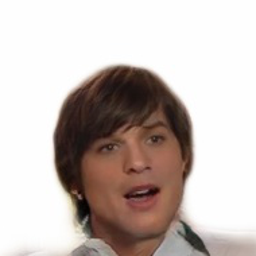} \\ %
        
        \includegraphics[align=c,bmargin=0.13cm,width=\wid]{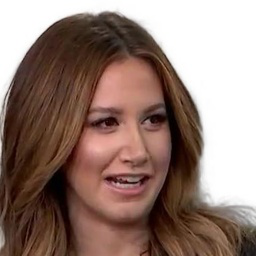} & \hspace{-0.0cm}
        \includegraphics[align=c,width=\wid]{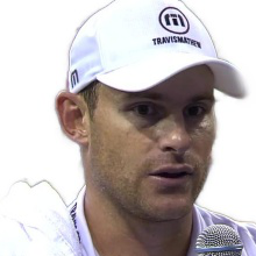} & \hspace{-0.0cm}
        \includegraphics[align=c,width=\wid]{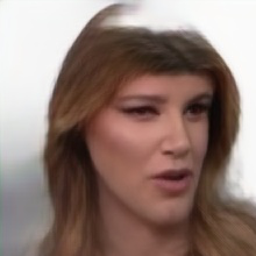} & \hspace{-0.0cm}
        \includegraphics[align=c,width=\wid]{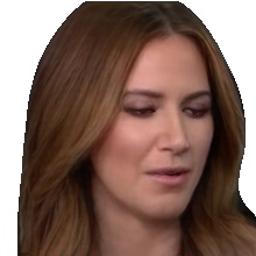} & \hspace{-0.0cm}
        \includegraphics[align=c,width=\wid]{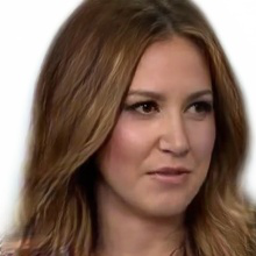} \\
        \includegraphics[align=c,bmargin=0.13cm,width=\wid]{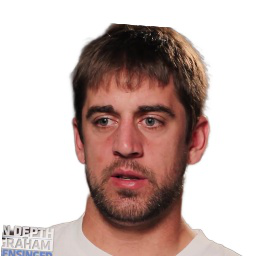} & \hspace{-0.0cm}
        \includegraphics[align=c,width=\wid]{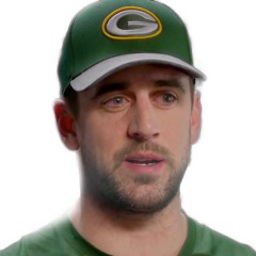} & \hspace{-0.0cm}
        \includegraphics[align=c,width=\wid]{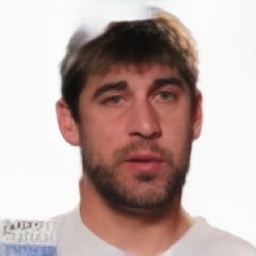} & \hspace{-0.0cm}
        \includegraphics[align=c,width=\wid]{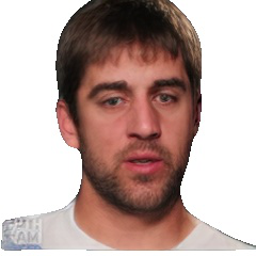} & \hspace{-0.0cm}
        \includegraphics[align=c,width=\wid]{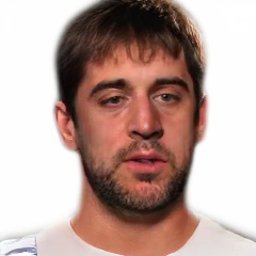} \\
        
        \includegraphics[align=c,bmargin=0.13cm,width=\wid]{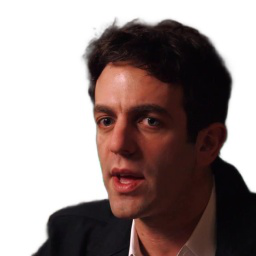} & \hspace{-0.0cm}
        \includegraphics[align=c,width=\wid]{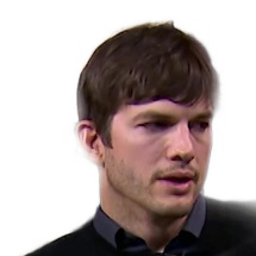} & \hspace{-0.0cm}
        \includegraphics[align=c,width=\wid]{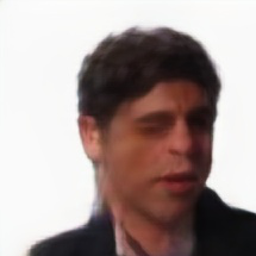} & \hspace{-0.0cm}
        \includegraphics[align=c,width=\wid]{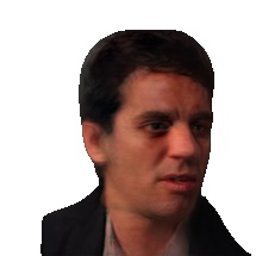} & \hspace{-0.0cm}
        \includegraphics[align=c,width=\wid]{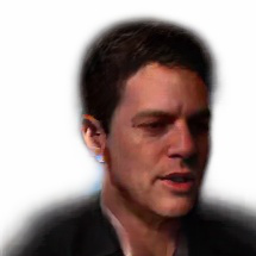} \\
        
        \includegraphics[align=c,bmargin=0.13cm,width=\wid]{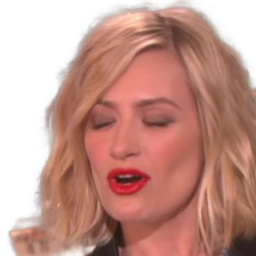} & \hspace{-0.0cm}
        \includegraphics[align=c,width=\wid]{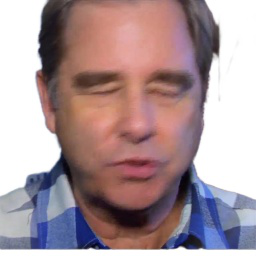} & \hspace{-0.0cm}
        \includegraphics[align=c,width=\wid]{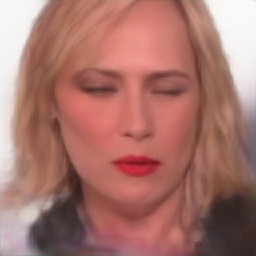} & \hspace{-0.0cm}
        \includegraphics[align=c,width=\wid]{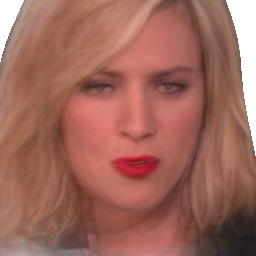} & \hspace{-0.0cm}
        \includegraphics[align=c,width=\wid]{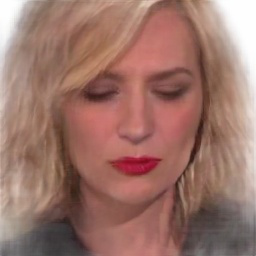} \\
        
        \includegraphics[align=c,bmargin=0.13cm,width=\wid]{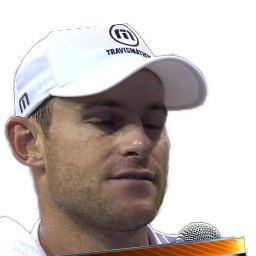} & \hspace{-0.0cm}
        \includegraphics[align=c,width=\wid]{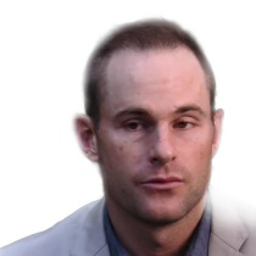} & \hspace{-0.0cm}
        \includegraphics[align=c,width=\wid]{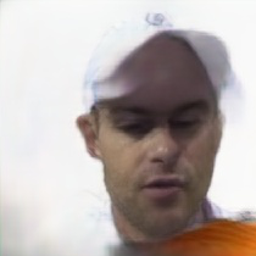} & \hspace{-0.0cm}
        \includegraphics[align=c,width=\wid]{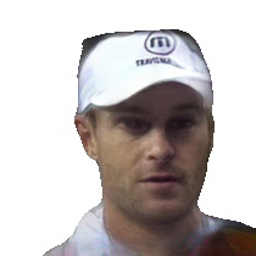} & \hspace{-0.0cm}
        \includegraphics[align=c,width=\wid]{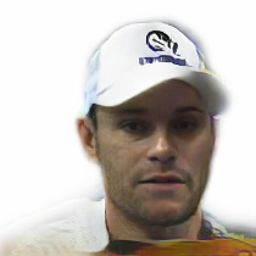} \\
        
        \includegraphics[align=c,bmargin=0.13cm,width=\wid]{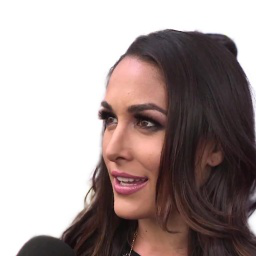} & \hspace{-0.0cm}
        \includegraphics[align=c,width=\wid]{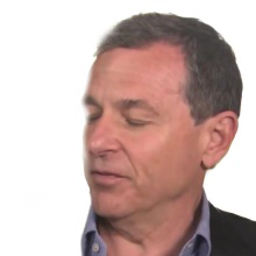} & \hspace{-0.0cm}
        \includegraphics[align=c,width=\wid]{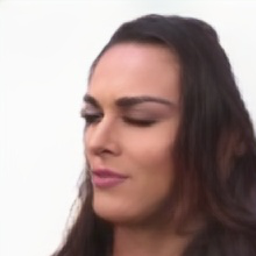} & \hspace{-0.0cm}
        \includegraphics[align=c,width=\wid]{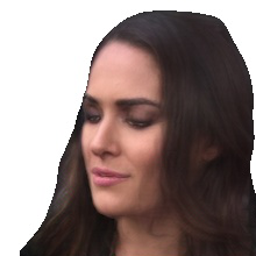} & \hspace{-0.0cm}
        \includegraphics[align=c,width=\wid]{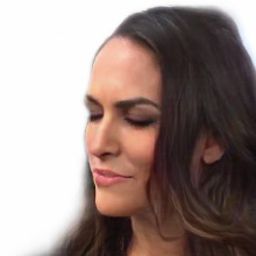} \\

        \textbf{Source} &\hspace{-0.0cm} \textbf{Driver} &\hspace{-0.0cm} \textbf{FOMM}  &\hspace{-0.0cm} \textbf{Bi-Layer} &\hspace{-0.0cm}  \textbf{ROME}
    \end{tabular}
    \caption{Additional comparison of renders on a VoxCeleb2 dataset. The task is to reenact the \textbf{source} image with the expression and pose of the \textbf{driver} image. This comparison is done in a cross-driving scenario, which we also use for quantitative comparison.}
    \label{fig:supp_vc2_cross_comp}
\end{figure*}

\begin{figure*}[h!]
    \centering    
    \setlength{\wid}{0.15\textwidth}
    \begin{tabular}{cccccc}

        \includegraphics[align=c,bmargin=0.13cm,width=\wid]{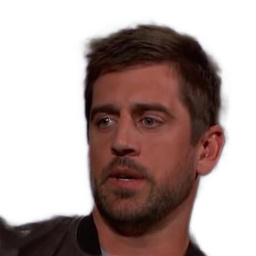} & \hspace{-0.0cm}
        \includegraphics[align=c,width=\wid]{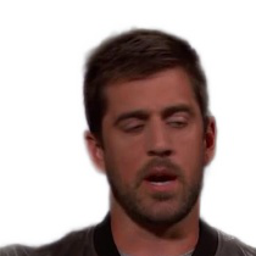} & \hspace{-0.0cm}
        \includegraphics[align=c,width=\wid]{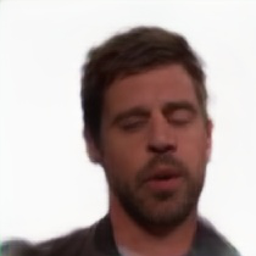} & \hspace{-0.0cm}
        \includegraphics[align=c,width=\wid]{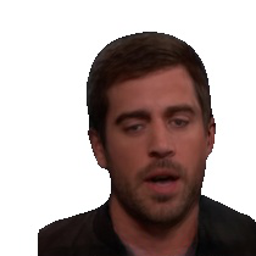} & \hspace{-0.0cm}
        \includegraphics[align=c,width=\wid]{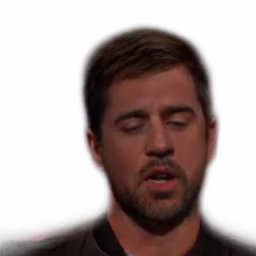} \\ %
         \includegraphics[align=c,bmargin=0.13cm,width=\wid]{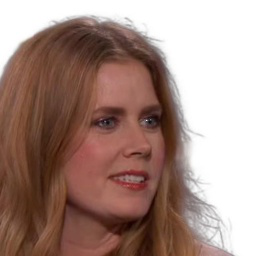} & \hspace{-0.4cm}
        \includegraphics[align=c,width=\wid]{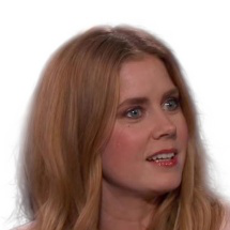} & \hspace{-0.0cm}
        \includegraphics[align=c,width=\wid]{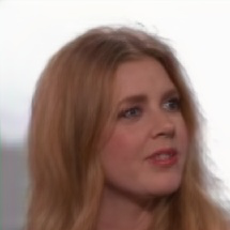} & \hspace{-0.0cm}
        \includegraphics[align=c,width=\wid]{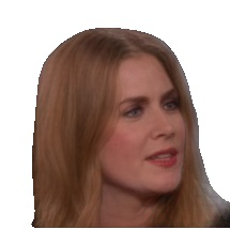} &
        \hspace{-0.0cm}
        \includegraphics[align=c,width=\wid]{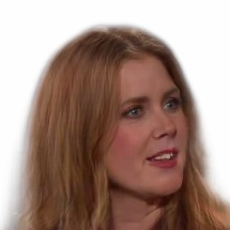} \\ %
        \includegraphics[align=c,bmargin=0.13cm,width=\wid]{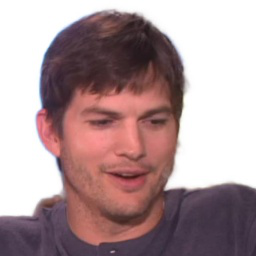} & \hspace{-0.4cm}
        \includegraphics[align=c,width=\wid]{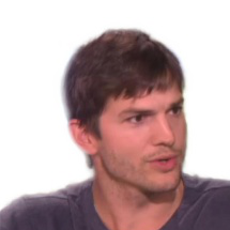} & \hspace{-0.0cm}
        \includegraphics[align=c,width=\wid]{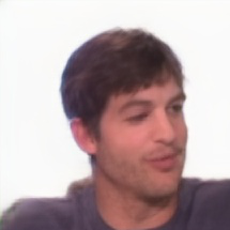} & \hspace{-0.0cm}
        \includegraphics[align=c,width=\wid]{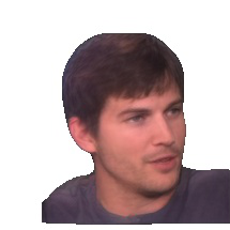} & \hspace{-0.0cm}
        \includegraphics[align=c,width=\wid]{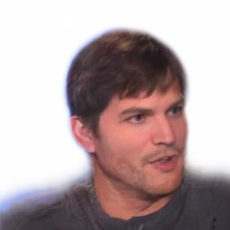} \\ %
        \includegraphics[align=c,bmargin=0.13cm,width=\wid]{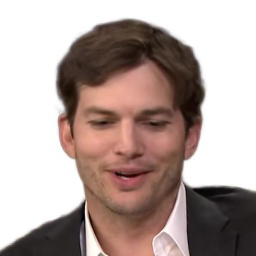} & \hspace{-0.0cm}
        \includegraphics[align=c,width=\wid]{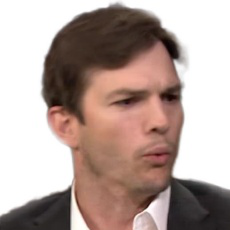} & \hspace{-0.0cm}
        \includegraphics[align=c,width=\wid]{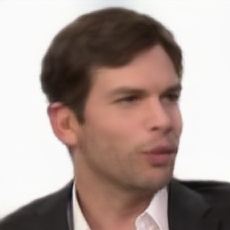} & \hspace{-0.0cm}
        \includegraphics[align=c,width=\wid]{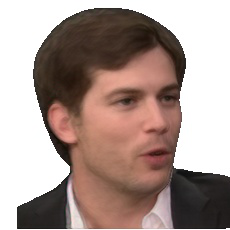} & \hspace{-0.0cm}
        \includegraphics[align=c,width=\wid]{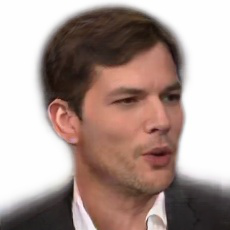} \\ %

        \includegraphics[align=c,bmargin=0.13cm,width=\wid]{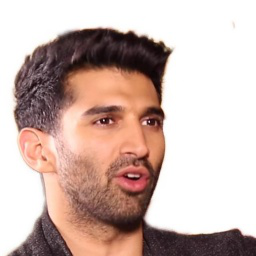} & \hspace{-0.0cm}
        \includegraphics[align=c,width=\wid]{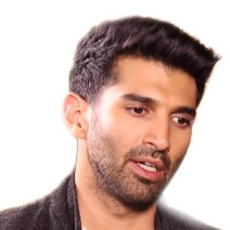} & \hspace{-0.0cm}
        \includegraphics[align=c,width=\wid]{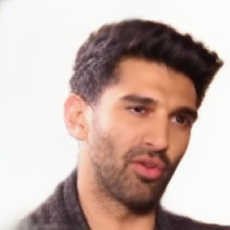} & \hspace{-0.0cm}
        \includegraphics[align=c,width=\wid]{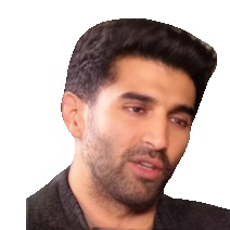} & \hspace{-0.0cm}
        \includegraphics[align=c,width=\wid]{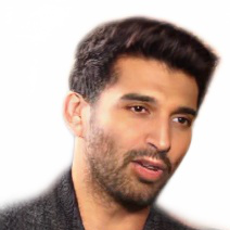} \\ %
        \includegraphics[align=c,bmargin=0.13cm,width=\wid]{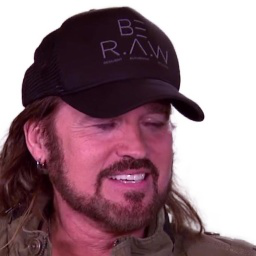} & \hspace{-0.0cm}
        \includegraphics[align=c,width=\wid]{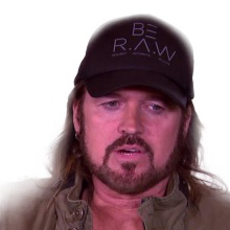} & \hspace{-0.0cm}
        \includegraphics[align=c,width=\wid]{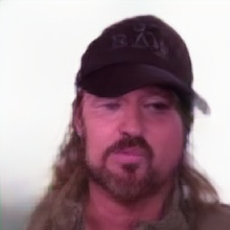} & \hspace{-0.0cm}
        \includegraphics[align=c,width=\wid]{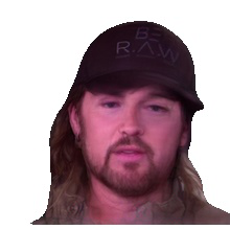} & \hspace{-0.0cm}
        \includegraphics[align=c,width=\wid]{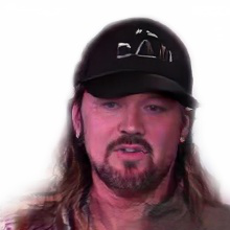} \\ %
        \includegraphics[align=c,bmargin=0.13cm,width=\wid]{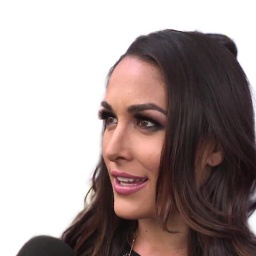} & \hspace{-0.0cm}
        \includegraphics[align=c,width=\wid]{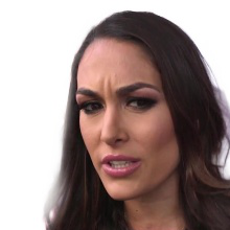} & \hspace{-0.0cm}
        \includegraphics[align=c,width=\wid]{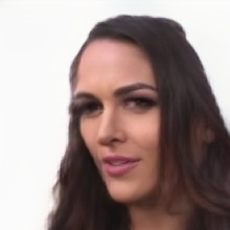} & \hspace{-0.0cm}
        \includegraphics[align=c,width=\wid]{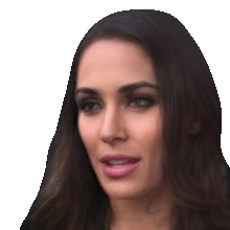} & \hspace{-0.0cm}
        \includegraphics[align=c,width=\wid]{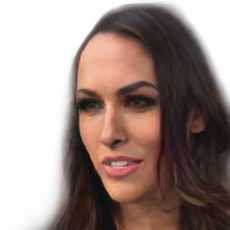} \\ %
        \includegraphics[align=c,bmargin=0.13cm,width=\wid]{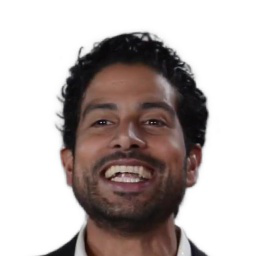} & \hspace{-0.0cm}
        \includegraphics[align=c,width=\wid]{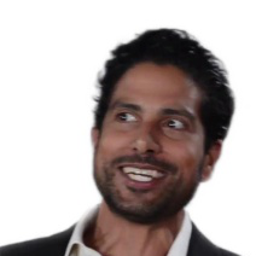} & \hspace{-0.0cm}
        \includegraphics[align=c,width=\wid]{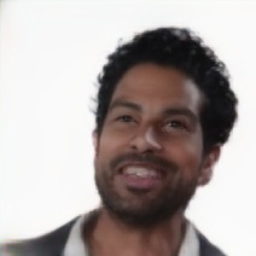} & \hspace{-0.0cm}
        \includegraphics[align=c,width=\wid]{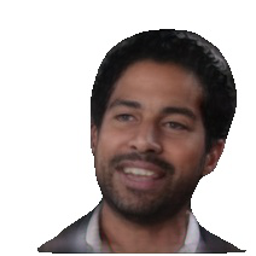} & \hspace{-0.0cm}
        \includegraphics[align=c,width=\wid]{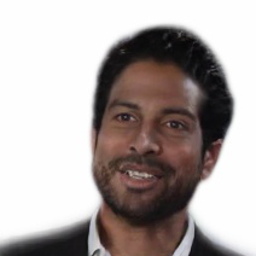} \\ %

        \textbf{Source} &\hspace{-0.0cm} \textbf{Driver} &\hspace{-0.0cm} \textbf{FOMM}  &\hspace{-0.0cm} \textbf{Bi-Layer} &\hspace{-0.0cm}  \textbf{ROME}
    \end{tabular}
    \caption{Additional comparison of renders on a VoxCeleb2 dataset. The task is to reenact the \textbf{source} image with the expression and pose of the \textbf{driver} image. This comparison is done in a self-driving scenario, which we also use for quantitative comparison.}
    \label{fig:supp_vc2_self_comp}
\end{figure*}

We provide an extended cross-driving qualitative comparison with neural-based rendering methods in~\fig{supp_vc2_cross_comp}. Then, we provide an additional self-driving qualitative comparison in~\fig{supp_vc2_self_comp}. Most of the results are consistent with the metrics obtained in the main text.

\begin{table}[!t]
    \begin{subtable}{\linewidth}
        \centering
        \begin{tabular}{ l c c c c c c}
            Method & \multicolumn{2}{c}{LPIPS$\downarrow$} & \multicolumn{2}{c}{SSIM$\uparrow$} & \multicolumn{2}{c}{PSNR$\uparrow$} \\
            \hline
            \hline
            w/o $\Delta\hat\v$ &  \multicolumn{2}{c}{$0.10$} 
            & \multicolumn{2}{c}{$0.81$}
            & \multicolumn{2}{c}{$23.1$} \\
            ROME &   \multicolumn{2}{c}{$0.08$} 
            & \multicolumn{2}{c}{$0.86$} & \multicolumn{2}{c}{$25.8$}  \\
        \end{tabular}
    \end{subtable}
    \caption{Quantitative ablation on a hold-out set of the VoxCeleb2 dataset. We observe that the deferred neural rendering trained without offsets achieves lower image quality for face and hair regions, than a full ROME system.}\label{tab:supp_vc2_abl}
\end{table}

We evaluate the effect that trained offsets have on the quality of rendering. To do that, we remove the head reconstruction step from the training pipeline and only train deferred neural rendering system using base FLAME meshes.
The quantitative results are in~\tab{supp_vc2_abl}. 
Notice how the quality degrades for the model with no offsets in the hair and shoulders areas and to the overall worse quantitatively measured performance. This leads us to the conclusion that our system for coarse mesh estimation can aid other neural rendering systems produce better quality reconstructions.

\subsection{Linear model}

\begin{figure*}[!h]
    \centering    
    \setlength{\wid}{0.110\textwidth}
    \def\bmarg{0.12cm}
    \def\sep{-0.1cm}
    \def\halfbigsep{0.2cm}
    \begin{tabular}{cccc cccc}
    
        \includegraphics[bmargin=\bmarg,width=\wid]{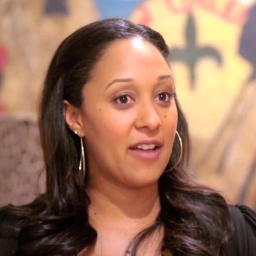} & \hspace{\sep}
        \includegraphics[width=\wid]{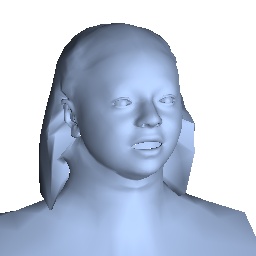} & \hspace{\sep}
        \includegraphics[width=\wid]{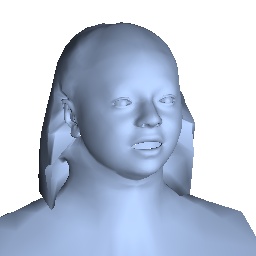} & \hspace{\sep}
        \includegraphics[width=\wid]{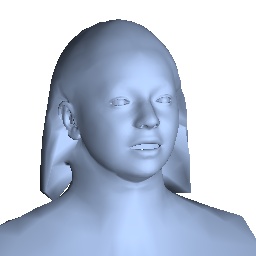} \hspace{\sep}
        \hspace{\halfbigsep} & \hspace{\halfbigsep}
        \includegraphics[width=\wid]{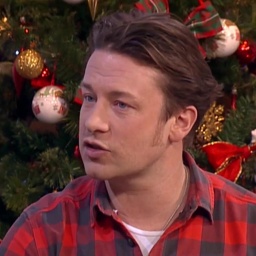} & \hspace{\sep}
        \includegraphics[width=\wid]{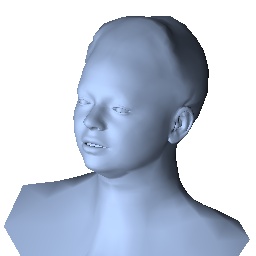} & \hspace{\sep}
        \includegraphics[width=\wid]{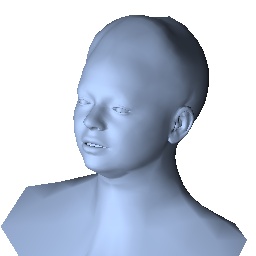} & \hspace{\sep}
        \includegraphics[width=\wid]{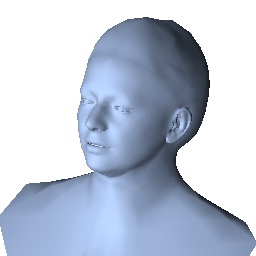} \\
        
        \includegraphics[bmargin=\bmarg,width=\wid]{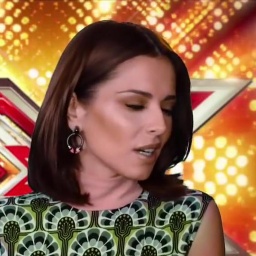} & \hspace{\sep}
        \includegraphics[width=\wid]{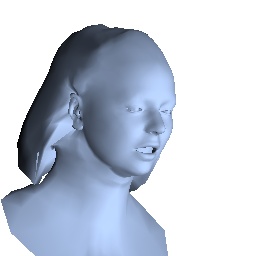} & \hspace{\sep}
        \includegraphics[width=\wid]{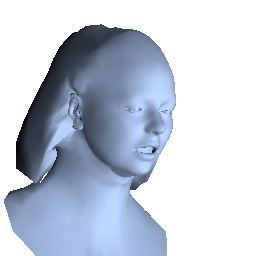} & \hspace{\sep}
        \includegraphics[width=\wid]{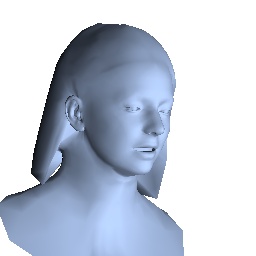} \hspace{\sep}
        \hspace{\halfbigsep} & \hspace{\halfbigsep}
        \includegraphics[width=\wid]{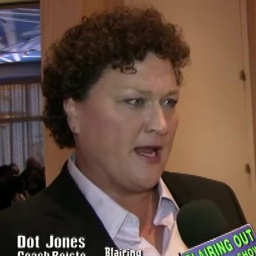} & \hspace{\sep}
        \includegraphics[width=\wid]{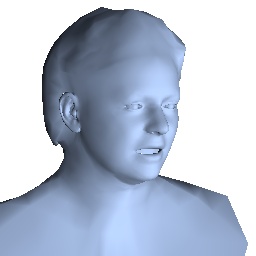} & \hspace{\sep}
        \includegraphics[width=\wid]{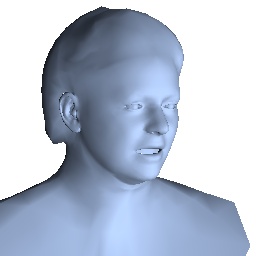} & \hspace{\sep}
        \includegraphics[width=\wid]{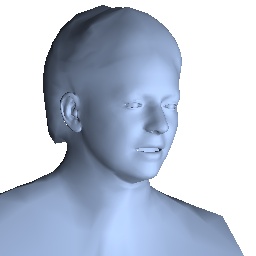} \\
        
        \includegraphics[width=\wid]{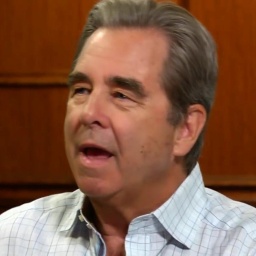} & \hspace{\sep}
        \includegraphics[width=\wid]{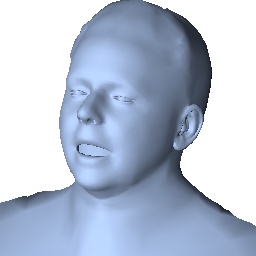} & \hspace{\sep}
        \includegraphics[width=\wid]{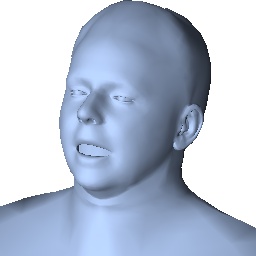} & \hspace{\sep}
        \includegraphics[width=\wid]{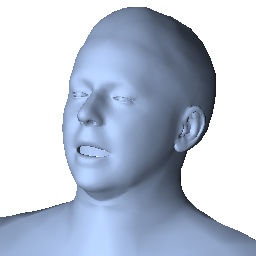}
        \hspace{\halfbigsep} & \hspace{\halfbigsep}
        \includegraphics[width=\wid]{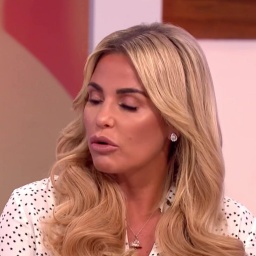} & \hspace{\sep}
        \includegraphics[width=\wid]{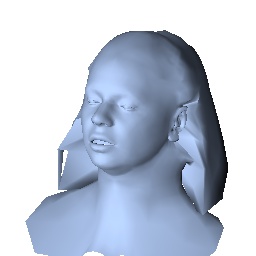} & \hspace{\sep}
        \includegraphics[bmargin=\bmarg,width=\wid]{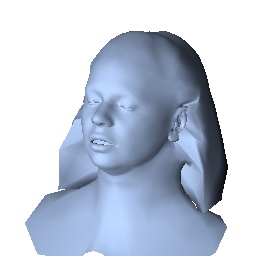} & \hspace{\sep}
        \includegraphics[width=\wid]{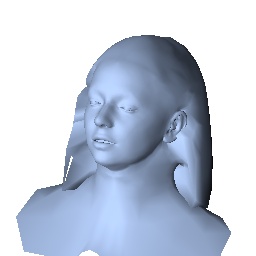} \\ 

        \textbf{Image} & \hspace{\sep} \textbf{ROME} & \hspace{\sep} \textbf{PCA} &\hspace{\sep} \textbf{Distilled} \hspace{\halfbigsep} & \hspace{\halfbigsep} \textbf{Image} & \hspace{\sep} \textbf{ROME} & \hspace{\sep} \textbf{PCA} &\hspace{\sep} \textbf{Distilled}  \\
    \end{tabular}
    \caption{We provide more examples to qualitatively evaluate the performance of the distilled linear model.}\label{fig:supp_linear_model_comp}
\end{figure*}

 We evaluate visual quality in~\fig{supp_linear_model_comp}. We note that the value of MSE that is achieved by our regressor leads to visually similar reconstructions. Our mesh reconstruction model can achieve up to \textbf{10 times} speed-up without any perceived degradation in reconstruction quality.

\begin{figure*}[h!]
    \centering    
    \setlength{\wid}{0.15\textwidth}
    \begin{tabular}{ccccccc}
        1 &
        \includegraphics[width=\wid]{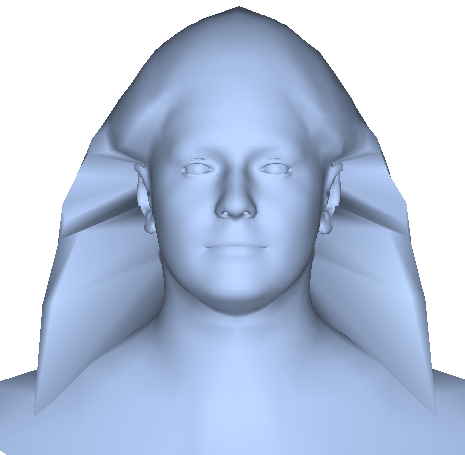} & \hspace{-0.4cm}
        \includegraphics[width=\wid]{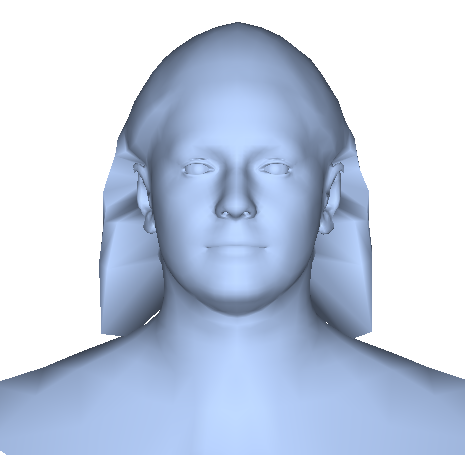} & \hspace{-0.4cm}
        \includegraphics[width=\wid]{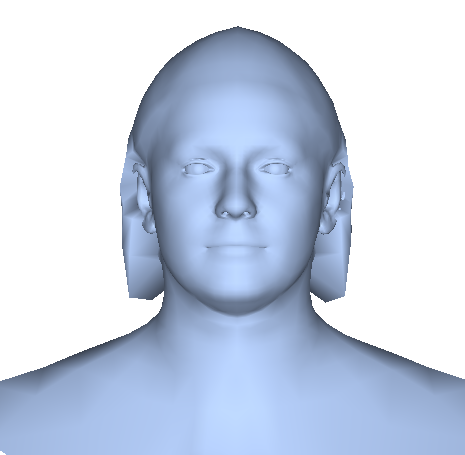} & \hspace{-0.4cm}
        \includegraphics[width=\wid]{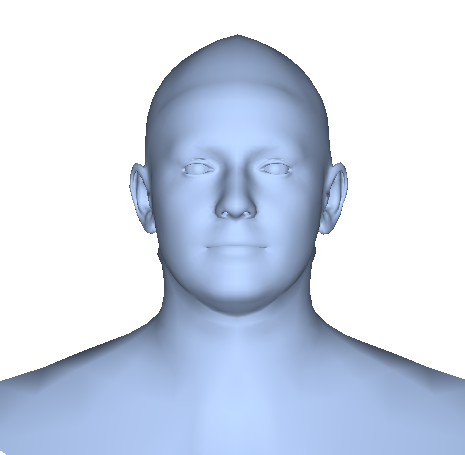} & \hspace{-0.4cm}
        \includegraphics[width=\wid]{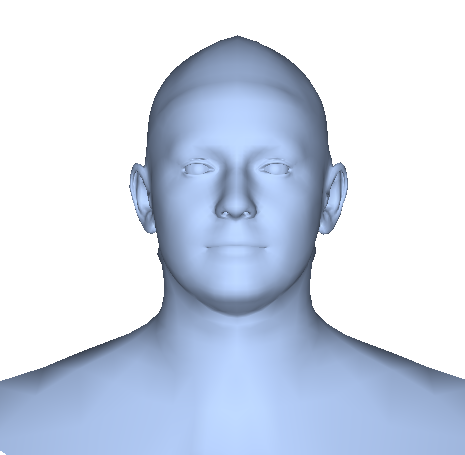} & \hspace{-0.4cm}
        \includegraphics[width=\wid]{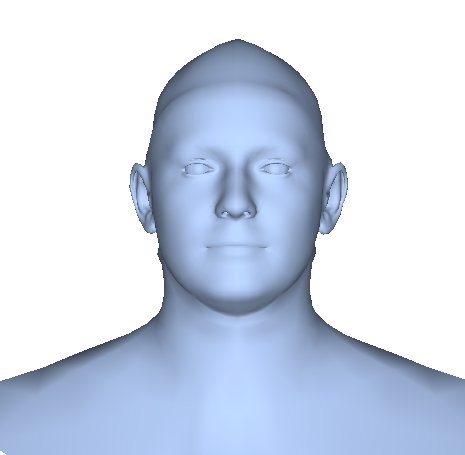} \\ %
        2 &
        \includegraphics[width=\wid]{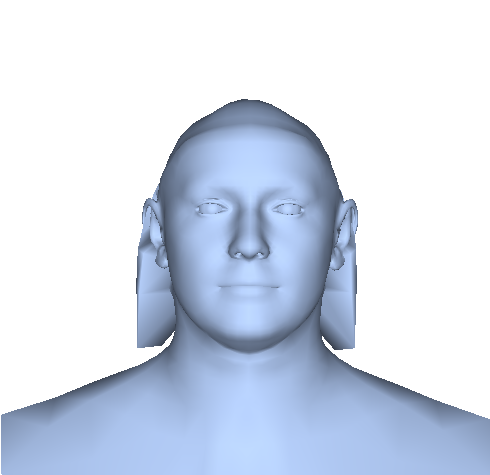} & \hspace{-0.4cm}
        \includegraphics[width=\wid]{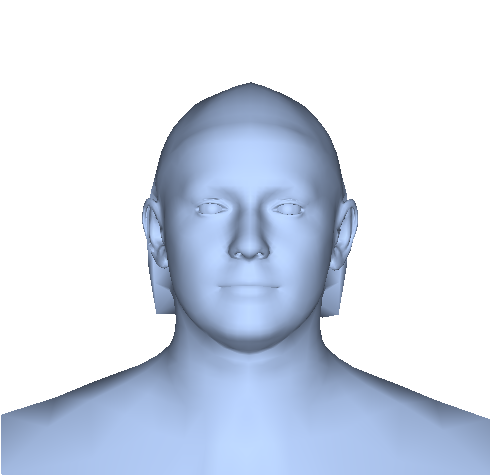} & \hspace{-0.4cm}
        \includegraphics[width=\wid]{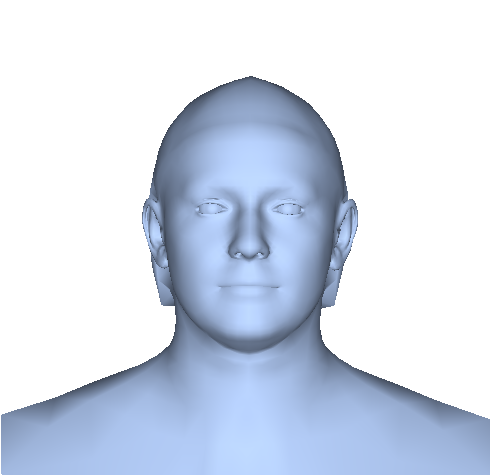} & \hspace{-0.4cm}
        \includegraphics[width=\wid]{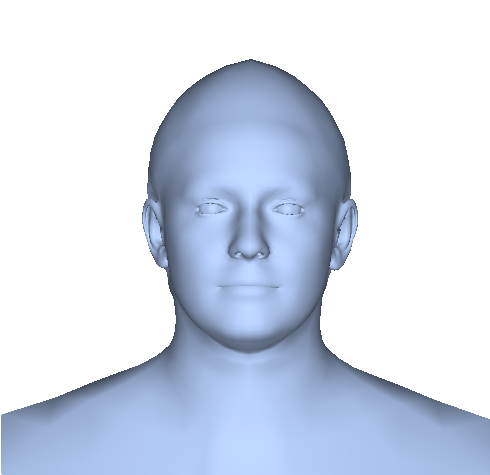} & \hspace{-0.4cm}
        \includegraphics[width=\wid]{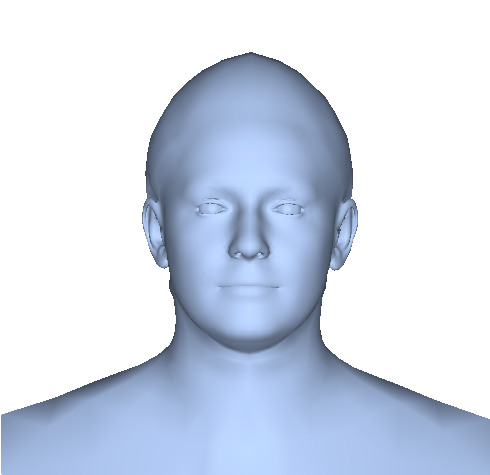} & \hspace{-0.4cm}
        \includegraphics[width=\wid]{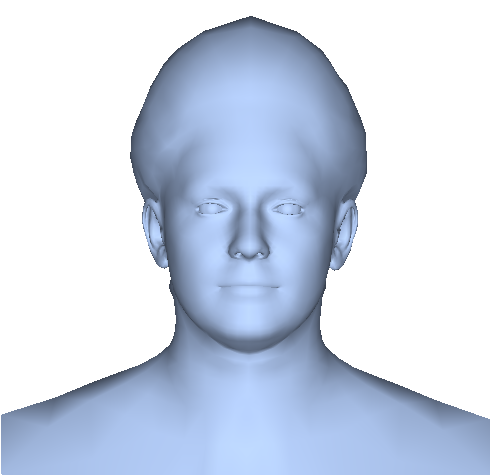} \\ 
        4 &
        \includegraphics[width=\wid]{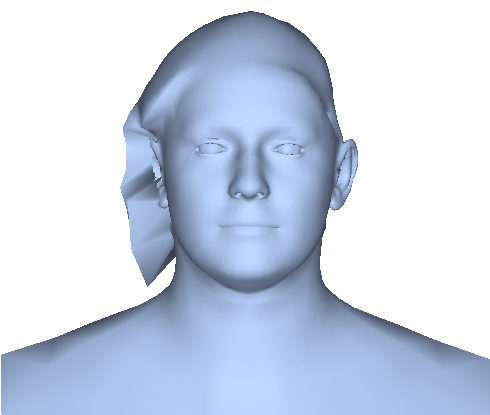} & \hspace{-0.4cm}
        \includegraphics[width=\wid]{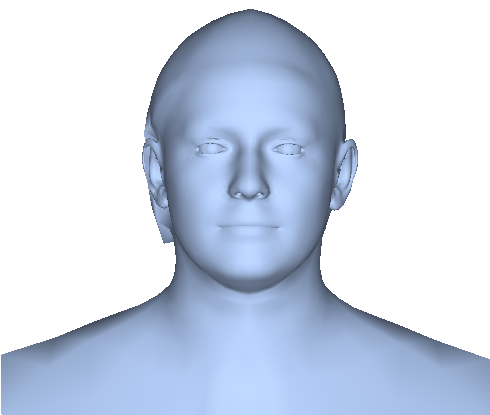} & \hspace{-0.4cm}
        \includegraphics[width=\wid]{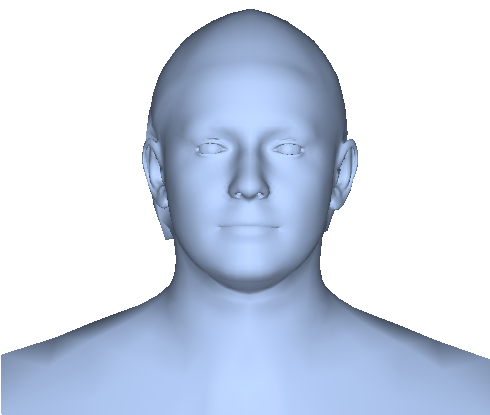} & \hspace{-0.4cm}
        \includegraphics[width=\wid]{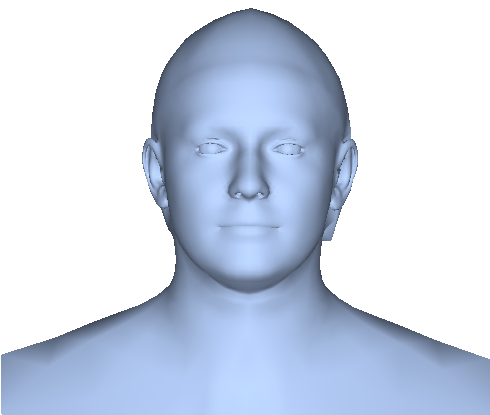} & \hspace{-0.4cm}
        \includegraphics[width=\wid]{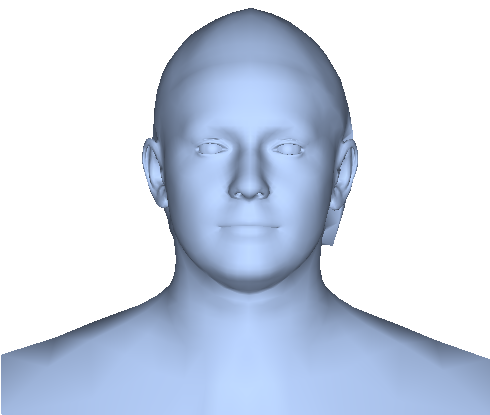} & \hspace{-0.4cm}
        \includegraphics[width=\wid]{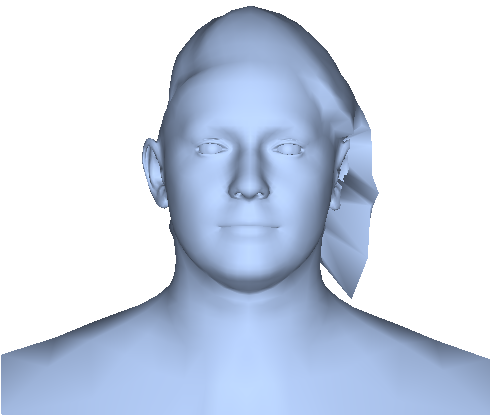} \\ 
        52 &
        \includegraphics[width=\wid]{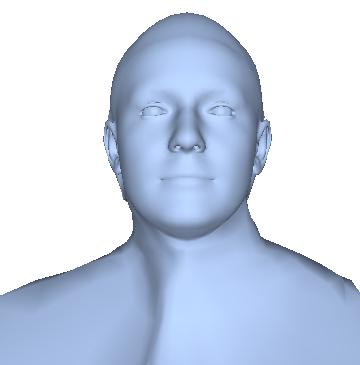} & \hspace{-0.4cm}
        \includegraphics[width=\wid]{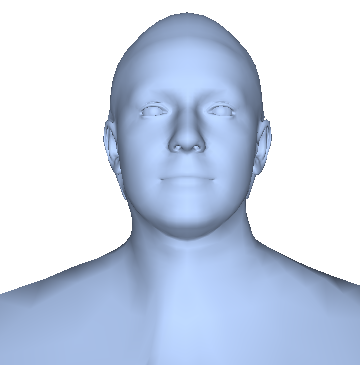} & \hspace{-0.4cm}
        \includegraphics[width=\wid]{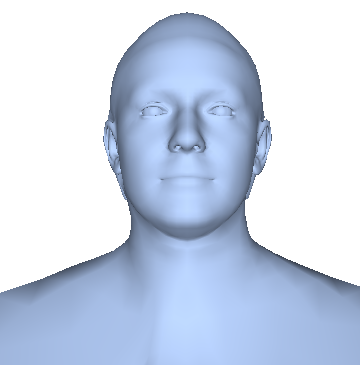} & \hspace{-0.4cm}
        \includegraphics[width=\wid]{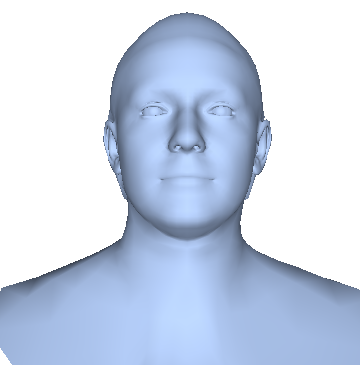} & \hspace{-0.4cm}
        \includegraphics[width=\wid]{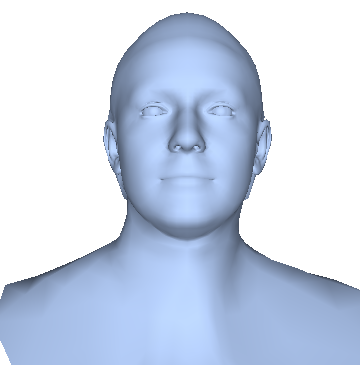} & \hspace{-0.4cm}
        \includegraphics[width=\wid]{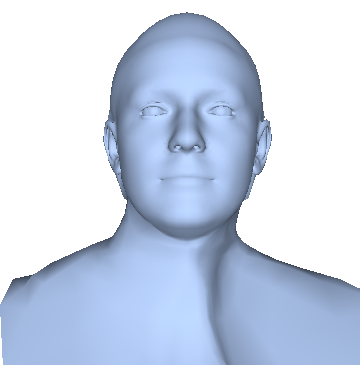} \\ 
        
        $k$ & $\eta^k_{(1)}$ &\hspace{-0.4cm} $\eta^k_{(\lfloor 0.1M \rfloor)}$ &\hspace{\mrgone} $\eta^k_{(\lfloor 0.25 M \rfloor)}$  &\hspace{\mrgone} $\eta^k_{(\lfloor 0.75 M \rfloor)}$ &\hspace{\mrgone} $\eta^k_{(\lfloor 0.9M \rfloor)}$ &\hspace{\mrgone} $\eta^k_{(M)}$
    \end{tabular}
    \caption{We show how the estimated meshes can be semantically manipulated by varying the individual components of the PCA basis. We assume that $M$ meshes were initially predicted by the ROME system. Then, we estimate $M$ coefficients $\eta_m \in \mathbb{R}^K$, each is used to reconstruct $m$-th mesh via the PCA basis. In our experiments, the set of 50 basis vectors are used to reconstruct the hair, and the set of 10 basis vectors are used to reconstruct the neck and the shoulders. We denote each component of $\eta_m$ as $\eta_m^k$. Then, we use the mean vector $\eta = \frac{1}{M} \eta_m$ to reconstruct the base mesh, and modify its $k$-th component to the $p$-th order statistic $\eta_{(p)}^k$ over the dataset $\{ \eta_m^k \}_{m=1}^M$, where $p \in \{1, \dots, M \}$. We show the resulting reconstructions above. Each row corresponds to a component which is modified, and each column corresponds to its new value (minimum, 10\%, 25\%, 75\%, 90\%, maximum). The first three lines correspond to the hair components, and the last line -- to the neck component.} %
    \label{fig:supp_pca_basis}
\end{figure*}

\begin{figure*}[h!]
    \centering    
    \setlength{\wid}{0.15\textwidth}
    \begin{tabular}{cccccc}
        \includegraphics[width=\wid]{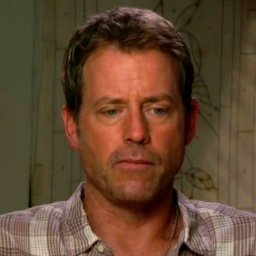} & \hspace{\mrgone}
        \includegraphics[width=\wid]{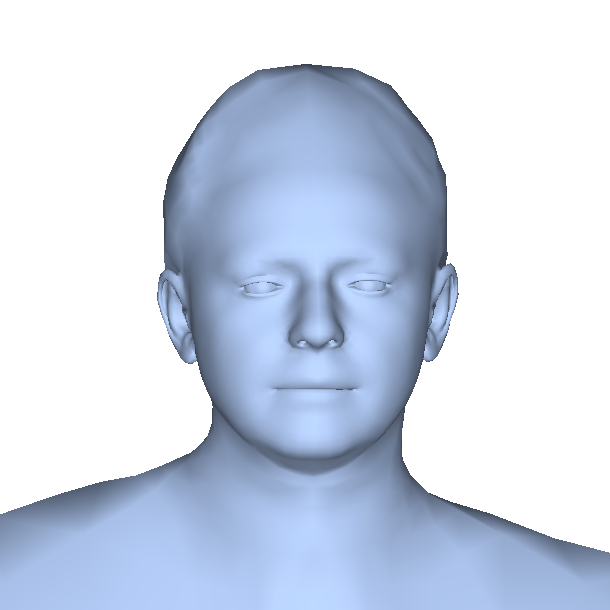} & \hspace{\mrgone}
        \includegraphics[width=\wid]{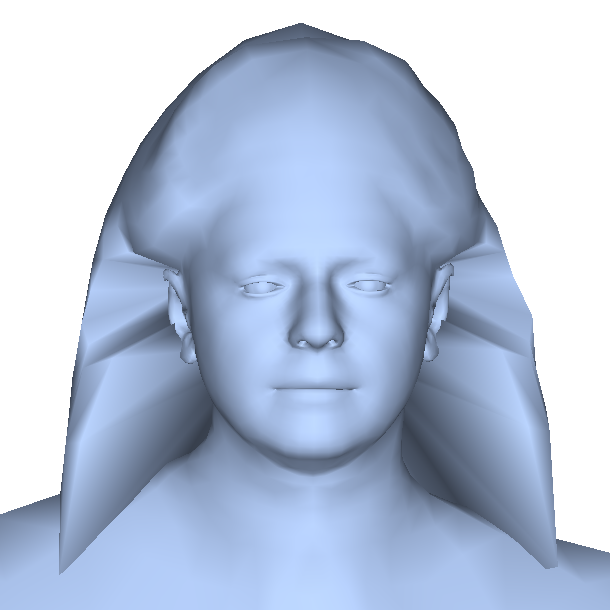} & \hspace{\mrgone}
        \includegraphics[width=\wid]{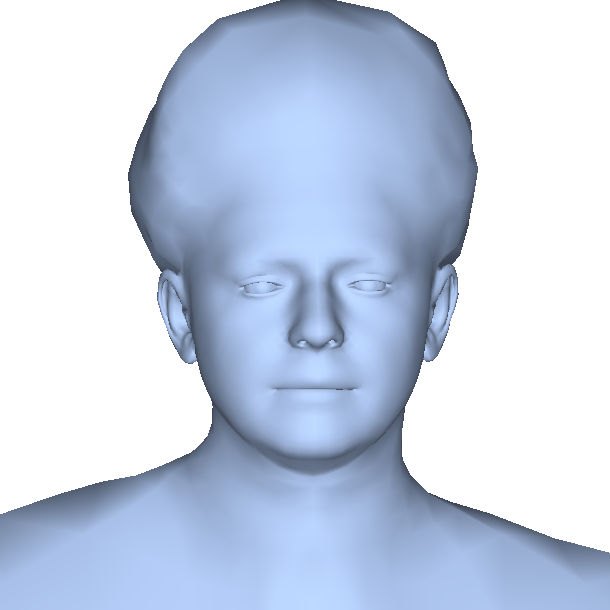} & \hspace{\mrgone}
        \includegraphics[width=\wid]{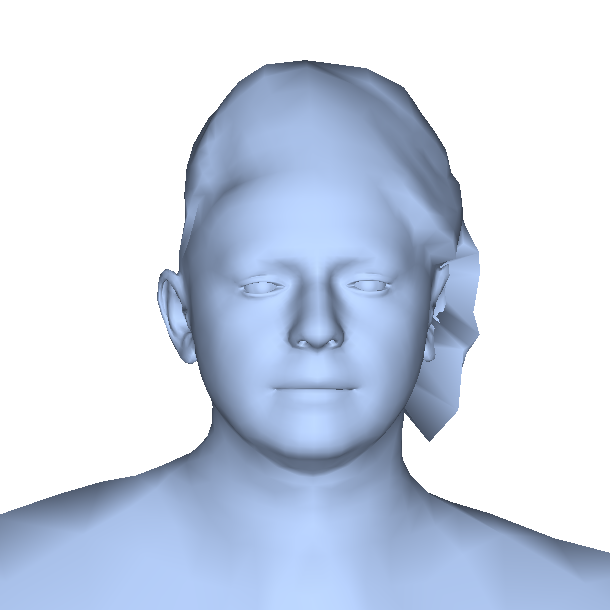} &
        \hspace{\mrgone}
       \includegraphics[width=\wid]{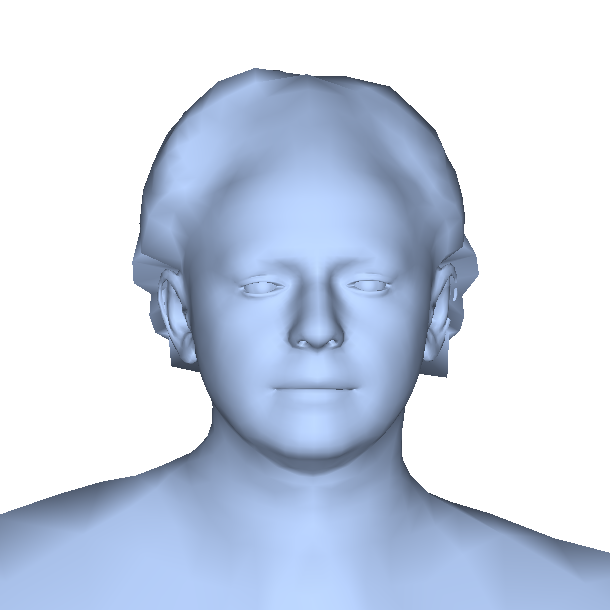} \\ %
        \includegraphics[width=\wid]{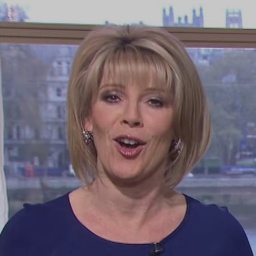}& \hspace{\mrgone}
        \includegraphics[width=\wid]{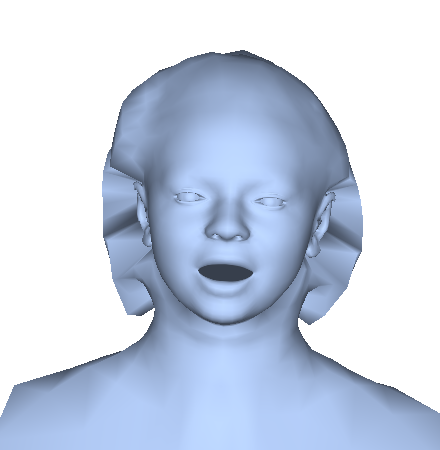} & \hspace{\mrgone}
        \includegraphics[width=\wid]{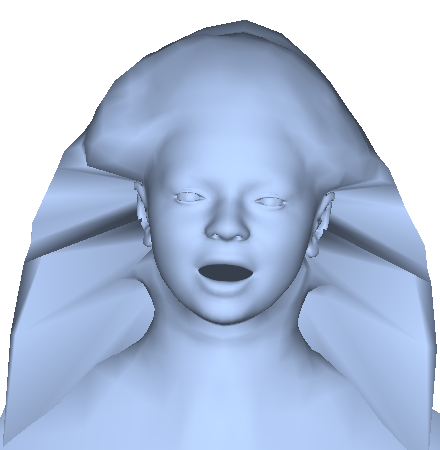} & \hspace{\mrgone}
        \includegraphics[width=\wid]{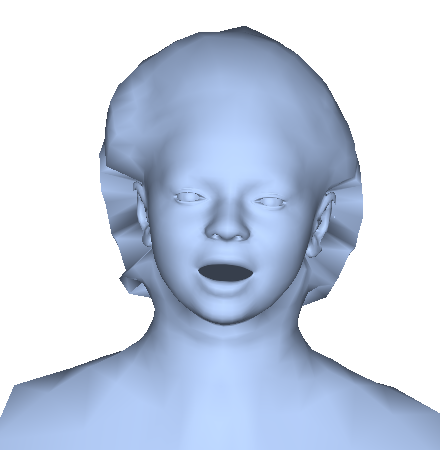} & \hspace{\mrgone}
        \includegraphics[width=\wid]{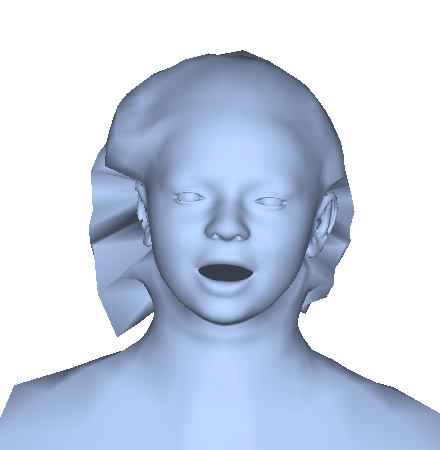} & \hspace{\mrgone}
        \includegraphics[width=\wid]{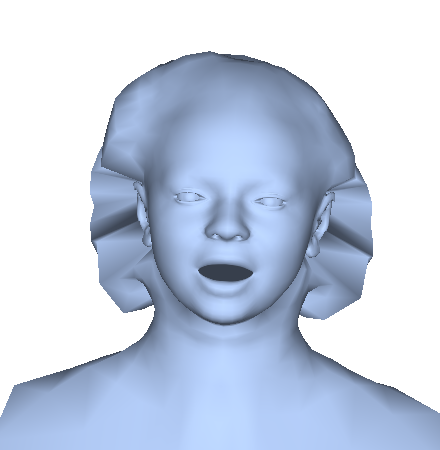} \\ %

        \textbf{Image} &\hspace{\mrgone} \textbf{Original} &\hspace{\mrgone} \textbf{Edit 1}  &\hspace{\mrgone} \textbf{Edit 2} &\hspace{\mrgone} \textbf{Edit 4} &\hspace{\mrgone} \textbf{Edit 5}
    \end{tabular}
    \caption{Examples of hair style manipulation using only individual components of a PCA basis.}
    \label{fig:supp_linear_model_manip}
\end{figure*}

Additionally, we evaluate the semantic manipulation capabilities of the linear model in a similar way to the face parametric models. Specifically, we pick individual basis vectors and see how varying their coefficient changes the reconstructed mesh. The results can be seen in~\fig{supp_pca_basis}. We observe a certain semantic disentanglement for the first hair and neck basis vectors. This disentanglement allows us to perform mesh and image editing tasks, like the editing of hairstyle, which we show in~\fig{supp_linear_model_manip}. Here, we obtain this modified reconstruction by simply varying one of the predicted coefficients for the linear basis.

\end{document}